\documentclass{article}

\usepackage{microtype}
\usepackage{graphicx}
\usepackage{subfigure}
\usepackage{booktabs} % for professional tables

\usepackage{amsmath,amsfonts,bm}

\def\eqref#1{equation~\ref{#1}}
\def\1{\bm{1}}

\def\vd{{\bm{d}}}

\def\vp{{\bm{p}}}

\def\vx{{\bm{x}}}

\def\mD{{\bm{D}}}

\def\mI{{\bm{I}}}

\def\mM{{\bm{M}}}
\def\mN{{\bm{N}}}

\def\mR{{\bm{R}}}

\def\mU{{\bm{U}}}

\def\mW{{\bm{W}}}

\DeclareMathAlphabet{\mathsfit}{\encodingdefault}{\sfdefault}{m}{sl}
\SetMathAlphabet{\mathsfit}{bold}{\encodingdefault}{\sfdefault}{bx}{n}

\def\sA{{\mathbb{A}}}
\def\sB{{\mathbb{B}}}

\def\sI{{\mathbb{I}}}

\def\sR{{\mathbb{R}}}
\def\sS{{\mathbb{S}}}

\def\sX{{\mathbb{X}}}

\newcommand{\E}{\mathbb{E}}

\newcommand{\R}{\mathbb{R}}

\usepackage{mathtools} 
\usepackage{graphicx} 
\usepackage{amsthm,amssymb}
\usepackage{xcolor}
\usepackage{caption}
\usepackage{multirow}

\newtheorem{lemma}{Lemma}
\newtheorem{prop}{Proposition}

\usepackage{hyperref}

\usepackage{xr-hyper} 
\usepackage{cleveref}
\usepackage{lineno,hyperref}

\usepackage[accepted]{icml2020}

\icmltitlerunning{Option Discovery in the Absence of Rewards with Manifold Analysis}
\begin{document}

\twocolumn[
\icmltitle{Option Discovery in the Absence of Rewards with Manifold Analysis}

% It is OKAY to include author information, even for blind
% submissions: the style file will automatically remove it for you
% unless you've provided the [accepted] option to the icml2019
% package.

% List of affiliations: The first argument should be a (short)
% identifier you will use later to specify author affiliations
% Academic affiliations should list Department, University, City, Region, Country
% Industry affiliations should list Company, City, Region, Country

% You can specify symbols, otherwise they are numbered in order.
% Ideally, you should not use this facility. Affiliations will be numbered
% in order of appearance and this is the preferred way.
\icmlsetsymbol{equal}{*}

\begin{icmlauthorlist}
\icmlauthor{Amitay Bar}{Technion}
\icmlauthor{Ronen Talmon}{Technion}
\icmlauthor{Ron Meir}{Technion}
\end{icmlauthorlist}

\icmlaffiliation{Technion}{Viterbi Faculty of Electrical Engineering, Technion, Israel Institute of Technology
}

\icmlcorrespondingauthor{Amitay Bar}{amitayb@campus.technion.ac.il}

% You may provide any keywords that you
% find helpful for describing your paper; these are used to populate
% the "keywords" metadata in the PDF but will not be shown in the document
\icmlkeywords{Machine Learning, ICML}

\vskip 0.3in
]

% this must go after the closing bracket ] following \twocolumn[ ...

% This command actually creates the footnote in the first column
% listing the affiliations and the copyright notice.
% The command takes one argument, which is text to display at the start of the footnote.
% The \icmlEqualContribution command is standard text for equal contribution.
% Remove it (just {}) if you do not need this facility.

%\printAffiliationsAndNotice{}  % leave blank if no need to mention equal contribution
\printAffiliationsAndNotice{} % otherwise use the standard text.

% TODO: author list and affiliations

% TODO: Mansour: Euler tour - good only for representing trees. BUT, for every connected graph, one can find a spanning tree (O(n), n - # of nodes using BFS or DFS), and then in the worst case, the length of a trajectory visiting all the states is at most 2n, and the expected length (expectation wrt the starting state) is n. Conversely, it is clear that there are graphs, e.g., dumbbell or two complete graphs connected by a path graph, for which a random walk will perform much worse than that. So, if the entire graph is known, the agent can store the spanning tree and march accordingly. So the random walk makes sense only if the graph is unknown (a-priori, or partially known). Anyway (although it's also an issue in previous work), we may consider stating the our scope is limited to agents with a policy that performs random walk with primitive actions or options. Alternatively, we need to claim that what we propose is beneficial only if the entire graph is not a-priori known and/or if the graph is unknown to the agent at the learning stage... Also, from the above, it seems that it makes sense to minimize the cover time. However, Jinai does it only for `RW' agents, and actually minimizes the upper bound rather the cover time itself.

\begin{abstract}
Options have been shown to be an effective tool in reinforcement learning, facilitating improved exploration and learning. In this paper, we present an approach based on spectral graph theory and derive an algorithm that systematically discovers options without access to a specific reward or task assignment. As opposed to the common practice used in previous methods, our algorithm makes full use of the spectrum of the graph Laplacian.
Incorporating modes associated with higher graph frequencies unravels domain subtleties, which are shown to be useful for option discovery. Using geometric and manifold-based analysis, we present a theoretical justification for the algorithm. In addition, we showcase its performance in several domains, demonstrating clear improvements compared to competing methods.

\end{abstract}

\section{Introduction} 
Reinforcement learning (RL) has attracted much attention in recent years thanks to its success in solving a broad range of challenging tasks.
Options (a.k.a.~skills) play an important role in RL  \citep{sutton1999between} and have opened the door to a series of studies demonstrating improvement in both learning and exploration  \citep{vezhnevets2017feudal,nachum2018data,eysenbach2018diversity,tang2017exploration,mannor2004dynamic,menache2002q}.
One important class of options consists of options that are not associated with any specific task and are acquired without receiving any reward. Such generic options often lead to efficient learning in various tasks that are not known a-priori, (e.g., \citep{eysenbach2018diversity}).

An effective approach to build such options is based on spectral graph theory, assuming a finite state domain in which each state is regarded as a node of a graph, and the graph edges represent the states connectivity. Such an approach led to the introduction of proto-value functions (PVFs)\citep{mahadevan2007proto}, which are the eigenvectors of the graph Laplacian \cite{chung1997spectral}. It was shown that the PVFs establish an efficient representation of the domain. Recently, these PVFs were used for options representation  \citep{machado2017laplacian,machado2017eigenoption}. There, eigenoptions were introduced by considering only the dominant eigenvectors (PVFs), where each eigenoption is formed based on a single eigenvector. In a related work,  \citet{jinnai2019discovering} presented cover options using only the Fiedler vector multiple times. On the one hand, option discovery with a graph-based representation is a powerful combination, since it facilitates options that are not task or reward-specific, yet it naturally incorporates the geometry of the domain. On the other hand, existing methods are based only on a single eigenvector or consider only the dominant eigenvectors while omitting the rest, leaving room for improvement and further investigation.

In this paper, we present a new scheme for defining options, relying on all the eigenvectors of the graph Laplacian. More concretely, we form a score function built from the eigenvectors, from which options can be systematically derived.
Since the agent acts without receiving reward, it is only natural to discover and analyze the options considering the geometry of the domain.
For analysis purposes, we model the domain as a manifold and consequently the graph as a discrete approximation of the manifold, allowing us to incorporate concepts and results from manifold learning, such as the diffusion distance \cite{coifman2006diffusion}.
We show that our options lead to improved performance both in learning and exploration compared to the eigenoptions as well as other option discovery schemes. 

Our main contributions are as follows.  
First, we present a new approach to principled option discovery with a theoretical foundation based on geometric and manifold analysis.
Second, this analysis includes novel results in manifold learning involving two key components: the stationary distribution of a random walk on a graph and the diffusion distance. To obtain these results, we employ a new concept in manifold learning, in which the entire spectrum of the underlying graph is considered rather than only its leading components.
Third, we propose an algorithm for option discovery, applicable in high-dimensional deterministic domains. We empirically demonstrate that the learning performance obtained by our options outperforms competing options on three small-scale domains. In addition, we show extensions to stochastic domains and to large scale domains.

\section{Background} 
\subsection{RL and Options}
We use the Markov decision process (MDP) framework to formulate the RL problem \citep{puterman2014markov}. An MDP is a 5-tuple $\left\langle \sS,\sA,p,r,\gamma\right\rangle$, where $\sS$ is the set of states, $\sA$ is the set of actions, $p$ is the transition probability such that $p\left(s'|s,a\right)$ is the probability of moving from state $s$ to state $s'$ by taking an action $a$, $r(s,a,s')$ is the reward function and $\gamma \in [0,1) $ is a discount factor. 
Consider an agent operating sequentially so that at time step $n$ it moves from state $s_n$ to state $s_{n+1}$, receiving a reward $R_{n+1}=r(s_n,a,s_{n+1})$. Its goal is to learn a policy $\pi:\sS\times\sA\rightarrow [0,1]$ which maximizes the expected discounted return $G_n\triangleq \E_{\pi,p}[\sum_{k=0}^{\infty}\gamma^{k}R_{n+k+1}|s_n]$.

An option is a generalization of an action (also known as a skill or a sub-goal) \citep{sutton1999between}.
Formally, an option $o$ is the 3-tuple $\left\langle \sI,\pi_o,\beta\right\rangle $ where $\sI$ is an initiation set $\sI\subseteq\sS$ (the states at which the option can be invoked), $\pi_o:\sS\times\sA\rightarrow\left[0,1\right]$ is the policy of the option to be followed by the agent, and $\beta:\sS \to\left[0,1\right]$ is the termination condition. By following an option $o$ the agent chooses actions according to the policy of the option $\pi_o$ until the option is terminated according to the termination condition $\beta$.

\subsection{Diffusion Distance}
The diffusion distance is a notion of distance between two points in a high-dimensional data set \citep{coifman2006diffusion}, where the points are assumed to lie on a manifold. It is widely used in many data science applications, e.g., in  \citet{mahmoudi2009three,bronstein2011shape,lafon2006data,liu2009learning,lederman2018learning,van2018recovering}, since it captures well the geometric structure of the data. 
While the formulation of diffusion distance is typically general, here we describe it directly in the MDP setting.

Consider a graph $G=(\sS,\mathbb{E})$, where the finite set of states $\sS$ is the node set and the edge set $\mathbb{E} \subset \sS \times \sS$ consists of all possible transitions between states. Define a random walk on the graph with transition probability matrix $\mW$, defined by $\mW_{ij}=p(s_{t+1}=i|s_t=j)$. Let $\vp_t^{(l)}$ denote the vector of transition probabilities from state $l$ to all states in $t$ random walk steps defined by the $l$th column of $\mW^t$. Throughout the paper the convention is a column-vector representation. With the above preparation, the diffusion distance is defined by
\begin{equation*}
D_t\left(s,s'\right)\triangleq\Vert \vp_t^{(s)} - \vp_t^{(s')} \Vert,
\label{eq:diffusion_distance}
\end{equation*}
where $\Vert\cdot \Vert$ is the $L_2$ norm. In contrast to the standard Euclidean distance, the diffusion distance does not depend solely on two individual points, namely, $s$ and $s'$, but takes into account the structure of the entire data sets. See a prototypical demonstration in the supplementary material (SM). Broadly, in short distances it is closely related to the geodesic distance (shortest path) \citep{portegies2016embeddings} and in long distances it demonstrates high robustness to noise and outliers \cite{coifman2006diffusion}. 
For more details on the advantages of the diffusion distance and its efficient computation using the eigenvectors of the graph Laplacian see the SM.

\section{Diffusion Options}

In standard, mostly goal-oriented RL, one learns to map states to actions in order to achieve a desired task. In situations with uncertainty (e.g., model uncertainty, reward uncertainty, etc.) exploration is essential in order to reduce uncertainty, thereby improving future actions. Exploration often consists of aspects that are specific to a given task, and aspects that are generic to the domain. For example, in an environment with multiple rooms, one may wish to learn how to reach the door of each room, thereby facilitating learning in later situations where a specific task is given, say, reaching a specific room (or set of rooms). This may also be useful if additional rooms are later added. In both cases (task-based or task-free), options can greatly facilitate the speed of exploration by forming shortcuts \cite{eysenbach2018diversity}. In this work we present a manifold-based approach to developing generic options that can be later used across multiple task domains.

To encourage exploration, a useful set of options will lead the agent to distant regions, visiting states that the uninformed random walk will seldom lead to. 
To this end, we exploit the diffusion distance and show that the strength of diffusion distance in the realm of high dimensional data analysis enables us to devise structure-aware options that improve both learning and exploration.

\subsection{Algorithm}
\label{sec: Algorithm}

The derivation of the algorithm for option discovery is carried out in a setting consisting of discrete and deterministic domains with a finite number of states, where the transitions between states are known. This allows us to focus on the development of the representation of the domain with multiple spectral components and on the analysis based on the interface of spectral graph theory and the diffusion distance. Nevertheless, the primary interest is large scale domains, which are only partially known a-priori. In Section \ref{Sec:Sacling up}, by relying on previous work, we show how to accommodate such domains.
%Initially, we consider discrete and deterministic domains with a finite number of states, where the transitions between states are known to the agent. \rsm{}{This kind of setting allows for a theoretical analysis from which diffusion options are derived. In this paper, we focus on the theoretical analysis (and its demonstration), which introduces diffusion distance and manifold learning ideas to option discovery. The analysis results in new concepts: considering multiple spectral components, and incorporating them together instead of considering them apart. In large scale domains, we only have access to sampled state transitions. We show in section \ref{Sec:Sacling up} that diffusion options could also be extended to large scale domains, where the entire underlying graph is not known.} 

%TODO: we don't term our options "diffusion options" yet

The proposed algorithm for systematic option discovery consists of two stages.
The first stage involves graph construction.
Let $G$ be a graph whose node set is the finite set of states $\sS$. Let $\mM\in\mathbb{R}^{|\mathbb{S}|\times|\mathbb{S}|}$ be the symmetric adjacency matrix of the graph, prescribing the possible transitions between states, namely $\mM_{s,s'}=1$ if a transition from state $s$ to state $s'$ is possible, and $\mM_{s,s'}=0$ otherwise. Based on $\mM$, define a non-symmetric lazy random walk matrix $\mW\triangleq \frac{1}{2}(\mI+\mM\mD^{-1})$, where the degree matrix $\mD$ is a diagonal matrix whose diagonal elements equal the sum of rows of $\mM$. Applying eigenvalue decomposition to $\mW$ yields two sets of left and right eigenvectors, denoted by $\{\phi_i\}$ and $\{\Tilde{\phi}_i\}$, respectively, and a set of real eigenvalues $\{\omega_i\}$. The $s$th component of $\phi_i$ is denoted by $\phi_i(s)$. 

The second stage of the algorithm relies on the following score function, $f_t:\sS\rightarrow\sR$, defined on the set of states $\mathbb{S}$, and assigns a score to each state $s \in \mathbb{S}$
\begin{equation}
    f_t(s)\triangleq \Vert\sum_{i\ge{2}}\omega_i^t\phi_i\left(s\right)\Tilde{\phi}_i\Vert^2,
    \label{eq:f_tDefinition}
\end{equation}
where $t>0$ is a scale parameter representing the diffusion time. 
By construction, $f_t(s)$ consists of the full spectrum of $\mW$, including both low and high frequencies, in contrast to common practice.
As we show in Proposition \ref{prop:MainPropAvDiff}, $f_t(s)$ is directly related to the average diffusion distance between state $s$ and all other states, making it a promising candidate for an option discovery criterion, as discussed below.

After computing $f_t(s)$, the states at which it attains a local maximum are extracted. We term these states option goal states, and denote them by $\{s_o^{(i)}\}$, where the index $i$ ranges between 1 and the number of local maxima. Each such state is associated with an option, which leads the agent from its current state to the option goal state.  
The options can start at any state ($\sI=\sS$), and terminate deterministically once reaching its option goal state, i.e. for option $i$, $\beta_i\left(s_o^{(i)}\right)=1$, and $\beta_i\left(s\right)=0\,\forall s\not=s_o^{(i)}$. In other words, once the agent chooses to act according to an option, it moves to $s_o$ via the shortest path from its current position.
We note that the scale parameter $t$ indirectly controls the number of options; since $0<\omega_i \le 1$ \citep{chung1997spectral}, the multiplication by $\omega_i^t$ in (\ref{eq:f_tDefinition}) makes $f_t(s)$ smoother as $t$ increases, analogously to a low pass filter effect. 
In addition, many eigenvalues are often negligible, and therefore, accurate reconstruction of $f_t(s)$ in (\ref{eq:f_tDefinition}) typically does not require all the spectral components.

The proposed algorithm for option discovery appears in Algorithm \ref{Alg: OptionsDiscovery}.
We term the discovered options \emph{diffusion options} because they are built from the eigenvalue decomposition of a discrete diffusion process, i.e., the lazy random walk on the graph. In addition, in Section \ref{subsec:analysis}, we show a tight relation to the diffusion distance.
Algorithm \ref{Alg: OptionsDiscovery} exhibits several advantages. First, the algorithm prescribes a systematic way to derive options which are not associated with any particular task or reward. Second, we empirically demonstrate the acceleration of the learning process and more efficient exploration in prototypical domains compared to competing methods for option discovery. Third, the computationally heavy part is performed only once and in advance. Fourth, the scale parameter $t$ enables to control the number of options and facilitates multiscale option discovery. 

We remark that the eigenvalue decomposition of $\mW$ used for the construction of $f_t(s)$ is related to the eigenvalue decomposition of the normalized graph Laplacian $\mN$, which traditionally forms the spectral decomposition of a graph. See the SM for details.

\begin{algorithm}
    \caption{Diffusion Options}
    \label{Alg: OptionsDiscovery}
    
     \textbf{Input:} Adjacency matrix $\mM$ and scale parameter $t>0$
     
     \textbf{Output:} $K$ options with policies $\{\pi_o^{(i)}\}_{i=1}^{K}$ 

    \begin{algorithmic}[1] 
     \STATE Compute the degree matrix $\mD$ from $\mM$
     \STATE Compute the random walk matrix \newline
     $\mW=\frac{1}{2}(\mI-\mM\mD^{-1})$   
     \STATE Apply EVD to $\mW$ and obtain \{$\phi_i$\}, \{$\Tilde{\phi}_i$\} and $\{\omega_i\}$
     \STATE Construct $f_t(s)= \Vert\sum_{i\ge{2}}\omega_i^t\phi_i\left(s\right)\Tilde{\phi}_i\Vert^2$ 
    
    \STATE Find the states $\{s_o^{(i)}\}_{i=1}^{K}$ of the local maxima of $f_t(\cdot)$ 

    \FOR{$i \in \{1,\ldots,K\}$}
     \STATE Build an option with policy $\pi_o^{(i)}$ s.t. it leads to $s_o^{(i)}$ 
     \ENDFOR
\end{algorithmic}
\end{algorithm}

\subsection{Extension to Large Scale Domains}
\label{Sec:Sacling up}
The exposition thus far focused on domains, whose full transition matrix is at hand when learning the representation of the domain. 
Suppose now that the considered set of states $\sS$ is only a subset of the entire set of states. 
% TODO: consider removing reference to algorithm1
The extension of diffusion options discovered by Algorithm \ref{Alg: OptionsDiscovery} to unseen states $s \notin \sS$ requires the extension of $f_t(s)$ and the extension of the option policies. Since the option policies can be trained off-policy as shown by \citet{jinnai2020exploration}, here we focus on extending $f_t(s).$

In the SM, we show that the extension of $f_t(s)$ to unseen states $s \notin \sS$ involves the extension of the eigenvectors $\phi$ and $\tilde{\phi}$ taking part in the construction of $f_t(s)$ in (\ref{eq:f_tDefinition}).
Since the eigenvectors admit a particular algebraic structure, their extension is naturally regularized, and therefore, often more accurate than a generic function extension. This fact was recently exploited by \citet{wu2018laplacian}, who developed a method that was later demonstrated by \citet{jinnai2020exploration}, to compute the eigenvectors of the graph Laplacian in large scale domains. Another approach for extending the eigenvectors was proposed by \citet{machado2017eigenoption} using deep successor representation. We note that due to the low pass filter effect in (\ref{eq:f_tDefinition}) not all the eigenvectors need to be extended. Additionally, only the locations of the local maxima of $f_t(s)$ are used in Algorithm \ref{Alg: OptionsDiscovery}, rather than all its values, thus we can extend a sufficient number of eigenvectors, so that the same local maxima are attained as in the construction with all the spectral components.

%In theory, using the full spectrum leads to Proposition \ref{prop:MainPropAvDiff} and Proposition \ref{prop:FuncStationaryDistribution} in section \ref{subsec:analysis} (TODO: not yet introduced). However, when the eigenvalue is small and the computation of its corresponding eigenvectors is typically less accurate, its influence on $f_t(s)$ is also small, due to the low pass filter effect (discussed in section \ref{sec: Algorithm}). Based on our empirical tests, using only the dominant 10-20 eigenvectors leads to $f_t$ with the same local maxima, resulting in the same options. We emphasize that using fewer is insufficient and does not capture well the geometry of the domain.

% TODO: we now don't include references to Chui and Mishne for extending eigenvectors with NN...

% TODO: I think it is still not clear how to find those states in large scale domains
After the score function $f_t(s)$ is approximated, extracting its local maxima requires not only going over all its entries, but also considering their connectivity. This additional complexity is negligible when the underlying graph is sparsely connected. Importantly, the more connected the graph is, the less significant the options are (e.g., as demonstrated by \citet{jinnai2019discovering}), and therefore, in the context of this paper, only sparsely connected graphs are of interest.

\subsection{Analysis}
\label{subsec:analysis}
We start the analysis with our main result relating $f_t(s)$ to the diffusion distance. The proof is provided in the SM.

\begin{prop}
\label{prop:MainPropAvDiff}
The function $f_t:\sS \rightarrow \R$ defined as  $f_t(s)\triangleq \Vert\sum_{i\ge{2}}\omega_i^t\phi_i\left(s\right)\Tilde{\phi}_i\Vert^2$ is equal to the mean squared diffusion distance between state $s$ and all other states, up to a constant independent of $s$, namely
\begin{equation}
    f_t(s)=\left< D_t^2\left(s,s'\right)\right>_{s'\in\sS}+\text{const},
\label{eq:f_tEQMeanDiffDist}
\end{equation}
where $\langle g(x) \rangle _{x \in \sX}$ represents the average on $\sX$:
\begin{equation*}
    \langle g(x) \rangle_{x\in\sX}\triangleq \frac{1}{|\sX|}\sum_{s\in\sX}g(x).
\end{equation*}
\end{prop}

An immediate consequence of Proposition \ref{prop:MainPropAvDiff} is that 
\begin{equation*}
\mathrm{max}_s f_t\left(s\right)=\mathrm{max}_s\left<D^2_t\left(s,s'\right)\right>_{s'\in\sS},    
\end{equation*}
implying that the option goal states, $\{s_o^{(i)}\}$, are the farthest states from all other states in terms of average squared diffusion distance.
Broadly, moving to such far states encourages exploration as the agent systematically travels through the largest number of states without, for example, the repetitions involved in the uninformed random walk. Additionally, by reaching different option goal states, the agent reaches different and distant regions of the domain, which also benefits exploration.
The particular notion of diffusion distance efficiently captures the geometry of the domain and demonstrates important advantages over the Euclidean and even the geodesic distances. See the SM for an illustrative example. The averaging operation $\langle \cdot \rangle$ incorporates the fact that the options are not related to a specific task, and therefore, the start state, the goal state, and the states at which the options are invoked, are all unknown a-priori.

Empirically we will demonstrate that the diffusion distance is related to the domain difficulty (see Section \ref{Sec:DiffDistDomComplx}). The larger the average pairwise diffusion distance is, the more difficult the domain is. As a result, when the agent follows options leading to distant states in terms of the diffusion distance, in effect, it reduces the domain difficulty.
In addition, we demonstrate that such goal states are typically ``special'' states such as corners of rooms or bottleneck states such as doors (see Fig. \ref{fig:4RoomsDiffOpt_tEq4} and Section \ref{Sec:ExperimentalResults}).

Proposition \ref{prop:FuncStationaryDistribution} offers an alternative perspective on $f_t(s)$, relating it to the stationary distribution of the graph, denoted by $\bm{\pi}_0$. The proof is in the SM.

\begin{prop}
\label{prop:FuncStationaryDistribution}
$f_t(s)$ can be recast as
\begin{equation*}
    f_t(s)=\Vert \vp_t^{(s)}-\bm{\pi}_0\Vert^2,
\end{equation*}
where $\bm{\pi}_0$ is the stationary distribution of the lazy random walk $\mW$ on the graph $G$. 
In addition, $f_t(s)$ is bounded from above by  
\begin{equation*}
  f_t\left(s\right) \le
    \omega_2^{2t}\left(\frac{1}{\bm{\pi}_0\left(s\right)}-1\right).  
\end{equation*}
\end{prop}

The first part of Proposition \ref{prop:FuncStationaryDistribution} relates $f_t(s)$ to the difference between the transition probability from state $s$ and the stationary distribution. As $t$ grows to infinity, the transition probability approaches the stationary distribution. For a fixed $t$, the states at which $f_t(s)$ gets a maximum value are the states that their transition probability differ the most from the stationary distribution.  

States $s$ for which $\pi_0 (s)$ is small are states that are least visited by an agent following a standard random walk. Arguably, these are exactly the states the agent should visit, for example by following options, to improve exploration. Indeed, we observe that the upper bound in Proposition \ref{prop:FuncStationaryDistribution} implies that these states allow for large $f_t(s)$ values.
We further discuss the relation between $f_t(s)$ and $\pi_0$ in a multi-dimensional grid domain in the SM.

Establishing the relation of $f_t(s)$ to the stationary distribution is important by itself because the stationary distribution is a central component in many applications and algorithms. Perhaps the most notable are PageRank \cite{page1999pagerank} and its variants \citep{kleinberg1999authoritative}, where the purpose is to discover important web pages that are highly connected and therefore can be considered as network hubs. In the exploration-exploitation terminology, one could claim that PageRank favors exploitation by identifying central pages. Conversely, the diffusion options lead the agent toward states that are least connected (with small stationary distribution values), and therefore, they encourage exploration.

We end this section with two remarks.
First, the upper bound in Proposition \ref{prop:FuncStationaryDistribution} generalizes a known bound on the convergence of the transition probability, starting from node $a$ in a graph, to the stationary distribution at node $b$ \citep{DanielSpielmanLectures},  
\begin{equation*}
  \left|p_t\left(b\right)-\pi_0\left(b\right)\right|\leq
\sqrt{\cfrac{d\left(b\right)}{d\left(a\right)}}\omega_2^t,  
\end{equation*}
where $d(a)$ and $d(b)$ are the degrees of nodes $a$ and $b$, respectively.

Second, combining Proposition \ref{prop:MainPropAvDiff} and Proposition \ref{prop:FuncStationaryDistribution} relates the diffusion distance to the distance from the stationary distribution of a random walk. This relation may have consequences in a broader context, when either the diffusion distance or the stationary distribution are used.

\subsection{Extension to Stochastic Domains}
\label{sec:Ext2StochDomainAnalys}

In the deterministic setting we considered thus far, we assumed that an action definitively leads the agent to a particular state, i.e., given an action $a$ and a state $s$ the probability $p(s'|s,a)$ is concentrated at a single state. 

Alternatively, one could consider a setting, where the domain is stochastic, and its stochasticity introduces uncertainty and decouples the action from the transition, namely, $p(s'|s,a)$ can be supported on more than one state. As a result, the agent following a random walk experiences a different number of transitions between states. 
The corresponding transition probability matrix leads to a non-symmetric normalized graph Laplacian $\mN$.
This poses a challenge since the eigenvalue decomposition of $\mN$ is not guaranteed to be real, and therefore, the construction of $f_t(s)$ in (\ref{eq:f_tDefinition}) needs a modification.
Note that other settings could lead to a non-symmetric Laplacian as well.

Here, we propose a remedy to support such cases.
Our solution follows the work presented by \citet{mhaskar2018unified}, which is based on the polar decomposition. Concretely, consider the polar decomposition of $\mN=\mR\mU$, where $\mR$ is a positive semi-definite matrix and $\mU$ is a unitary matrix. Since $\mR$ is uniquely determined, the spectral analysis applied to $\mN$ in the deterministic case can be applied to $\mR$ in a similar manner.
As observed by \citet{mhaskar2018unified}, there exist efficient algorithms for computing $\mR$ \citep{nakatsukasa2010optimizing}. 
Accordingly, the required modification applied to the option discovery in Algorithm \ref{Alg: OptionsDiscovery} is minimal. After the computation  of $\mN$, its polar decomposition is computed. Then, the eigenvalue decomposition of the positive part $\mR$ is used for the construction of $f_t(s)$. 
See the SM for the modified algorithm. 
In Section \ref{sec:StochasticDomainResult}, we demonstrate its performance.

\section{Experimental Results}
\label{Sec:ExperimentalResults}

We demonstrate empirically that the diffusion options are generic and useful, allowing for improvement in both learning unknown tasks and in exploring domains efficiently. 
Particularly, using Q learning \citep{watkins1992q}, we show that equipped with the diffusion options, which are computed in a reward-free domain, the agent is able to learn tasks that are unknown a-priori faster and to explore a domain more effectively.
In addition, we demonstrate the relation between the diffusion options and the stationary distribution.

We focus on three domains: a Ring domain, which is the $2$D manifold of the placement of a 2-joint robotic arm \citep{verma2008mathematical}, a Maze domain \citep{wu2018laplacian}, and a 4Rooms domain \citep{sutton1999between}. The set of actions are: left, right, up and down. In every domain, we pre-define a single start state and a set of goal states.

The agent performs several trials, where each trial is associated with a different goal state from the set of goal states. In each trial, the agent starts at the same start state and is assigned with the task of reaching the trial goal state.
We implement Q learning \citep{watkins1992q} with $\alpha=0.1$ and $\gamma=0.9$ for $400$ episodes, containing $100$ steps each. 
The agent follows the Q function at states for which it exists, and otherwise chooses a primitive action or an option with equal probability. In case the agent does not reach the goal state after $100$ steps, a default value of $101$ is set for the number of steps.

Since options typically consist of multiple steps, for a fair comparison, we take them into account in the total steps count at each episode. Note that this might lead to terminating an option without reaching its option goal state in case the episode reaches $100$ steps. 

We compare the diffusion options with the eigenoptions presented by \citet{machado2017laplacian} and with the cover options from \citet{jinnai2019discovering}. 
As a baseline, we also show results for a random walk consisting of only primitive actions without options. In the SM, we include comparison to random options as well.

We evaluate the performance using three objective measures. The first measure is the standard learning convergence. We compute the average number of steps to a goal over all learning trials (goal states), where each trial consists of $30$ Monte Carlo iterations. The average number of steps is presented as a function of the learning episode.
Second, we present the average number of visitations at each state during learning (over all episodes and goal states).
Third, to evaluate the exploration efficiency, we compute the number of steps between every two states, following \citet{machado2017laplacian}. 

The main hyperparameter of the algorithm is $t$. In our implementation, we set $t=4$. 
Our empirical study shows that different values of $t$ lead to similar results.
For results using other $t$ values and for a further discussion on the choice of $t$, see the SM.

\subsection{Diffusion Options Generation}

\begin{figure}[t]
\centering
    \subfigure[]{\label{fig:RingStatDistribution}\includegraphics[width=.32\columnwidth]{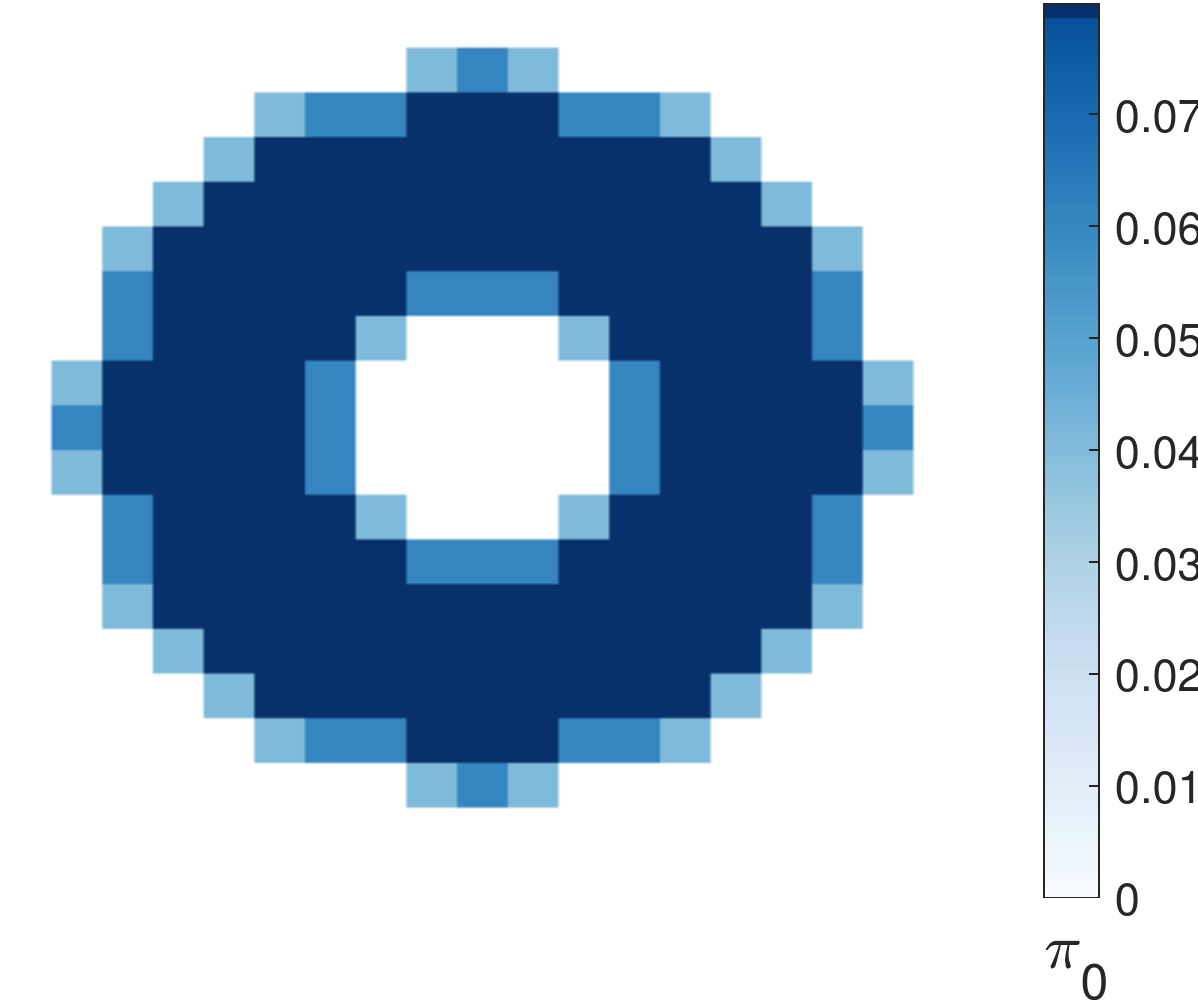}}
    \subfigure[]{\label{fig:RingDiffOpt_tEq4}\includegraphics[width=.32\columnwidth]{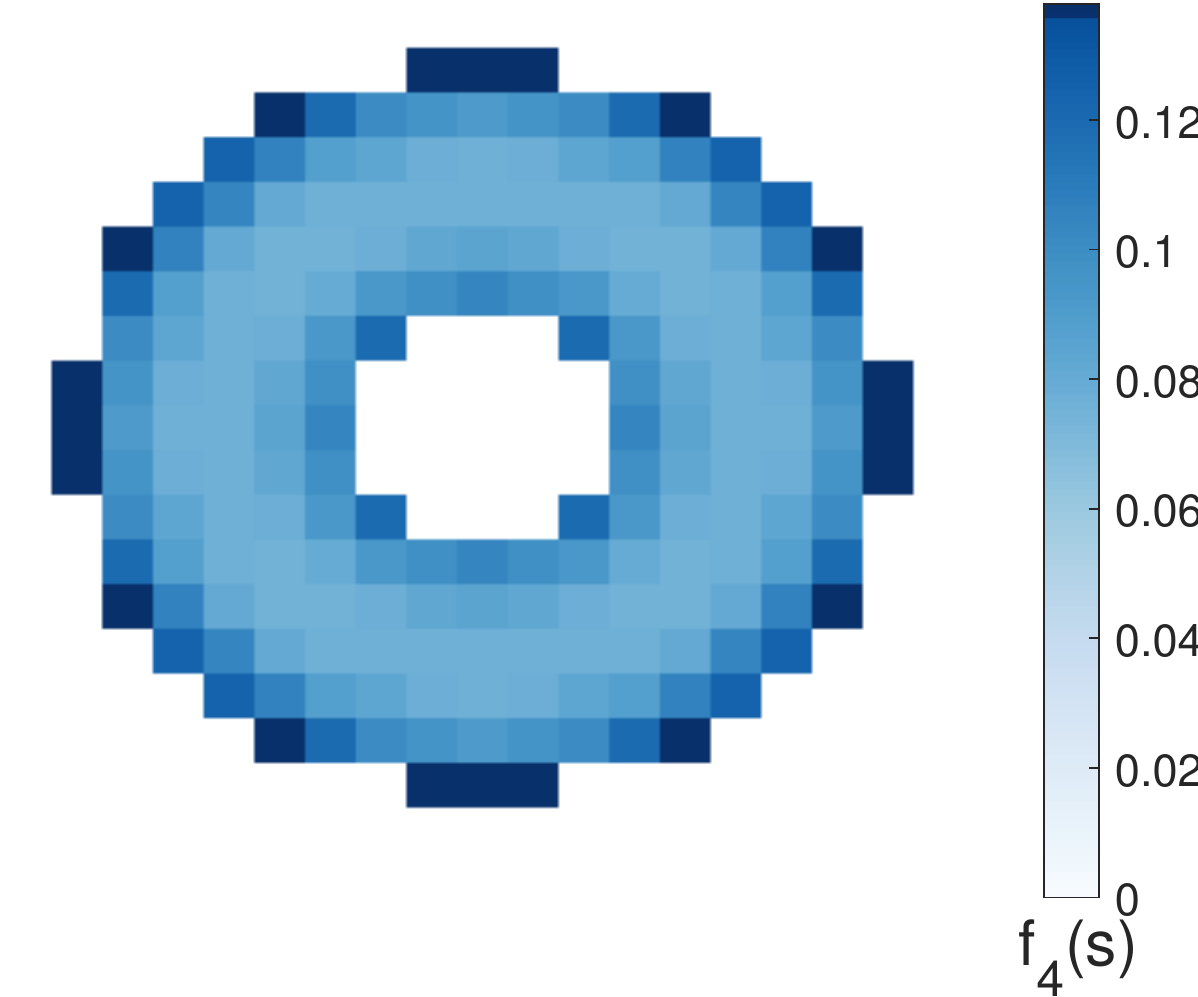}}
    \subfigure[]{\label{fig:RingDiffOpt_tEq13}\includegraphics[width=.32\columnwidth]{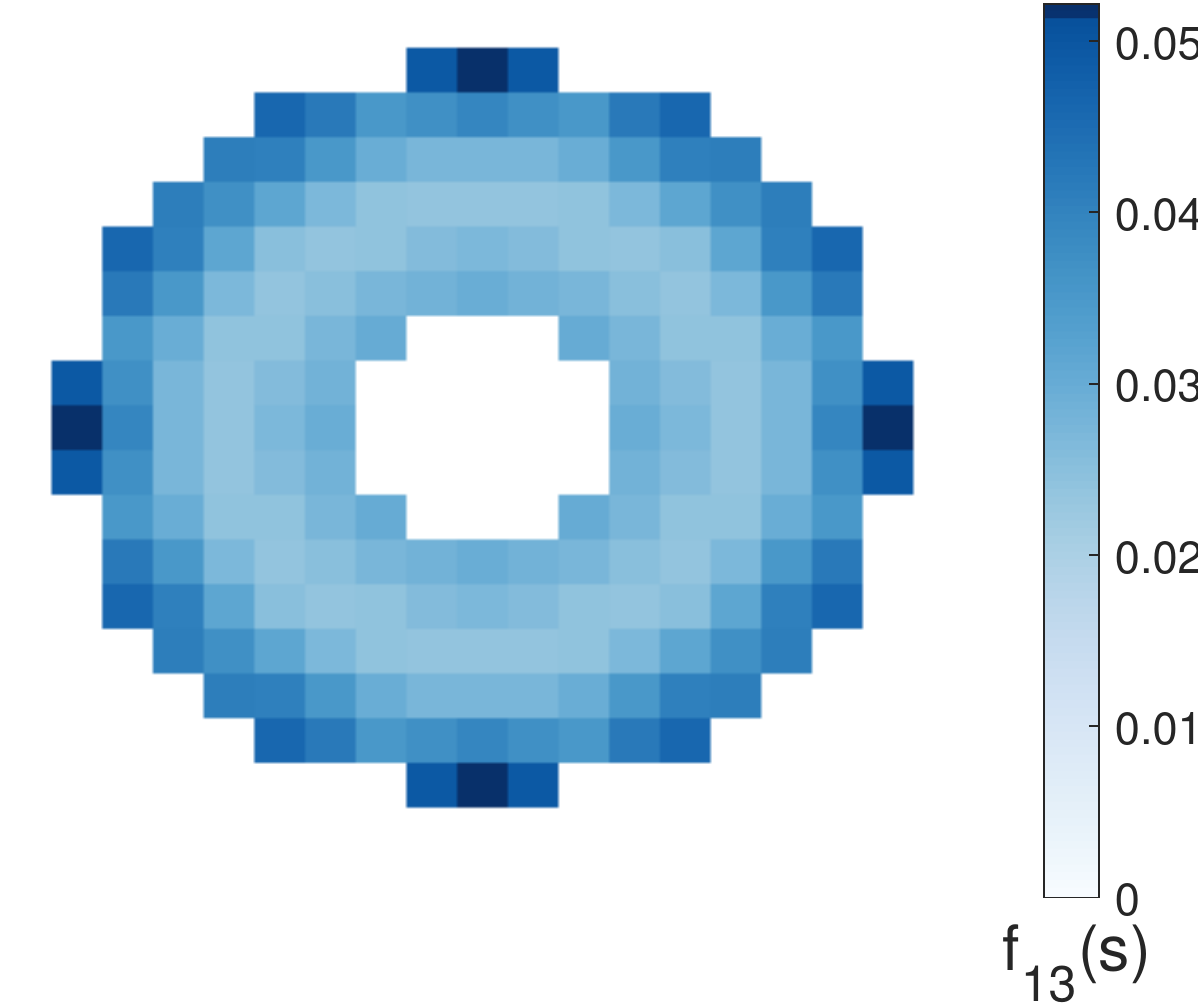}}
    \subfigure[]{\label{fig:MazeStatDistribution}\includegraphics[width=.32\columnwidth]{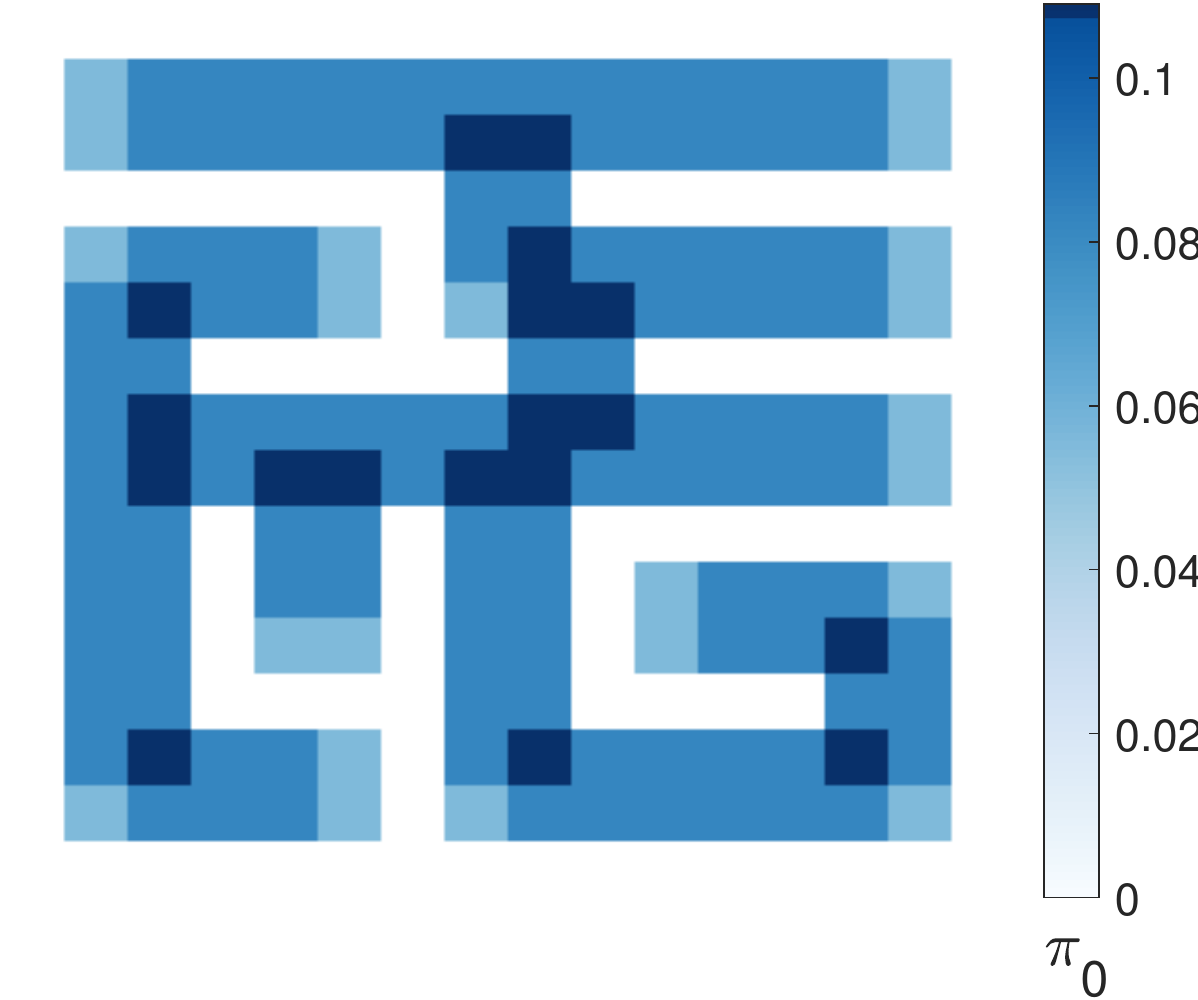}}
    \subfigure[]{\label{fig:MazeDiffOpt_tEq4}\includegraphics[width=.32\columnwidth]{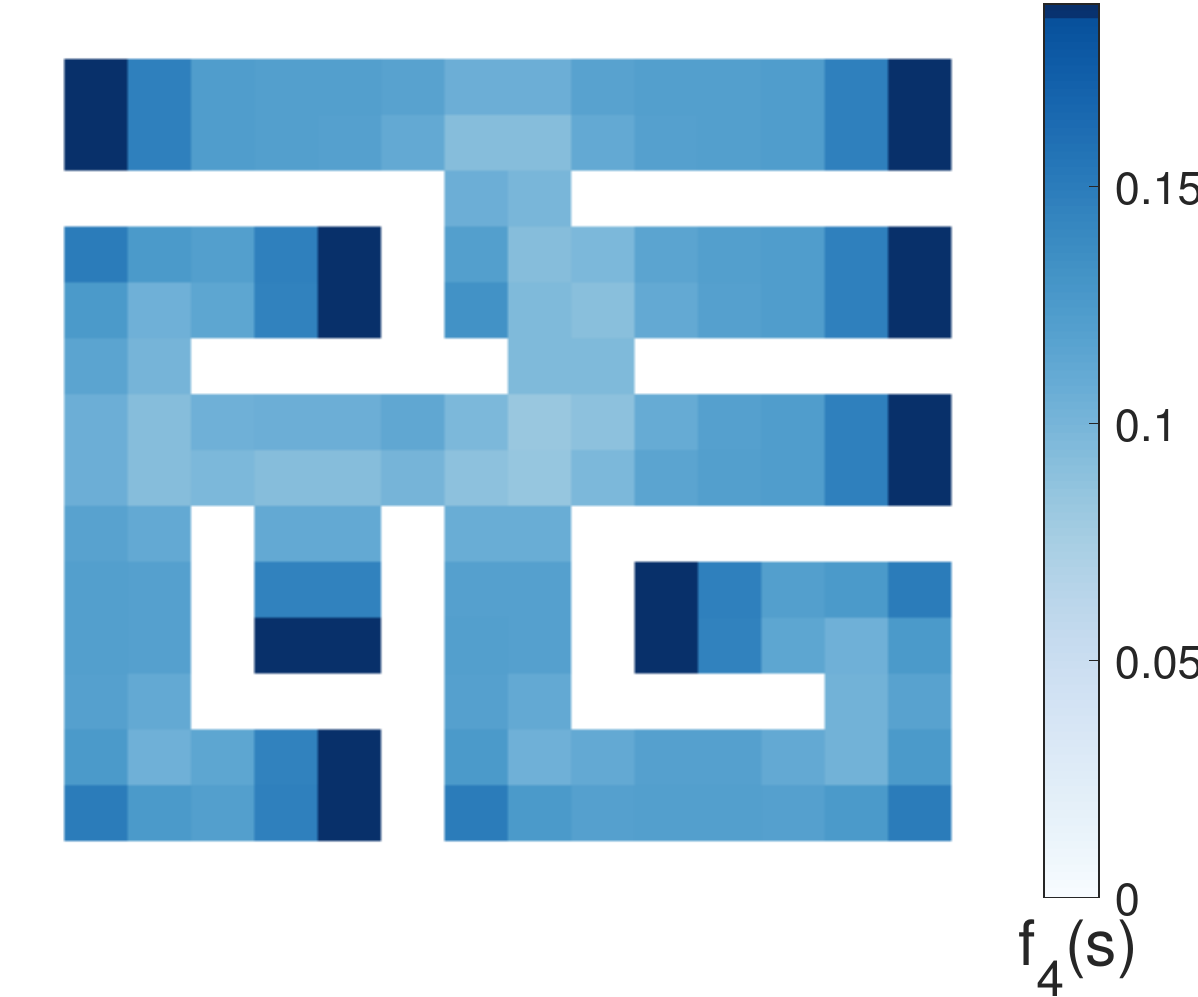}}
    \subfigure[]{\label{fig:MazeDiffOpt_tEq13}\includegraphics[width=.32\columnwidth]{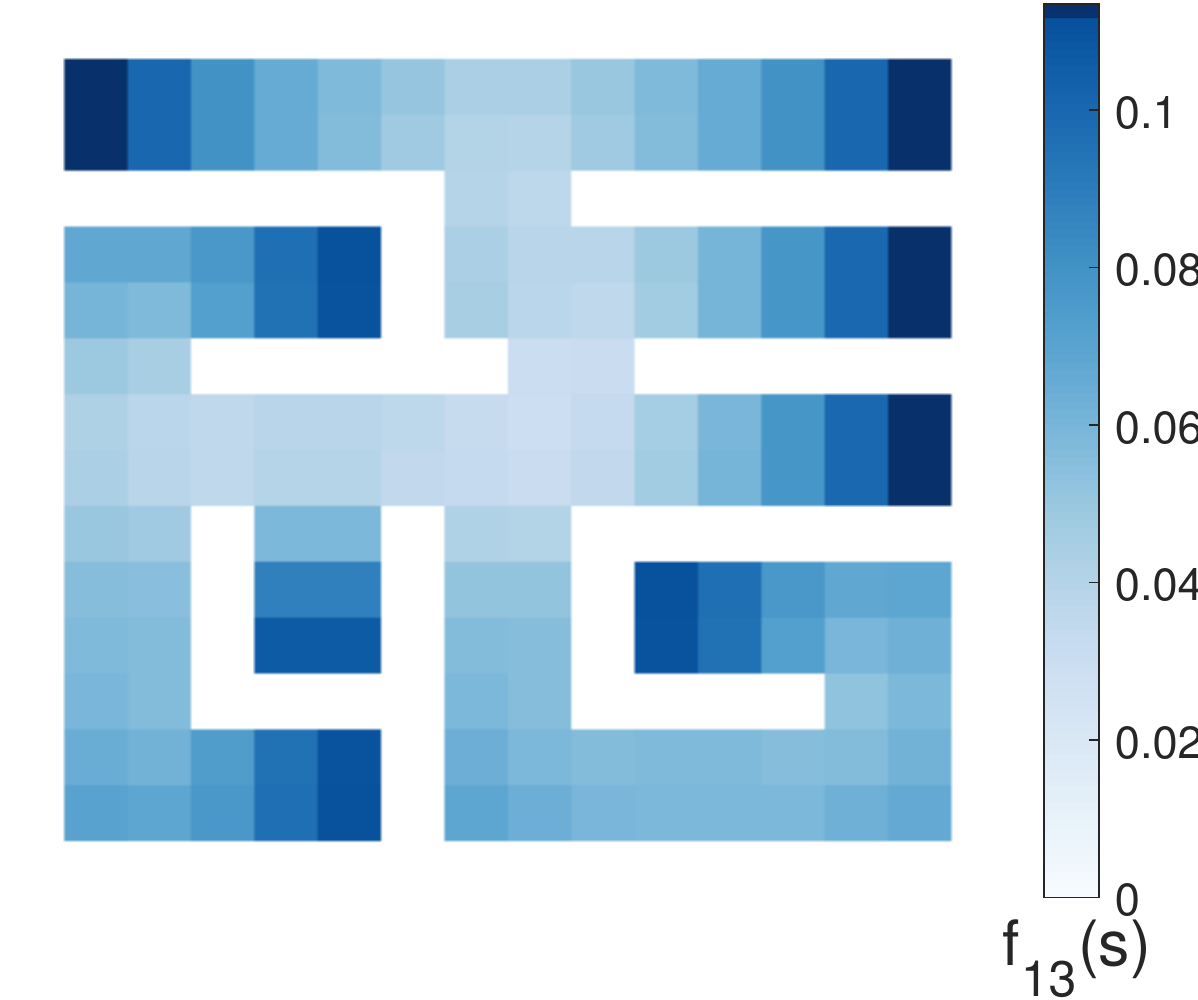}}
    \subfigure[]{\label{fig:4RoomsStatDistribution}\includegraphics[width=.32\columnwidth]{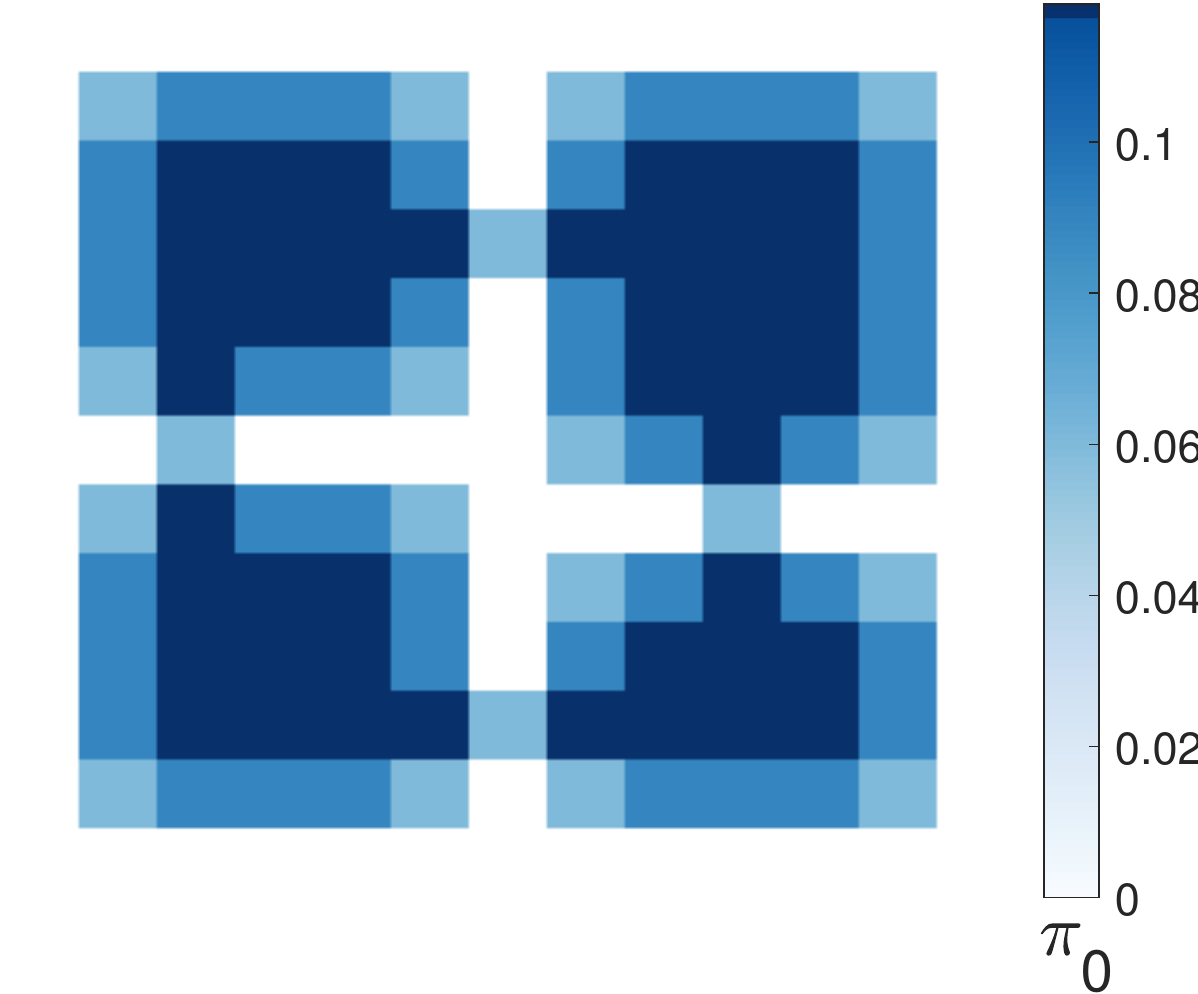}}
    \subfigure[]{\label{fig:4RoomsDiffOpt_tEq4}\includegraphics[width=.32\columnwidth]{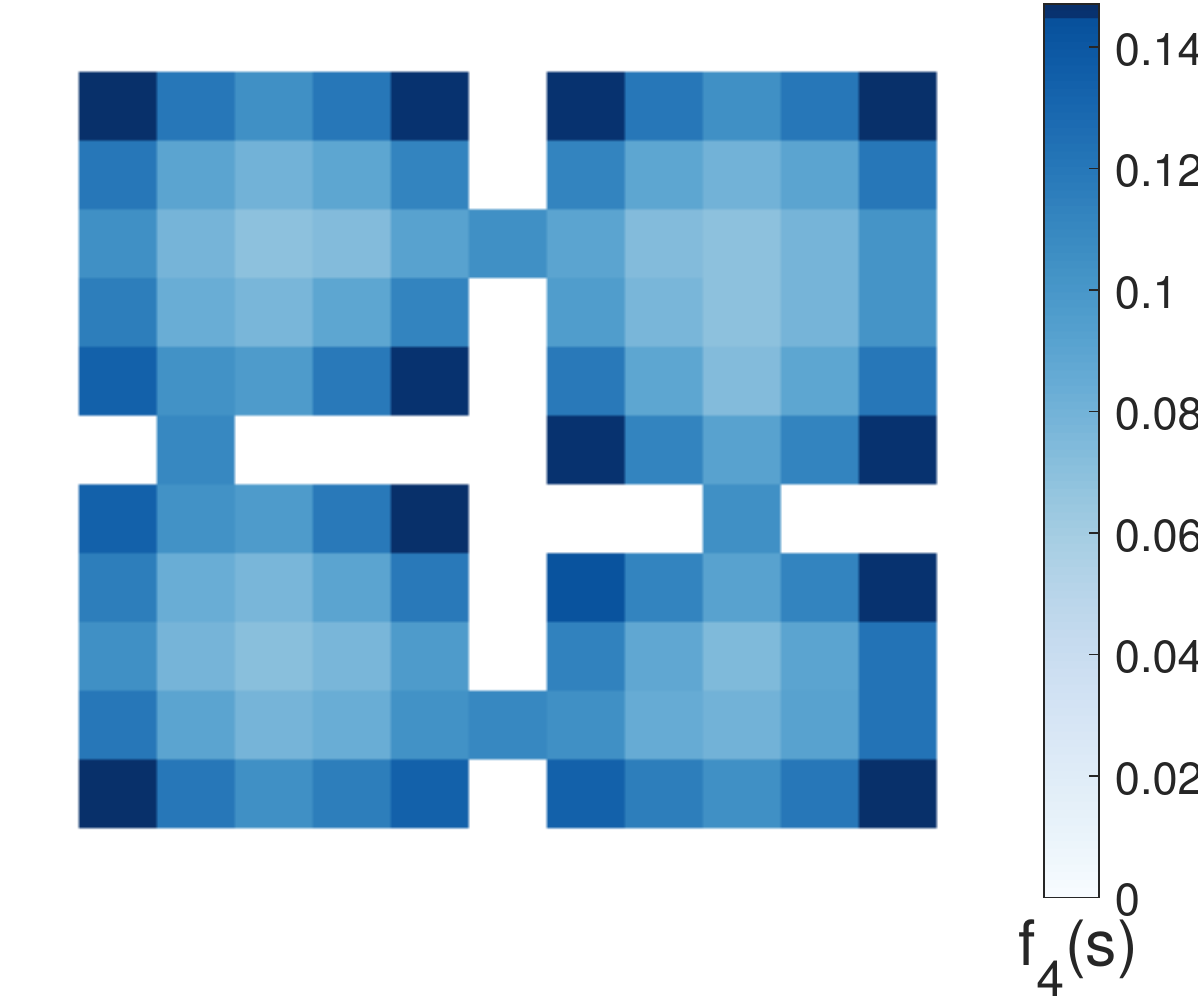}}
    \subfigure[]{\label{fig:4RoomsDiffOpt_tEq13}\includegraphics[width=.32\columnwidth]{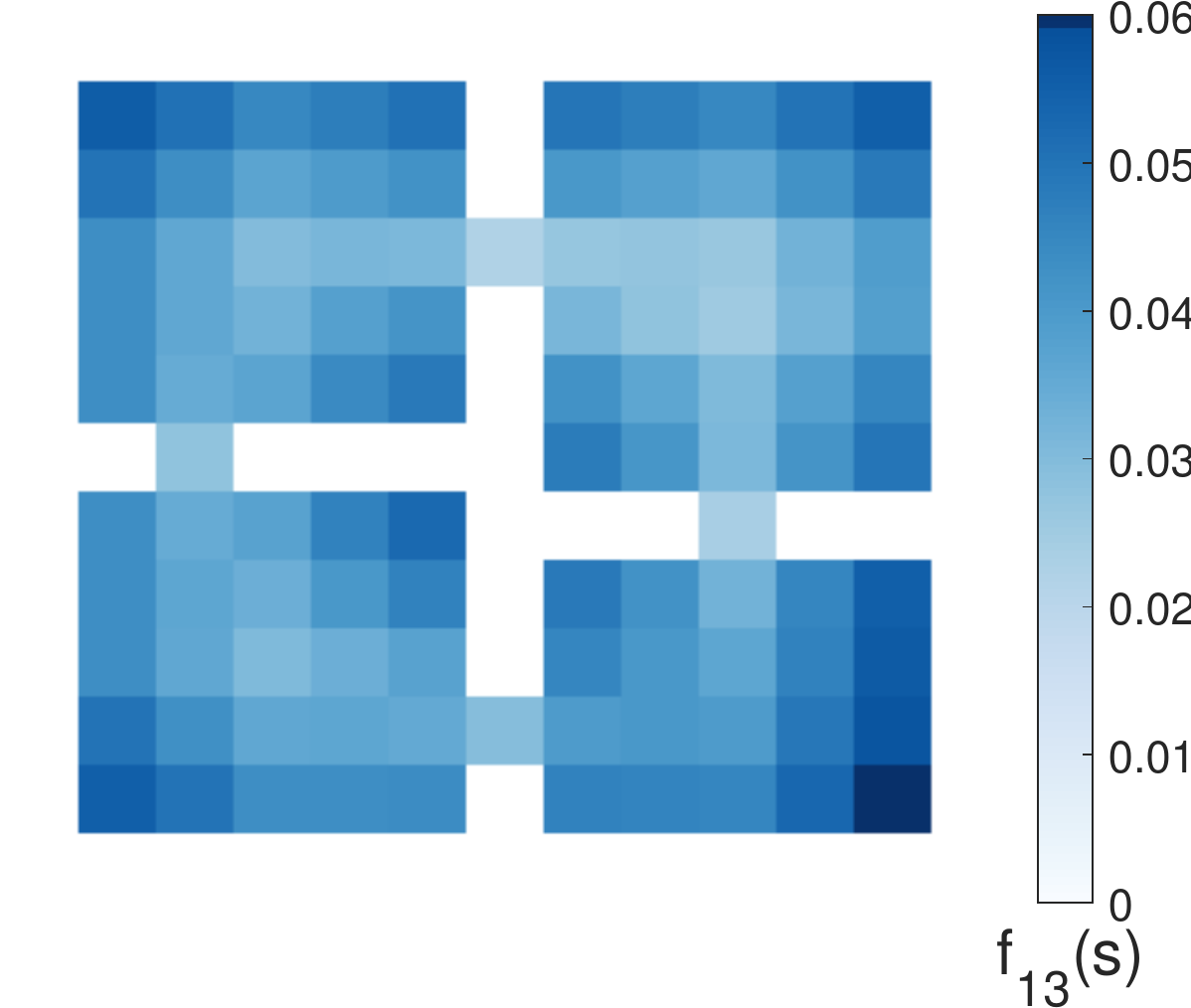}}
    \caption{The domains colored according to (a,d,g) the stationary distribution $\pi_0$, (b,e,h) the options generating function $f_4(s)$, and (c,f,i) the options generating function $f_{13}(s)$.}
   \label{fig:f_t}
\end{figure}

In Fig. \ref{fig:f_t}, we plot the options generating function $f_t(s)$ for two values of $t$ as well as the stationary distribution.
First, we observe the low pass filter effect obtained by increasing the scale parameter $t$. Particularly, we see that $f_{13}(s)$ is smoother, containing fewer peaks, than $f_4(s)$. Based on our empirical tests, using only the dominant 10-20 eigenvectors leads to $f_t$ with the same local maxima, resulting in the same options. As discussed in Section \ref{Sec:Sacling up}, this facilitates the extension to large scale domains. We emphasize that using fewer eigenvectors is insufficient and does not capture well the geometry of the domain.
Second, we observe that the minima of the stationary distribution coincide with the local maxima of $f_t(s)$ for some cases, in accordance with Proposition \ref{prop:FuncStationaryDistribution}. For example, note the corners of the rooms and the doors in the 4Rooms domain (Figs. \ref{fig:4RoomsStatDistribution} and \ref{fig:4RoomsDiffOpt_tEq4}). Nevertheless, we observe that the local minima of the stationary distribution might also capture irrelevant states in evolved domains. For example, in the Maze domain, in contrast to the stationary distribution, $f_t(s)$  captures the end of the corridors \emph{only} (see Figs. \ref{fig:MazeStatDistribution} and  \ref{fig:MazeDiffOpt_tEq13}), which are important for efficient exploration and learning in this domain. 

\subsection{Exploration and Learning}

\begin{figure*}[t]
\centering
    \subfigure[]{\label{fig:RingStartGoal}\includegraphics[width=.19\textwidth]{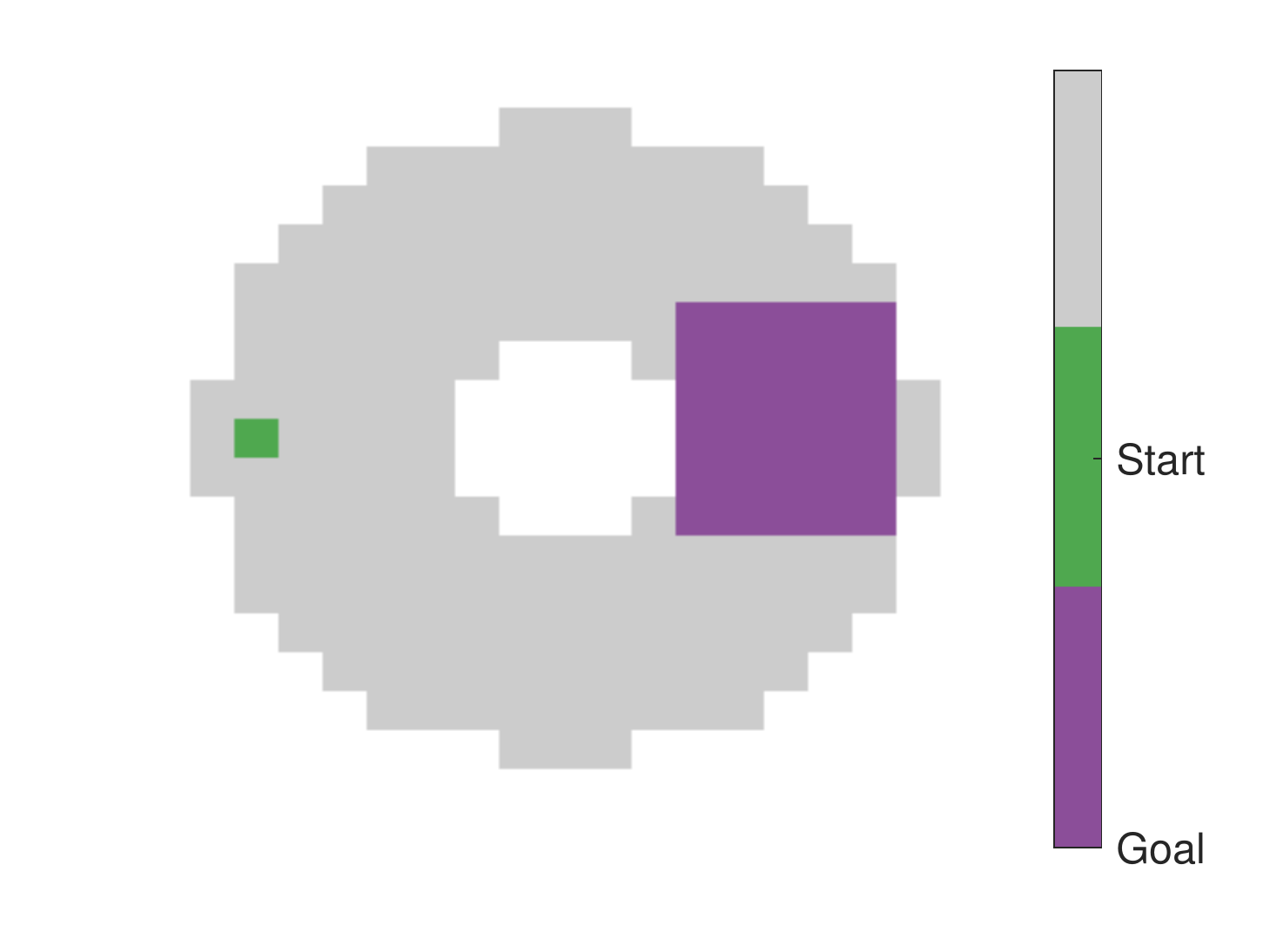}}
    \subfigure[]{\label{fig:RingOurVisitation}\includegraphics[width=.16\textwidth]{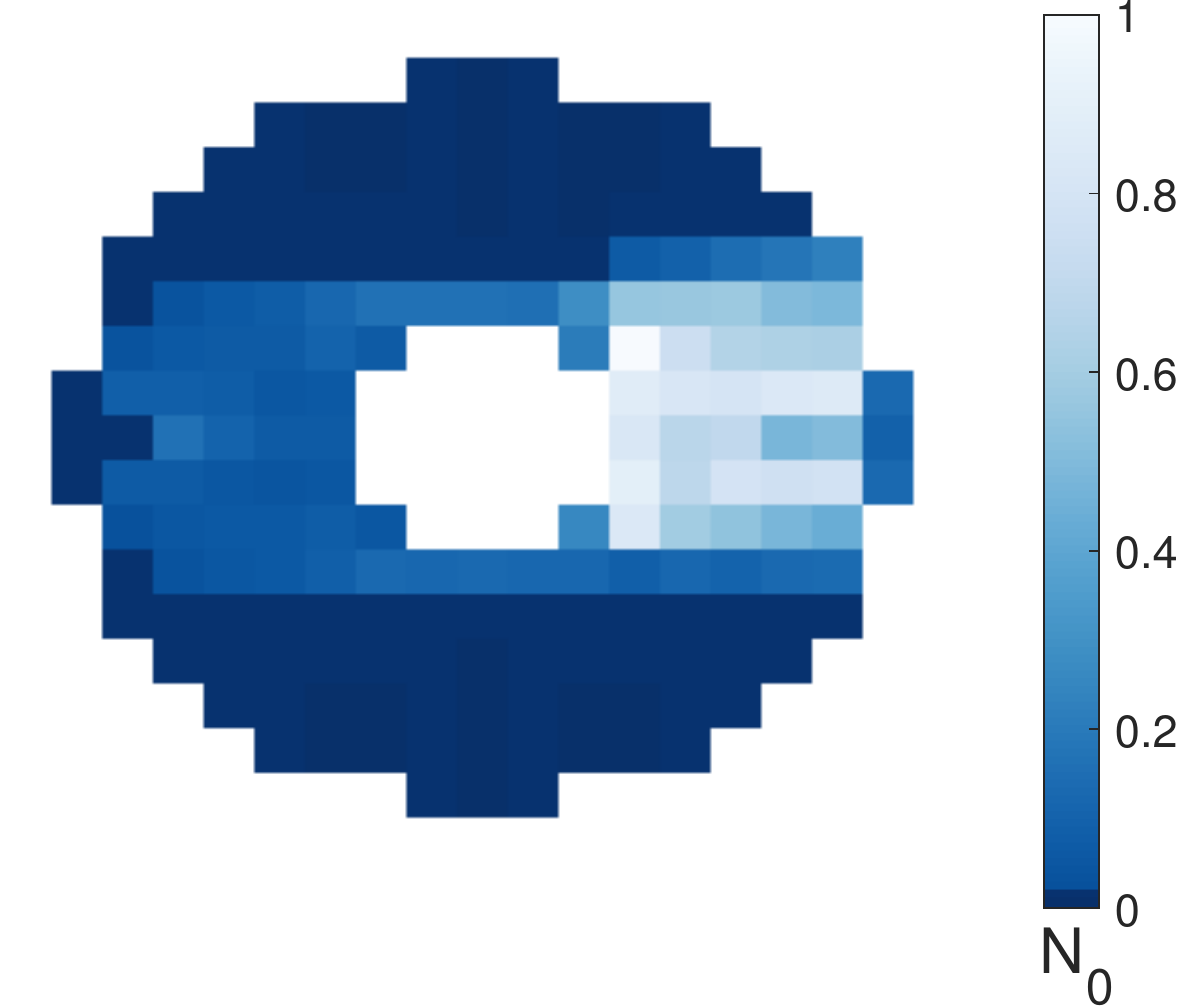}}
    \subfigure[]{\label{fig:RingLaplacianVisitation}\includegraphics[width=.16\textwidth]{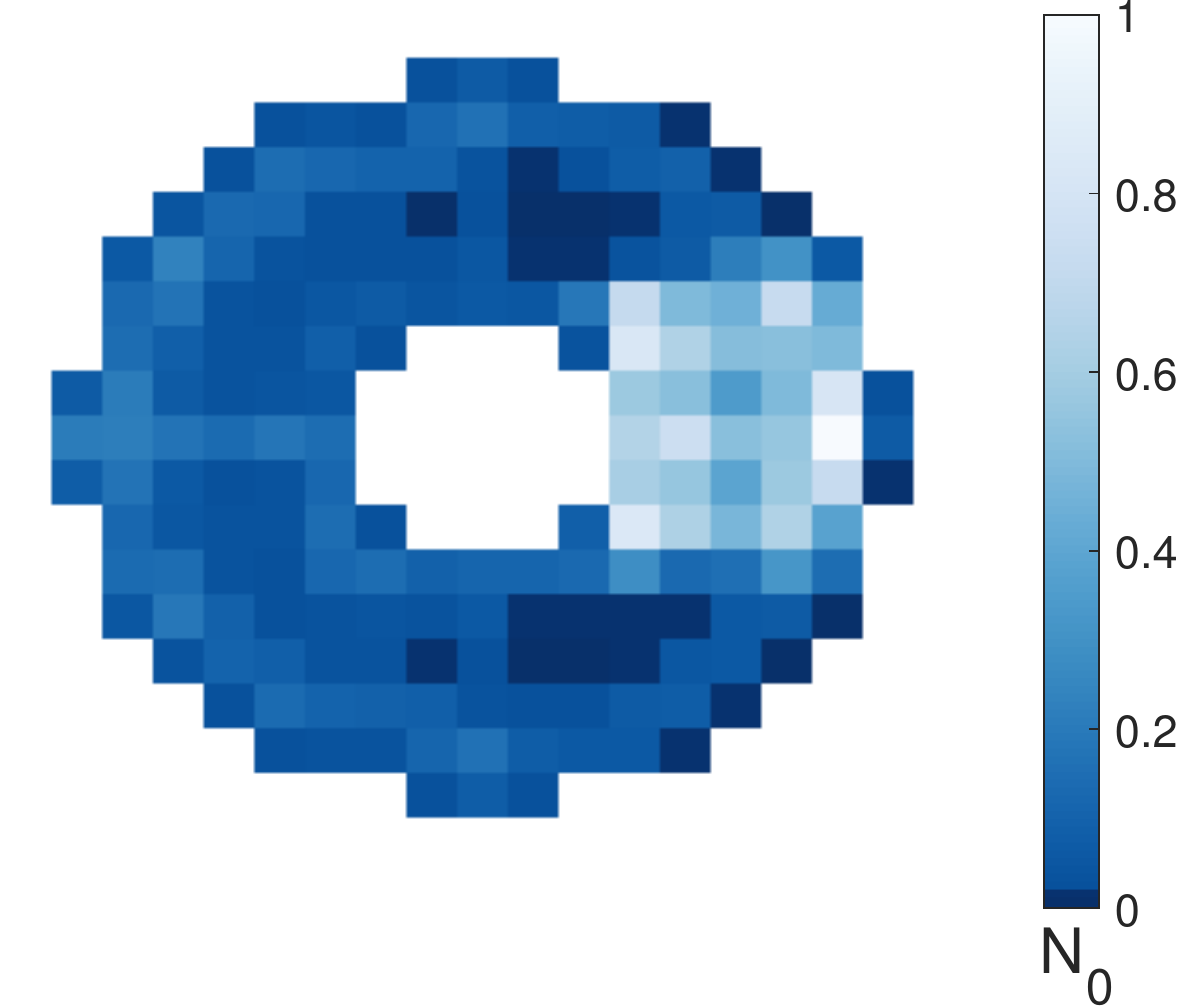}}
    \subfigure[]{\label{fig:RingRWVisitation}\includegraphics[width=.16\textwidth]{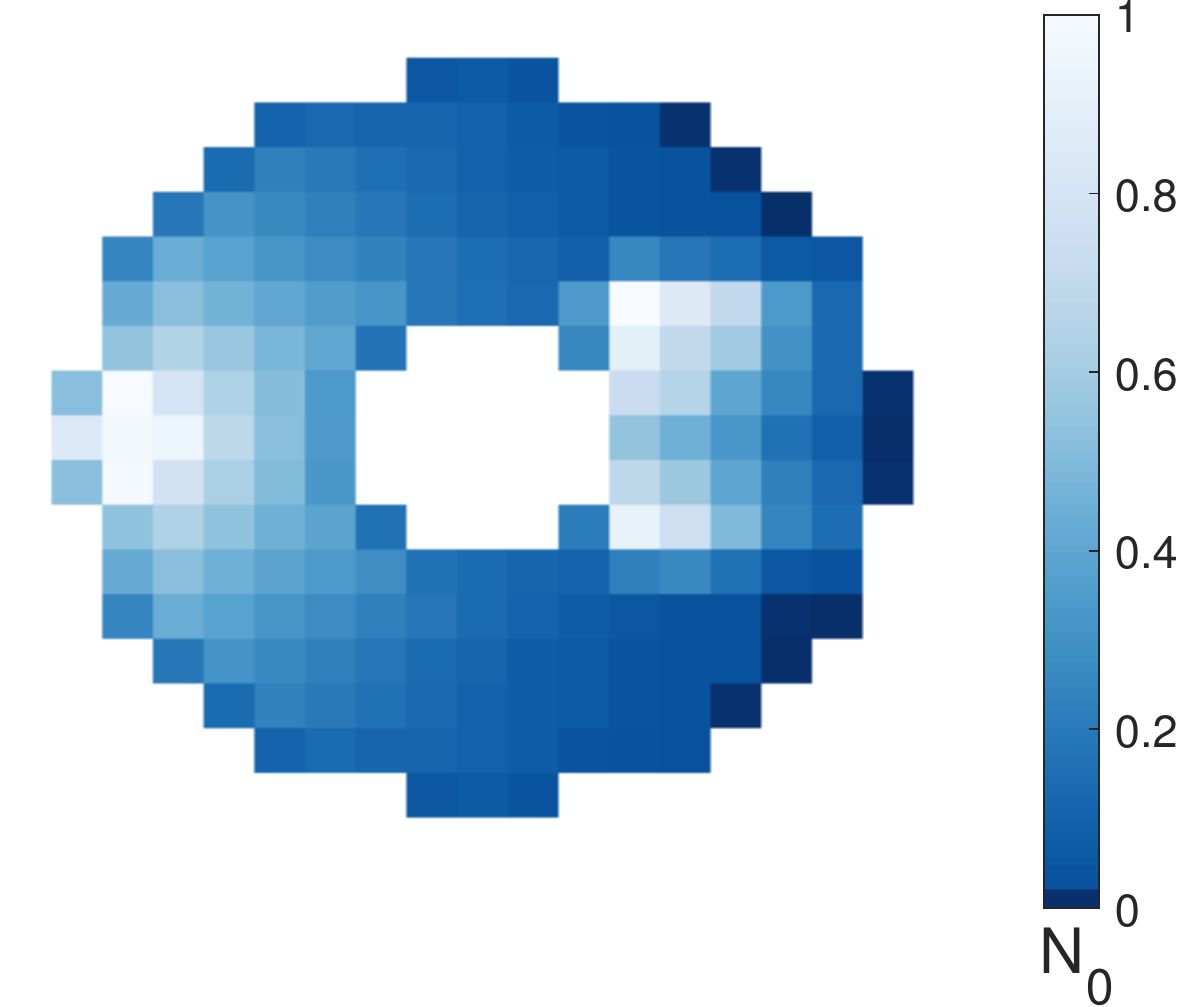}}
    \subfigure[]{\label{fig:RingQlearning}\includegraphics[width=.19\textwidth]{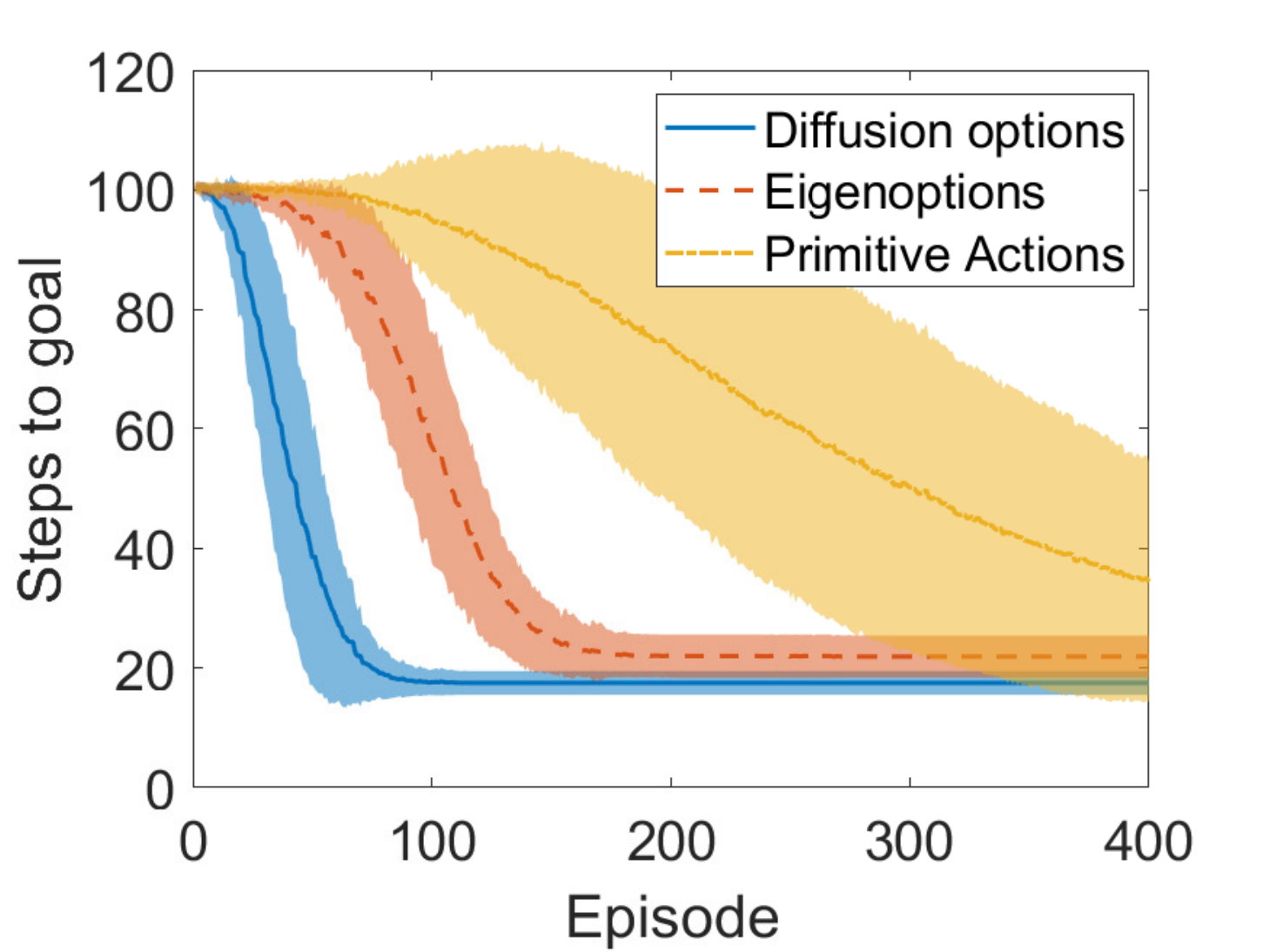}}
    
    \subfigure[]{\label{fig:MazeStartGoal}\includegraphics[width=.19\textwidth]{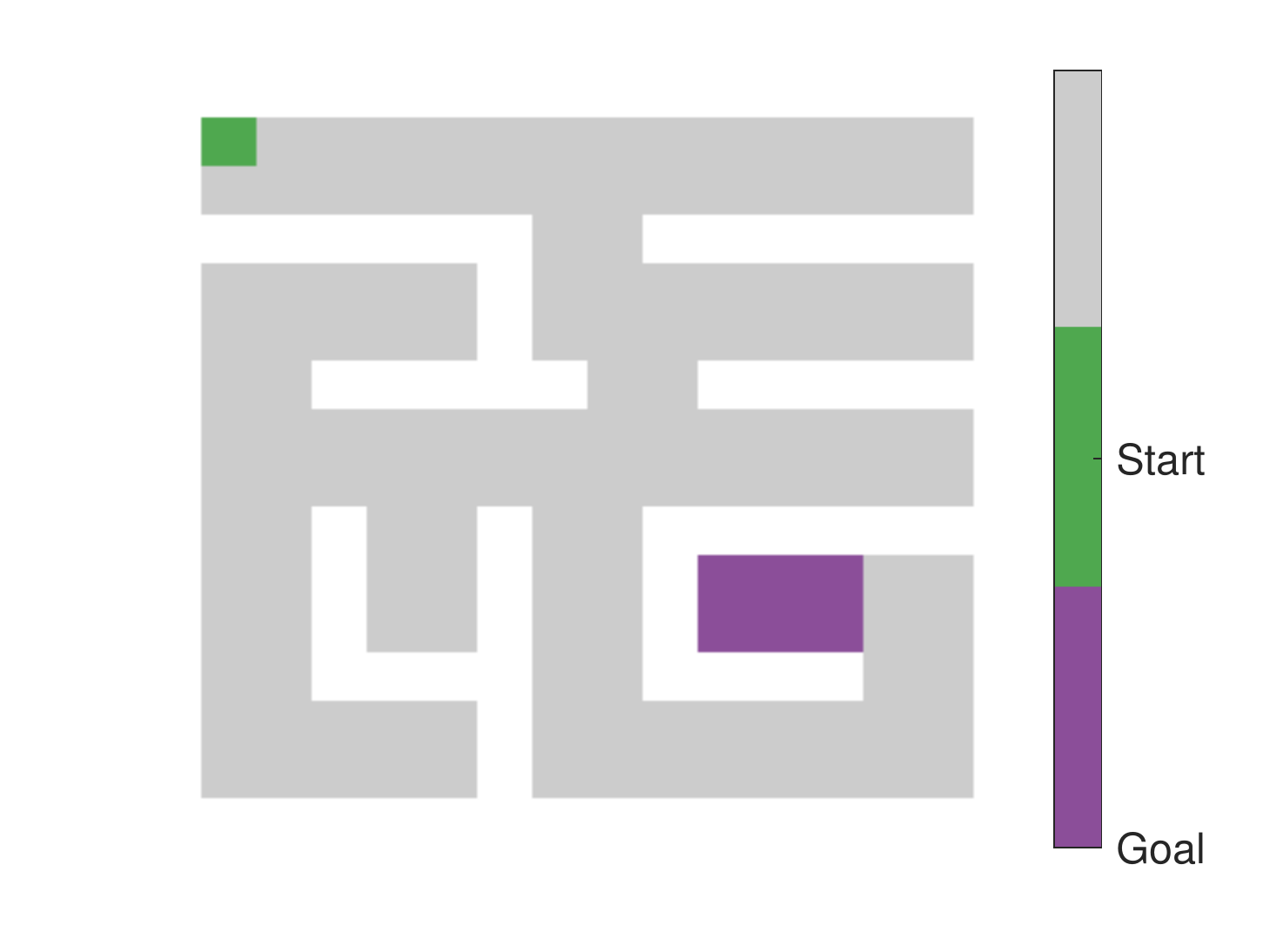}}
    \subfigure[]{\label{fig:MazeDiffusionVisit}\includegraphics[width=.16\textwidth]{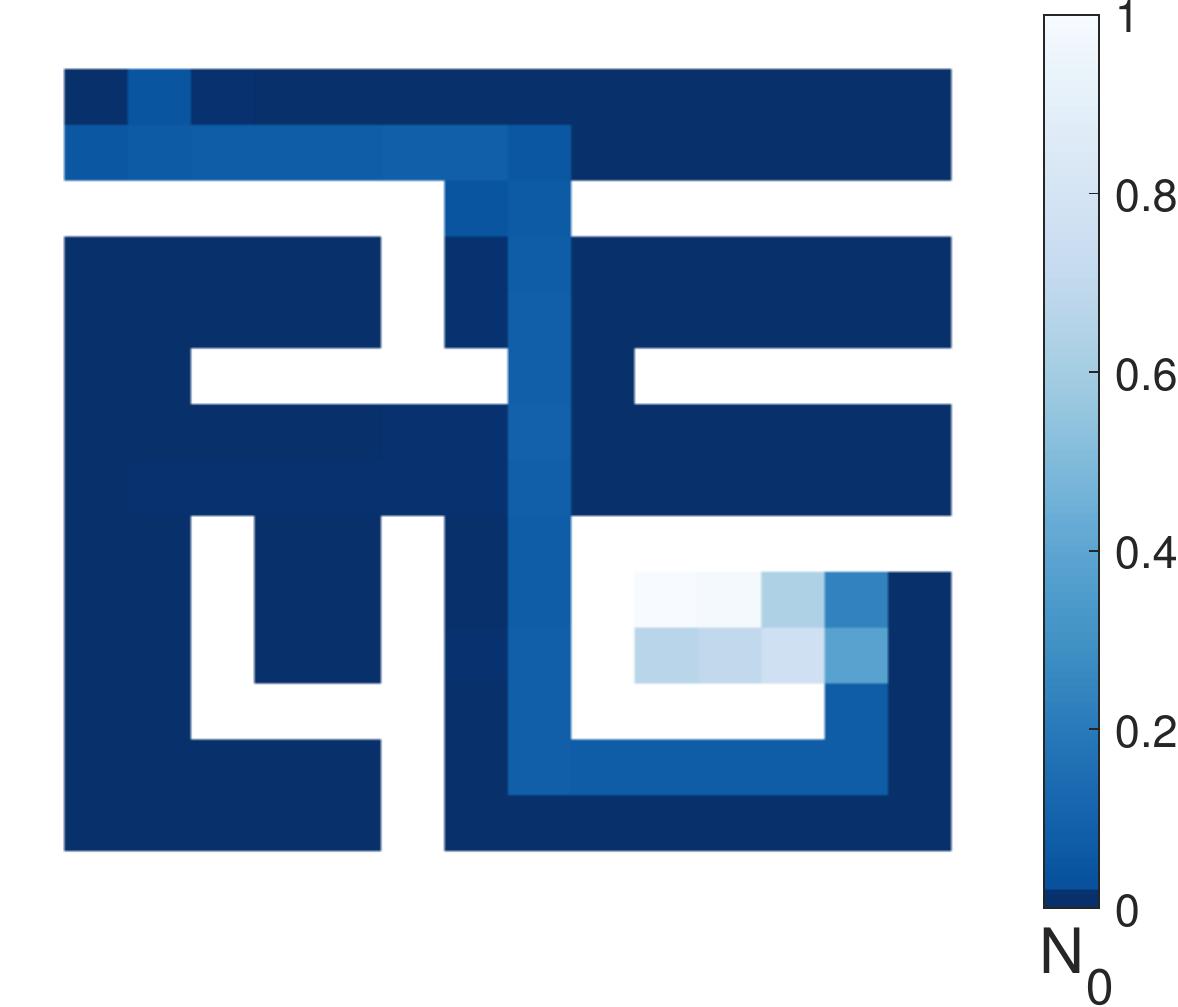}}
    \subfigure[]{\label{fig:MazeLaplacVisit}\includegraphics[width=.16\textwidth]{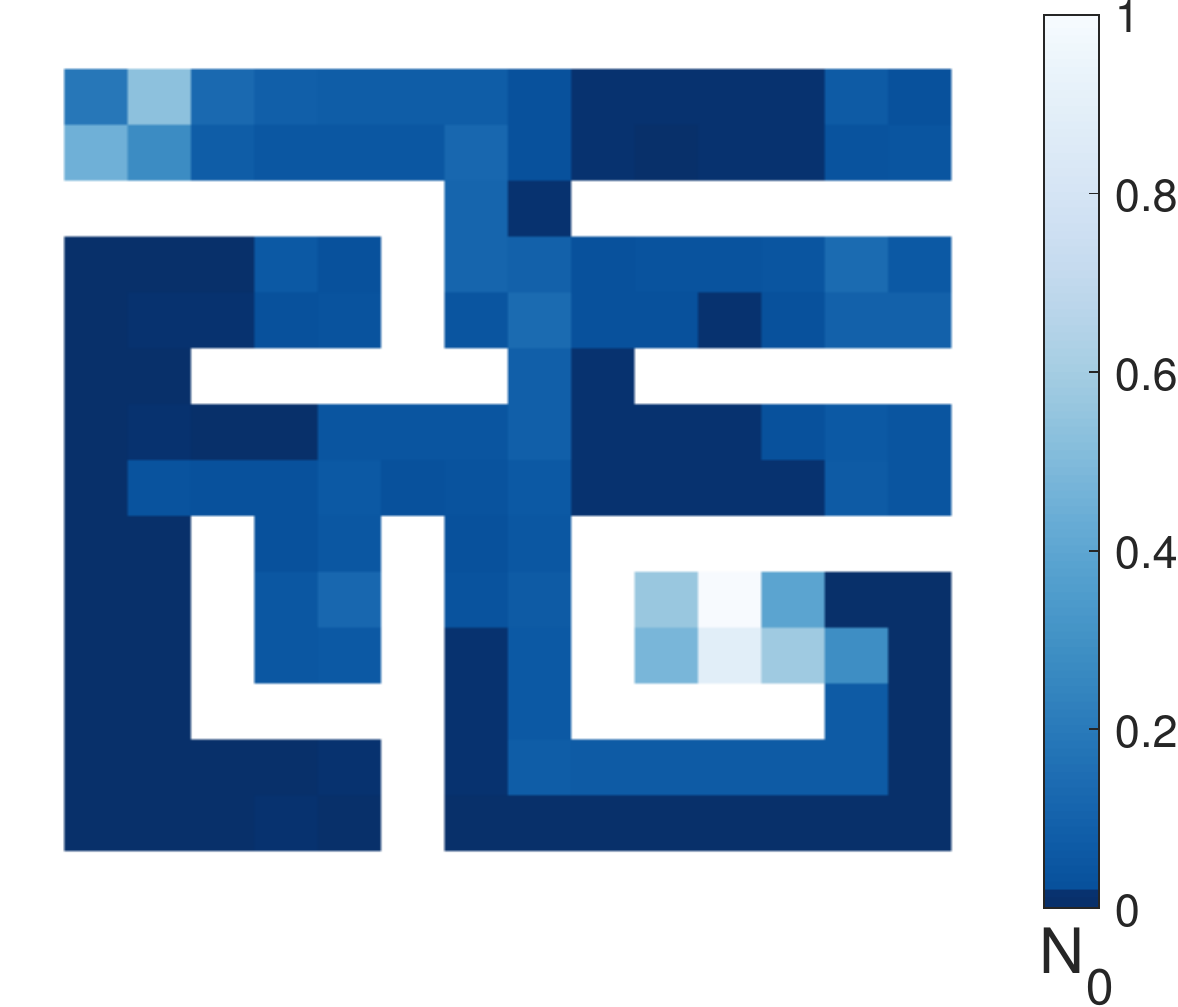}}
    \subfigure[]{\label{fig:MazeRWVisit}\includegraphics[width=.16\textwidth]{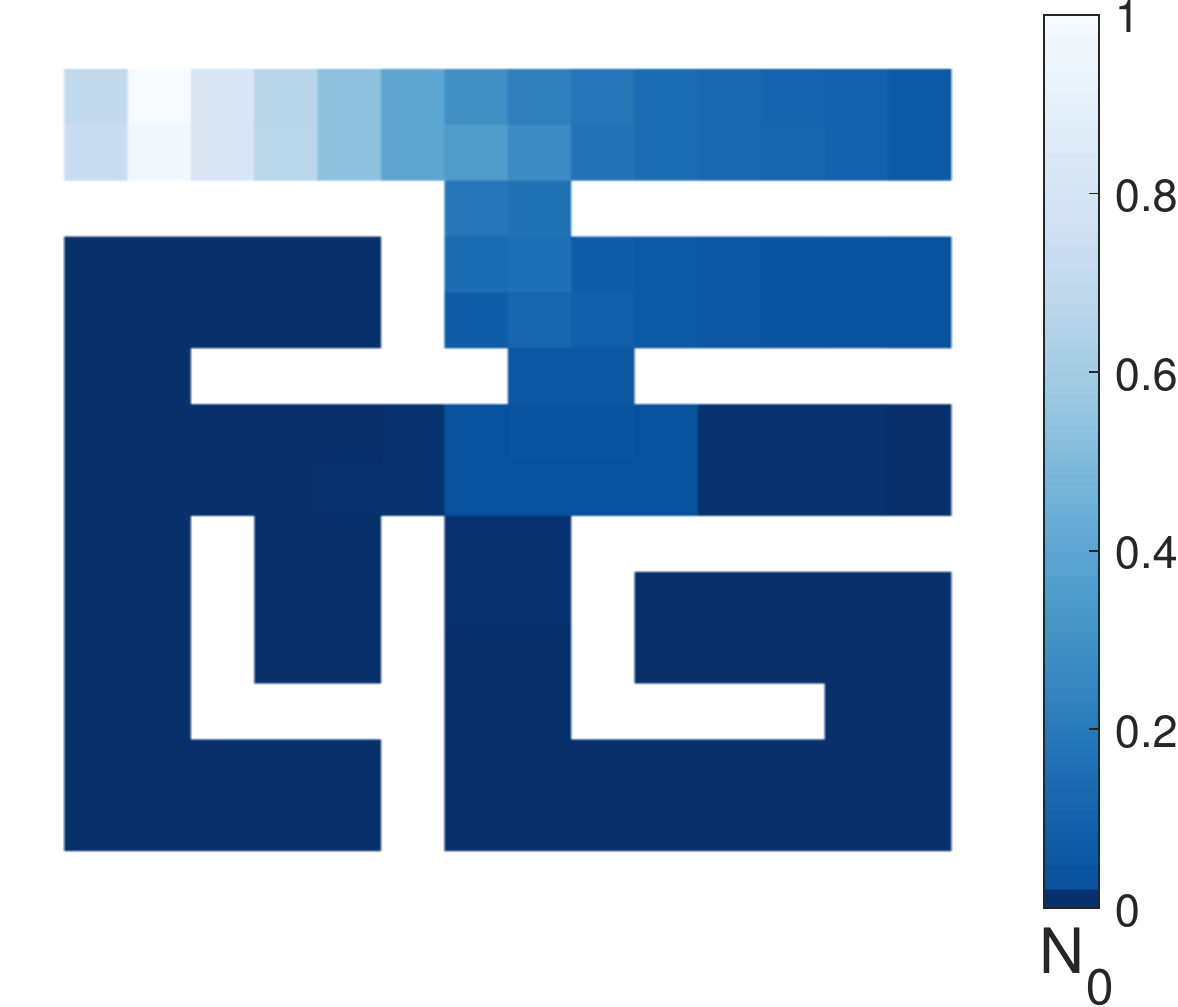}}
    \subfigure[]{\label{fig:MazeQLearning}\includegraphics[width=.19\textwidth]{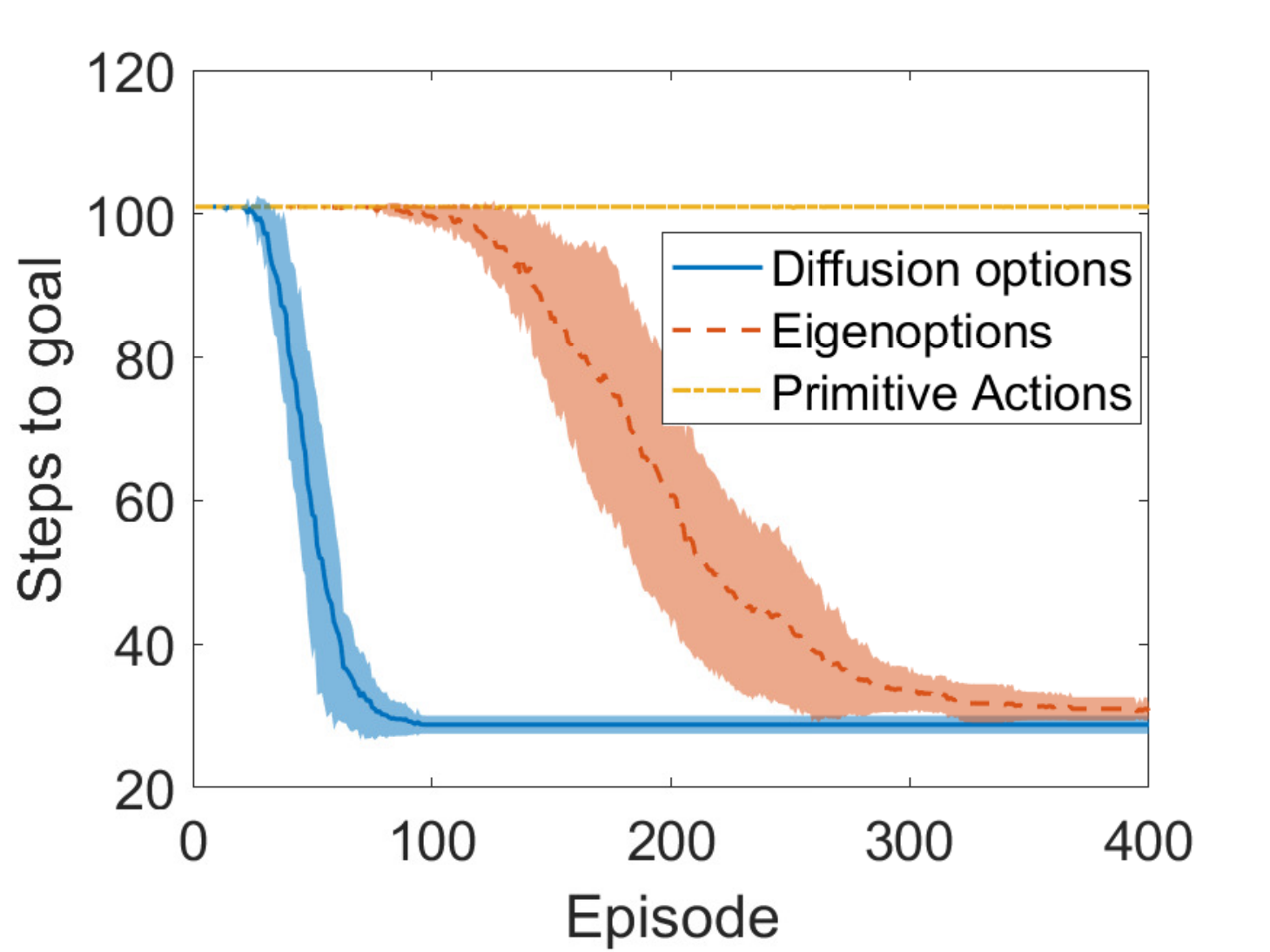}}
    
    \subfigure[]{\label{fig:4RoomsStartGoal}\includegraphics[width=.19\textwidth]{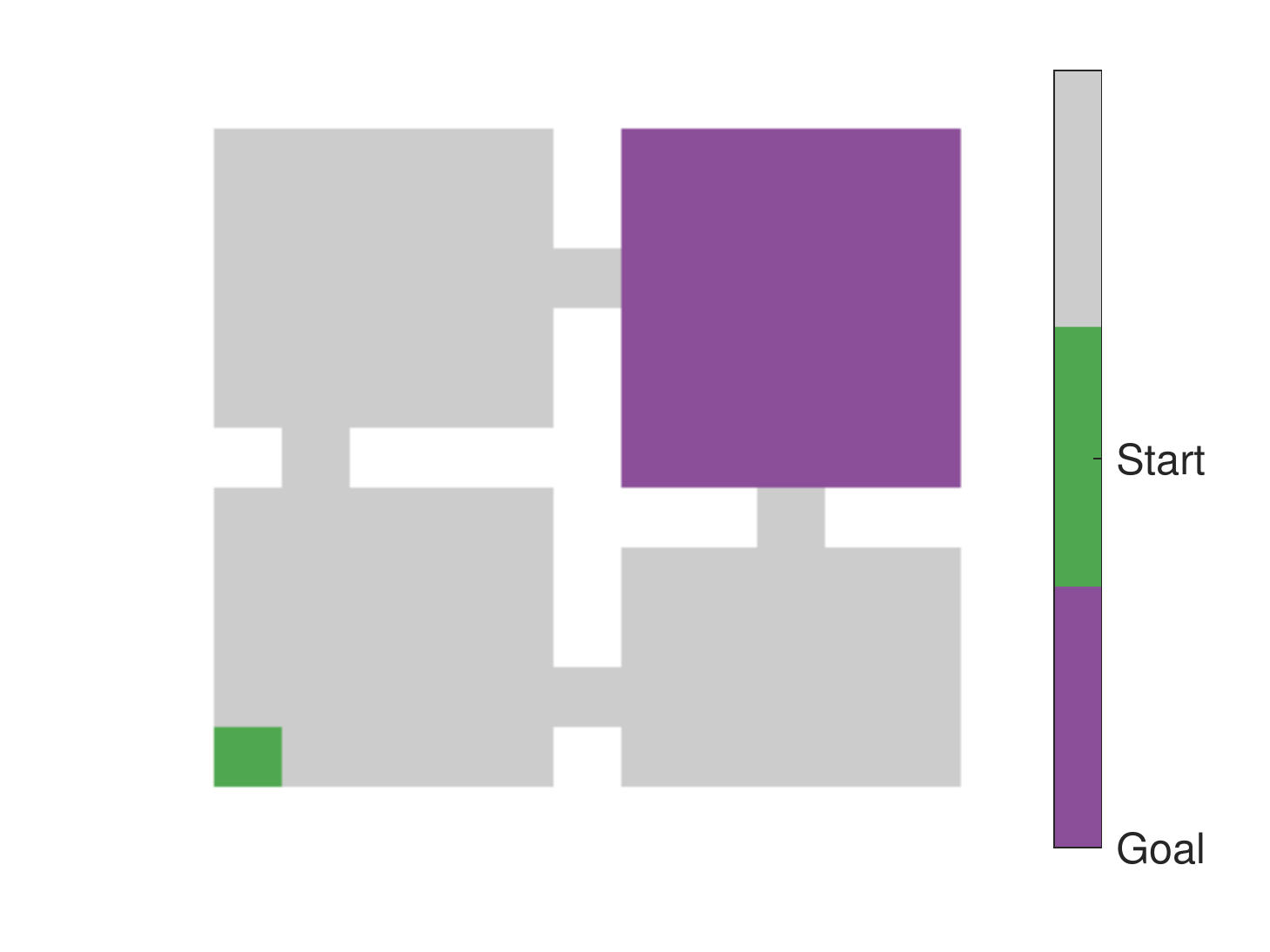}}
    \subfigure[]{\label{fig:4RoomsDiffusionVisit}\includegraphics[width=.16\textwidth]{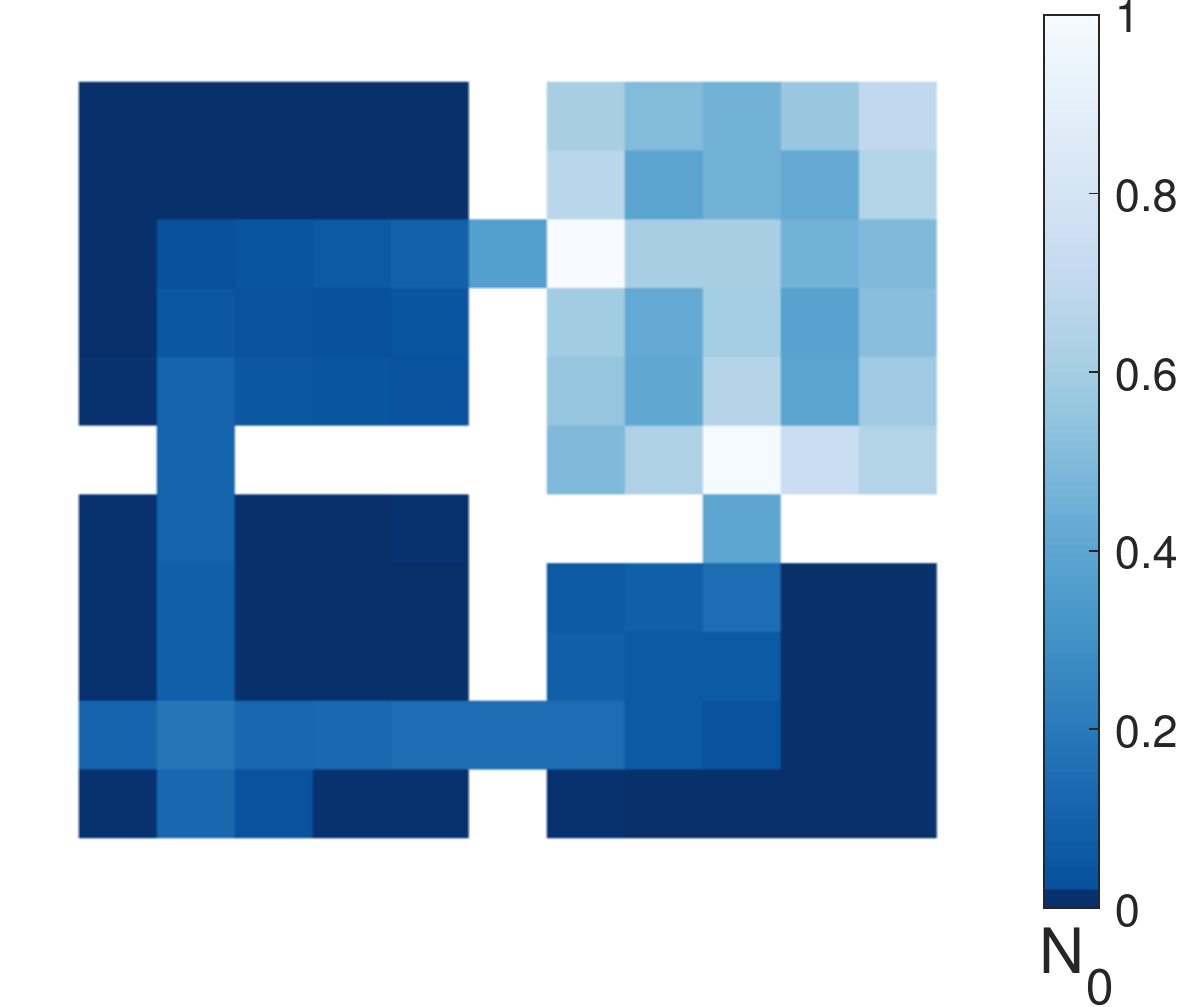}}
    \subfigure[]{\label{fig:4RoomsLAplacianVisit}\includegraphics[width=.16\textwidth]{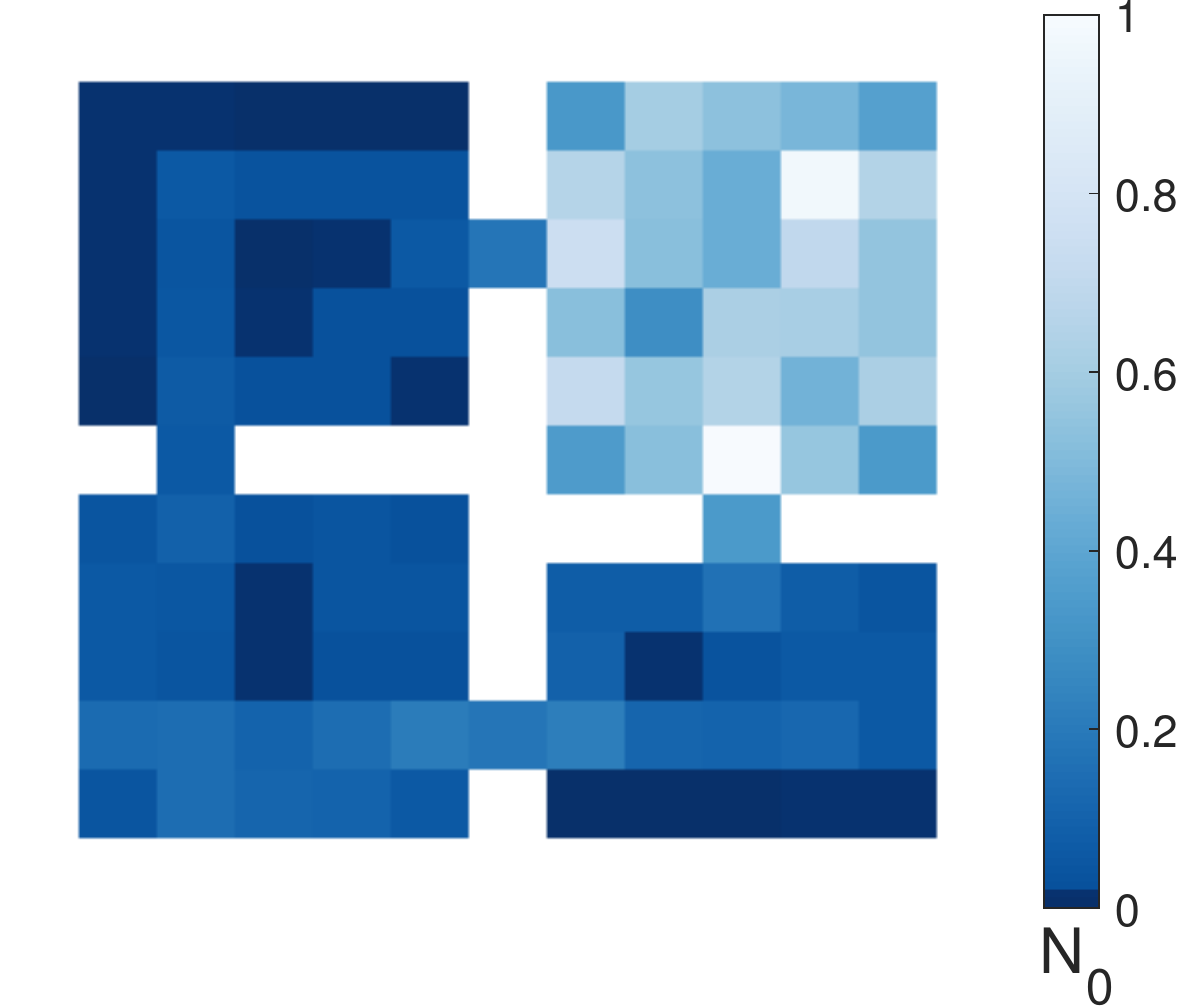}}
    \subfigure[]{\label{fig:4RoomsRWVisit}\includegraphics[width=.16\textwidth]{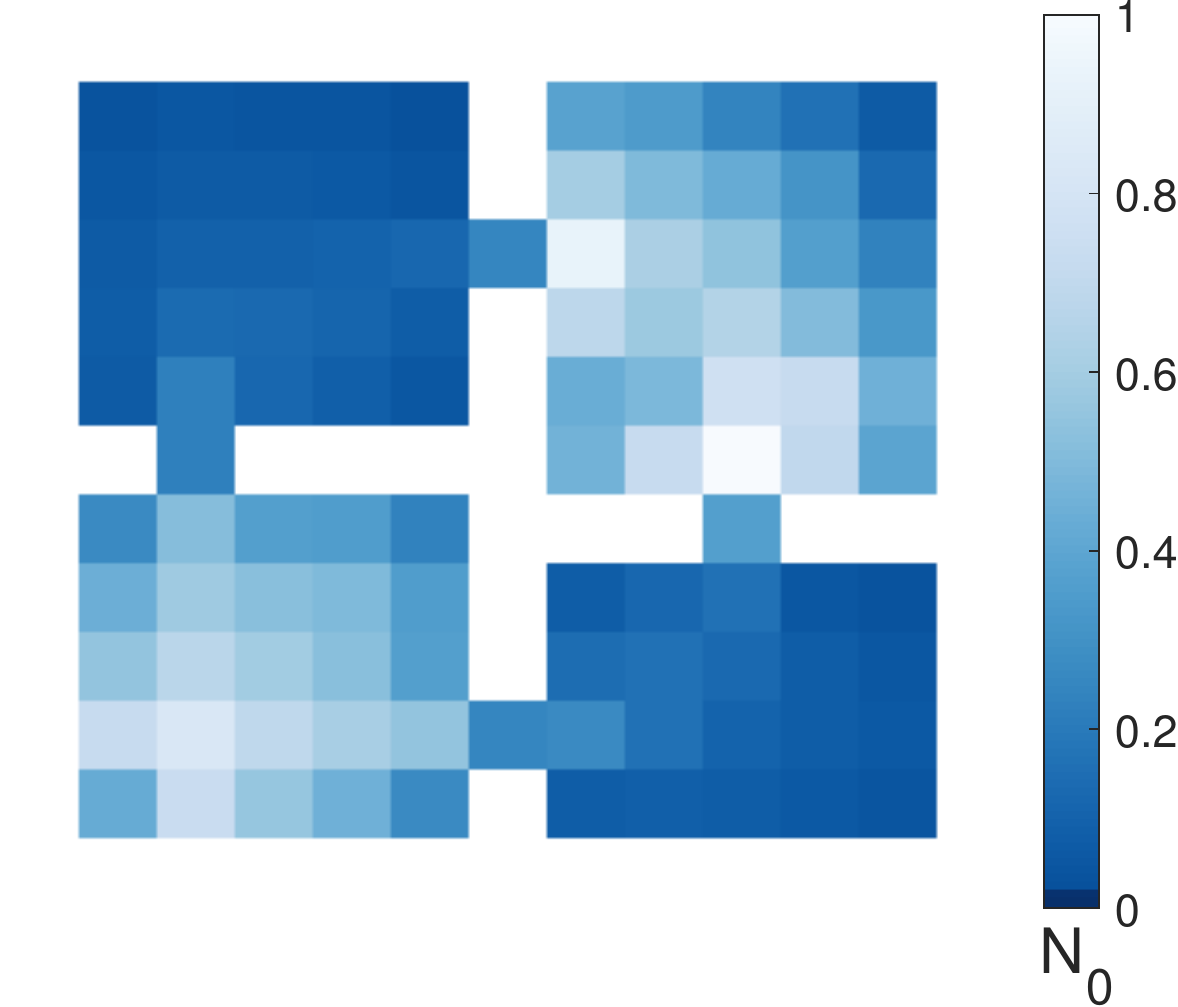}}
    \subfigure[]{\label{fig:4RoomsQLearn}\includegraphics[width=.19\textwidth]{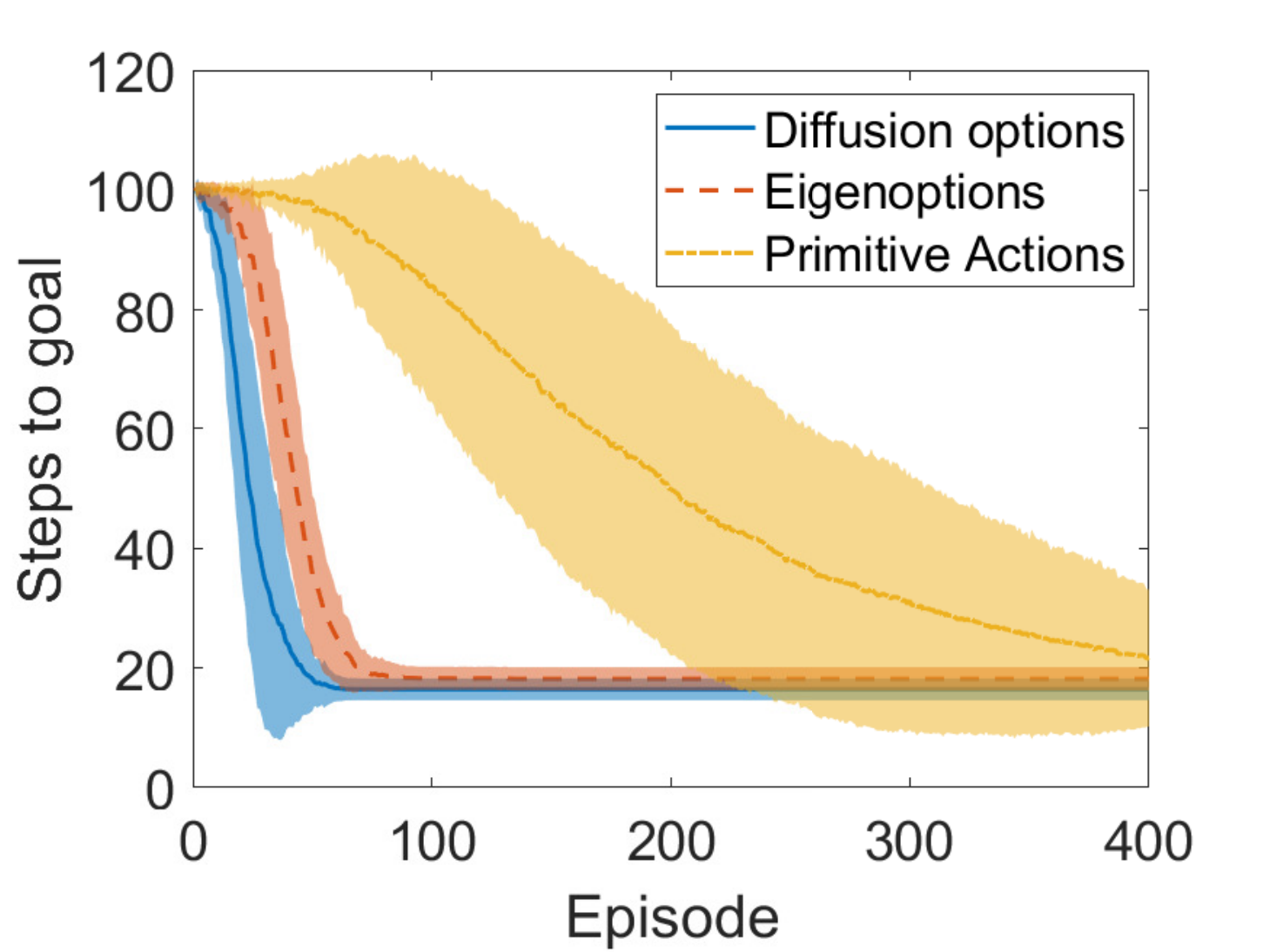}}
    
  \caption{Learning results on the Ring domain (top row), the Maze domain (middle row), and 4Rooms domain (bottom row). (a,f,k) The start state (green) and goal states (purple). (b-d,g-i,l-m) Normalized visitation count $N_0$ obtained based on (b,g,l) the diffusion options, (c,h,m) the eigenoptions, and (d,i,n) a random walk (d). For visualization purposes, the visitation number is normalized to the range of $[0,1]$ by dividing by the maximum number of visitations. (e,j,o) The learning convergence depicting the average number of steps to goal for each learning episode. The solid line represents the mean value and the light colors represent the standard deviation.}
    \label{fig:learning_res}

\end{figure*}

Figure \ref{fig:learning_res} presents the results obtained by setting $t=4$ for all domains. 
We observe in the visitation count plots that the diffusion options lead the agent to the goal states through the shortest path, e.g., in the Ring domain, following the inner ring. 
Importantly, these results are obtained by the diffusion options that were built in advance without access to the location of the start and goal states. 
Conversely, we observe that the eigenoptions lead the agent less efficiently, for example, in the Ring domain, through both the inner and the outer rings. 
While both the diffusion options and the eigenoptions result in informed trajectories to the goal, we observe that the na\"{i}ve random walk tends to concentrate near the start state.

Figure \ref{fig:learning_res} also shows that the diffusion options demonstrate the fastest learning convergence, followed by the eigenoptions and then the random walk. In addition, the diffusion options lead to convergence to shorter paths to a goal compared to the eigenoptions. These convergence results coincide with the visitation count. For example in the Ring domain, by employing the eigenoptions, the agent travels via states at the outer ring which are not on the shortest path to the goal. The significant gap in performance between the diffusion options and the eigenoptions in the Maze domain may be explained by the fact that the option goal states of the diffusion options are located at the end of the corridors (see Fig. \ref{fig:f_t}), leading to efficient exploration, and in turn, to this fast learning convergence.
We note that the zero variance in the learning curves at the beginning of the learning implies that the agent did not reach its goal during the episode, so the same default value was set.

For a fair comparison, we use the same number of options in both algorithms with the same Q learning configuration described above. 
In the SM, we present results, where the number of eigenoptions is tuned to attain maximal performance.
Even after tuning, the diffusion options outperform the eigenoptions.

Table \ref{Table:StepsStates} shows the number of steps between states.
%TODO: move to ack - We thank the reviewers for bringing cover options \citep{jinnai2019discovering,jinnai2020exploration} to our attention. We managed to implement and test them in terms of exploration. 
We note that in contrast to eigenoptions and diffusion options, cover options are point options; see further discussion in Section \ref{Sec: Existing work}.
We observe that the diffusion options lead to more efficient transitions between states compared to the eigenoptions, cover options, and a random walk. This suggests that diffusion options demonstrate better exploration capabilities.

\begin{table*}[t]
\caption{Number of steps between any pair of states using options induced by $t=4$ and by $t=13$. We report the median value and the interquartile range (IQR) over all pairs. See the SM, for mean and standard deviation.}
\label{Table:StepsStates}
\begin{center}
\begin{tabular}{*{11}{c|}}
\multicolumn{1}{c|}{\bf Domain (\#states)}
&\multicolumn{1}{c|}{\bf t}
&\multicolumn{1}{c|}{\bf \#options}
&\multicolumn{2}{c|}{\bf Diffusion Options }
&\multicolumn{2}{c|}{\bf Eigenoptions }
&\multicolumn{2}{c|}{\bf Cover Options }
&\multicolumn{2}{c|}{\bf Random Walk }
\\ 
                      &  &  &  Median  & IQR       &  Median    & IQR       &  Median   &IQR  &  Median   &IQR \\
\hline
\multirow{2}{*}{Ring (192)}     & 4 & 32  & \bf 217       & 101                 &  301    & 210     & 361  & 536      &  565   & 160\\
                                & 13 & 28 & \bf 219       & 110                 &  279    & 232     & 363  & 481       &  565  & 160\\
\hline
\multirow{2}{*}{Maze (148)}     & 4  & 19 & \bf 282      & 194                  & 446      & 573       & 525  & 812    & 1280   & 960 \\
                                & 13 & 14 & \bf 249      & 160                   & 641      & 781      & 498  & 842   & 1280   & 960 \\
\hline
\multirow{2}{*}{4Rooms (104)}  & 4  & 20 &  \bf 147     & 137                    &  160   & 114         & 179  & 512  & 487       & 104\\
                                 & 13 & 15 & \bf 140     & 96                       &  162   & 151      & 175  & 442  & 487    & 104 \\
\end{tabular}
\end{center}
% \label{Table:StepsStates}
\end{table*}
 
\subsection{Stochastic Domains}
\label{sec:StochasticDomainResult}

We revisit the 4Rooms domain with the addition of a stochastic wind blowing downwards. The presence of wind is translated to the probability of $1/3$ that the agent moves down, regardless of its chosen action. As a result, the agent is more likely to visit states at the bottom of the domain, so in principle, the desired options should favor states at the upper parts of the domain.

\begin{figure}[t]
\centering
    \subfigure[]{\label{fig:4RoomsStochasticOurFunc}\includegraphics[width=.44\columnwidth]{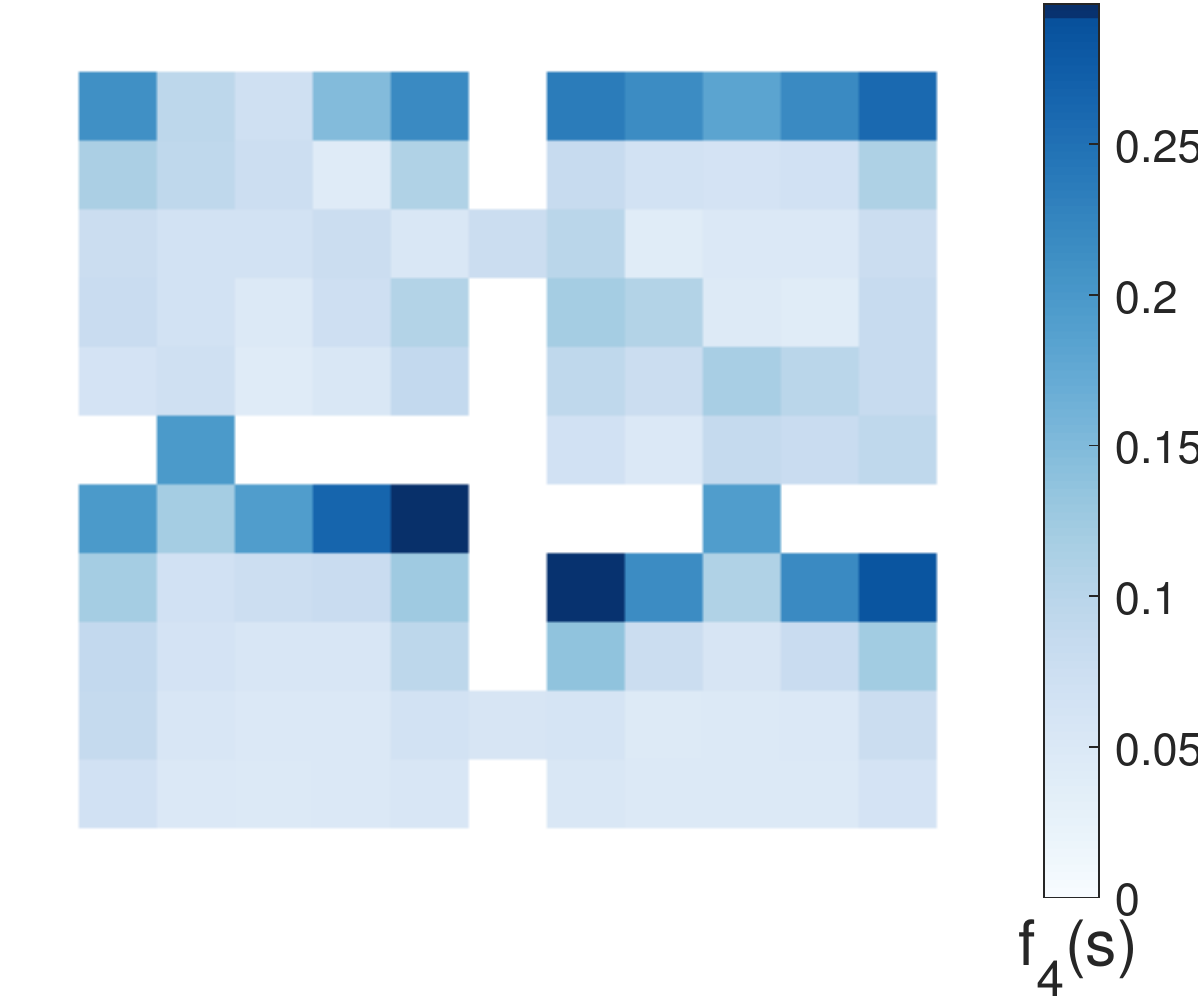}}
    \subfigure[]{\label{fig:4RoomsStochasticQLearnPerf}\includegraphics[width=.48\columnwidth]{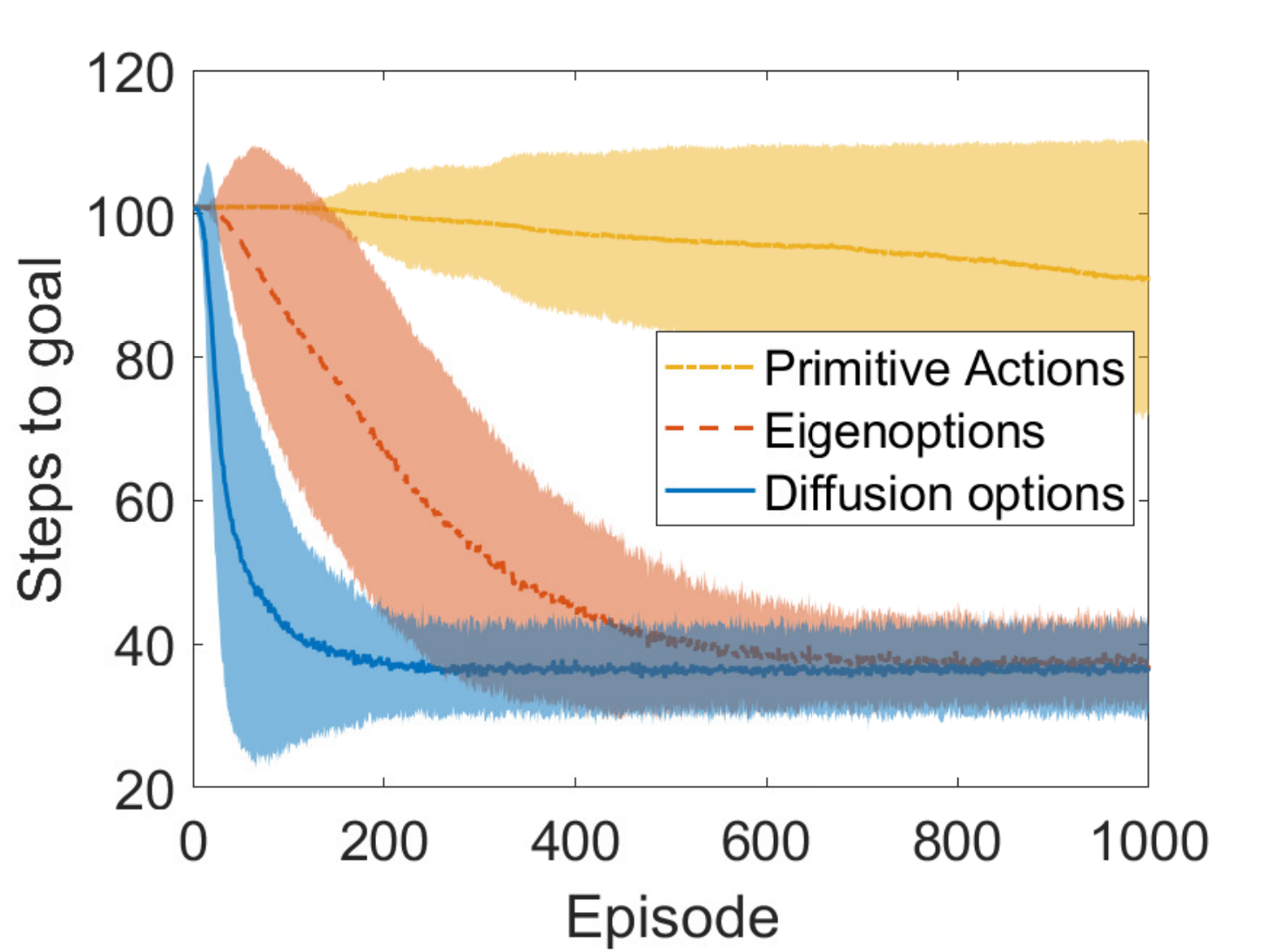}}
 \caption{4Rooms domain with stochastic wind blowing downwards. (a) The domain is colored by $f_4(s)$, where we observe that the local maxima are at the top rooms, compensating for the wind. See Fig. \ref{fig:f_t} for comparison to the result without wind. (b) The obtained learning convergence.}
 \end{figure}

In Fig. \ref{fig:4RoomsStochasticOurFunc}, we observe that $f_4(s)$ now exhibits high values at the upper part of the rooms, rather than high values at the corners and boundaries as in Fig. \ref{fig:4RoomsDiffOpt_tEq4} without the wind. To compare the learning convergence, we adapt the eigenoptions to the stochastic domain by considering the eigenvectors of the positive part of the polar decomposition of the Laplacian as eigenoptions. Figure \ref{fig:4RoomsStochasticQLearnPerf}, presenting the learning convergence, shows a clear advantage to the use of the diffusion options compared to the eigenoptions in this stochastic setting.  

\subsection{Diffusion Distance and Domain Difficulty}
\label{Sec:DiffDistDomComplx}
We empirically show that the diffusion distance is related to the ``domain difficulty''. We propose to approximate the difficulty by the average diffusion distance between every pair of states, and compare it with two other measures of difficulty: the average time duration required for learning a task using primitive actions, i.e. the learning rate, and the average number of steps between pairs of states.
Note that the computation of diffusion distances is intrinsic, i.e., it takes into account only the geometry of the domain. Consequently, it can be computed per domain a-priori without any task assignment or access to rewards. Conversely, the learning rate and the average number of steps are computed in the context of learning particular tasks and rewards, and as a result, convey their difficulties as well.

For each domain, the average diffusion distance between all states is computed. To account for the domain size, we multiply the average diffusion distance by the number of accessible states. 
In addition, we compute the average of diffusion distance over $100$ different scales of $t$ from a regular grid between $1$ and $1000$.

The results are: $13.6$, $20.5$, and $8.6$ for the Ring, the Maze, and the 4Rooms domains, respectively.
We observe that the obtained value in the Maze is higher than the obtained value in the Ring, despite having fewer states.
Indeed, the learning convergence in the Maze is slower (see Figs. \ref{fig:MazeQLearning} and \ref{fig:RingQlearning}) and the average number of steps between states is higher as well (see Table \ref{Table:StepsStates}). 

The relation between the domain difficulty and the diffusion distance gives another justification to the proposed algorithm. 
By Proposition \ref{prop:MainPropAvDiff}, acting according to a diffusion option leads the agent to a distant state in terms of the diffusion distance.
As a result, it can be seen as a way to effectively reduce the domain difficulty.

\section{Relation to Existing Work}
\label{Sec: Existing work}
Option discovery has attracted much interest in recent years, resulting in numerous methods from various perspectives such as information theoretic \citep{mohamed2015variational,florensa2017stochastic,hausman2018learning}, learning hierarchy \citep{bacon2017option,vezhnevets2017feudal}, and curiosity \citep{pathak2017curiosity}, to name but a few.
Discovering options without reward has been a recent active research subject. Combining information theory and skill discovery, \citet{eysenbach2018diversity} proposed to view skills as mixtures of policies, and to derive policies without a reward using an information theoretic objective function. There, a two-stage approach, similar to the present paper, was presented. In the first stage, the domain is scanned with no reward and the options are computed, and in the second stage, the options are utilized for learning in the context of particular rewards.

The notion of ``bottleneck'' states has assumed a central role in option discovery. For example, \citet{menache2002q,csimcsek2005identifying,mannor2004dynamic} propose to define and to identify bottleneck states using graph and spectral clustering methods. Unfortunately, these approaches fail in domains such as the Ring domain, for which clustering is not well defined. An alternative approach presented by \citet{stolle2002learning} defines bottleneck states as frequently visited states. Recently, \citet{goyal2019infobot} showed that this definition might lead to the discovery of redundant options in domains such as a T-shaped domain. 

Perhaps the closest algorithm to ours for option discovery was presented by \citet{machado2017laplacian}. There, the agent uses a subset of eigenvectors of the graph Laplacian of the domain. Each eigenvector (up to a sign) prescribes a value function assigned to each option (termed eigenoption). The agent follows the eigenvector until it reaches a local extremum, where the option terminates. A natural question that arises is why the extrema of the eigenvectors are good option goal states. 
Here we offer a plausible answer from a diffusion distance perspective. \citet{cheng2019diffusion} defined the diffusion distance to a subset $\sB\subset \sS$, and derived a lower bound. In the present paper notation, the formulation of the bound is as follows.
Let $d_{\sB_i}(s)$ be the smallest number of steps, such that the random walk starting from state $s$ reaches the subset $\sB_i$ with probability greater than $\frac{1}{2}$. Then for $\sB_i=\{s\in \sS : -\epsilon\le\psi_i\left(s\right)\le\epsilon\}$ the following holds:
\begin{equation*}
 d_{\sB_i}\left(s\right)\mathrm{log}\left(\frac{1}{\vert1-\nu_i\vert}\right)\ge
\mathrm{log}\left(\frac{\vert\psi_i\left(s\right)\vert}{\Vert\psi_i\Vert_{L^\infty}}\right)-
\mathrm{log}\left(\frac{1}{2}+\epsilon\right),   
\end{equation*}
where $\nu_i$ and $\psi_i$ are a pair of eigenvalue and its associated eigenvector of the normalized graph Laplacian $\mN$.
For small $\epsilon$, the set $\sB_i$ is the set of states for which $\psi_i(s)$ is close to zero. By following an eigenoption defined by the eigenvector $\psi_i$, the agent moves toward states that are distant from the states in $\sB_i$. For instance, consider a domain that is comprised of 2 clusters. For such a domain, $\sB_2$, derived from $\psi_2$, is the set of bottleneck states separating the 2 clusters. Thus, the eigenoption leads the agent away from bottleneck states.  

In contrast, by Proposition \ref{prop:MainPropAvDiff}, the goal states of diffusion options are states that are distant from \emph{all} states (on average). 
Diffusion distance, which is closely related to the proposed options via Proposition \ref{prop:MainPropAvDiff}, takes into account the structure of the domain, including bottlenecks. In addition, Proposition \ref{prop:FuncStationaryDistribution} also implies on the tight relation between diffusion options and bottleneck states because bottlenecks often lie at the minima of the stationary distribution. 

Other graph-based options are cover options \citep{jinnai2019discovering}, which are based on different principles than diffusion options. Cover time is the number of steps it takes to reach every state at least once by a random walk, and cover options attempt to minimize the cover time. Diffusion distance is based on the difference between transition probabilities, and diffusion options attempt to reach states that are seldom visited by a random walk. Perhaps the strongest evidence for the difference between the two options is the fact that cover options are derived only from the Fiedler vector (multiple times), whereas diffusion options from the entire spectrum. We claim that using multiple spectral components captures better the structure of the domain. Another difference is that cover options are point options with limited initiation sets. This limitation does not apply to diffusion options. 
% TODO: not sure if we want to include the following
%The analysis of cover options relies on a bound, similarly to our interpretation of eigenoptions, but in contrast to the exact expressions derived for diffusion options. These bounds rely on one spectral component, whereas we show that incorporating many leads to improved results.

Our options-generating function $f_t(s)$ in (\ref{eq:f_tDefinition}) is related to recent work in data analysis as well.
Similar functions to $f_t(s)$, constructed from the eigenvectors of the graph Laplacian, were proposed for anomaly detection and clustering (\citep{cheng2018geometry,cheng2018spectral}, respectively). Particularly, \citet{cheng2018spectral} introduced and analyzed a function called spectral norm, and showed that the proliferation of eigenvectors is beneficial for clustering. In this work, we show that the same approach of combining all eigenvectors together, rather than using them separately (as the common practice is, for instance in PCA), is beneficial for option discovery.

\section{Conclusions}
We presented a method to derive options based on the full spectrum of the graph Laplacian. The main ingredient in the derivation and the subsequent analysis is the diffusion distance, a notion that was introduced in the context of manifold learning primarily for high-dimensional data analysis. We tested our options using Q learning in three domains, demonstrating improved exploration and learning compared to competing options. 

We believe that a similar approach with such geometric considerations can be beneficial in other problems. Particularly, in future work we plan to explore its use for state aggregation \citep{singh1995reinforcement,duan2019state}. States that belong to the same partition have the same transition probabilities, and as a consequence, the diffusion distance between them is zero. Therefore, it seems only natural to utilize this notion of distance for this problem. 
In addition, we will study the possibility to combine model-based state transition learning with the formation of an empirical graph Laplacian.

\section*{Acknowledgements}
% We thank the reviewers for their helpful suggestions and especially for bringing cover options \citep{jinnai2019discovering,jinnai2020exploration} to our attention. 
This research was partially supported by the Technion Hiroshi Fujiwara cyber security research center and by the Pazy Foundation. The work of RM is partially supported by the Ollendorff Center of the Viterbi Faculty of Electrical Engineering at the Technion, and by the Skillman chair in biomedical sciences.

\bibliography{OptionDiscoveryManifoldAnalysis}
\bibliographystyle{icml2020}

%%%%%%%%%%%%%%%%%%%%%%%
% SM
%%%%%%%%%%%%%%%%%%%%%%%

\title{Supplementary Material for Option Discovery in the Absence of Rewards with Manifold Analysis}
\date{}
\onecolumn

\maketitle

\setcounter{section}{0}
\renewcommand{\thesection}{\Alph{section}}
\section{Diffusion Distance}

The diffusion distance between two points $s$ and $s'$, as defined in section 2.2, can be computed using all eigenvectors of the random walk matrix $\mW$ \citep{coifman2006diffusion}: 
\begin{equation}
    \label{eq:DiffDistEigenVec}
    D_t^2(s,s')=\sum_{i\ge0}\omega_i^{2t}(\tilde{\phi}_i(s)-\tilde{\phi}_i(s'))^2,
\end{equation}
where $\{\tilde{\phi}_i\}$ and $\{\omega_i\}$ are the right eigenvectors and eigenvalues of $\mW$, respectively.
Expressing the diffusion distance using the full spectrum of $\mW$ is an important property, which is shown to be useful for option discovery, as we show in this work.

Consider the $l$-dimensional \emph{diffusion maps} embedding, where $l \le |\sS|$, defined by $\left[\Psi_t(x)\right]_i\triangleq\omega^t_i\tilde{\phi}_{(i)}(x)$ for $i=1,\ldots,l$. It was shown in \citet{coifman2006diffusion} that the diffusion distance can be well approximated by the Euclidean distance between the the diffusion maps, i.e., \begin{equation*} 
    D_t(s,s') \approx \Vert \Psi_t(s)-\Psi_t(s') \Vert
\end{equation*}
where equality holds if all the eigenvectors are used ($l=|\sS|$). This approximation prescribes a convenient way to compute the diffusion distance efficiently.

\begin{figure}[h]
  \begin{center}
\subfigure[]{\label{Fig:EucDist}\includegraphics[width=.39\columnwidth]{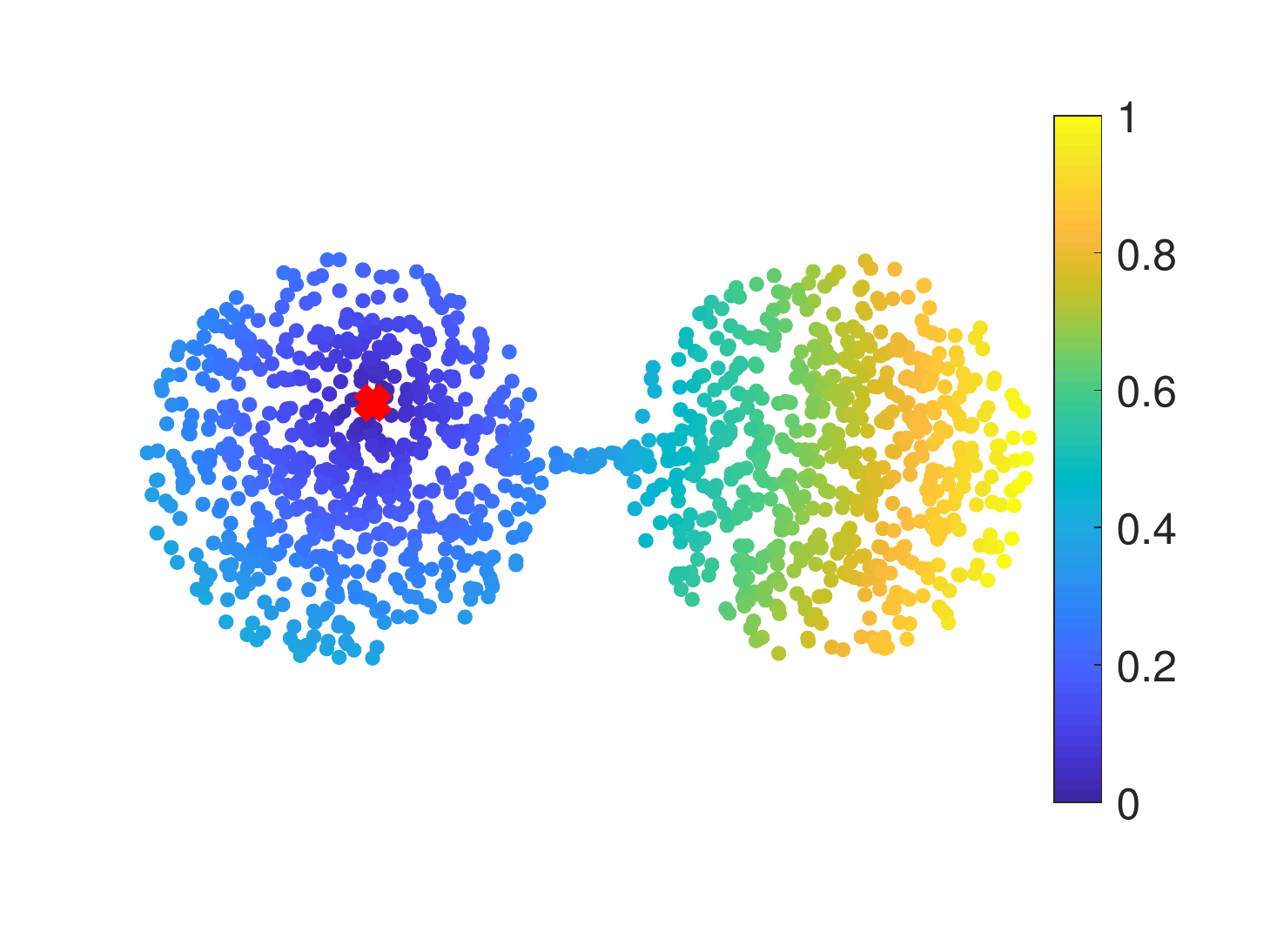}}
\subfigure[]{\label{Fig:DiffDist}\includegraphics[width=.39\columnwidth]{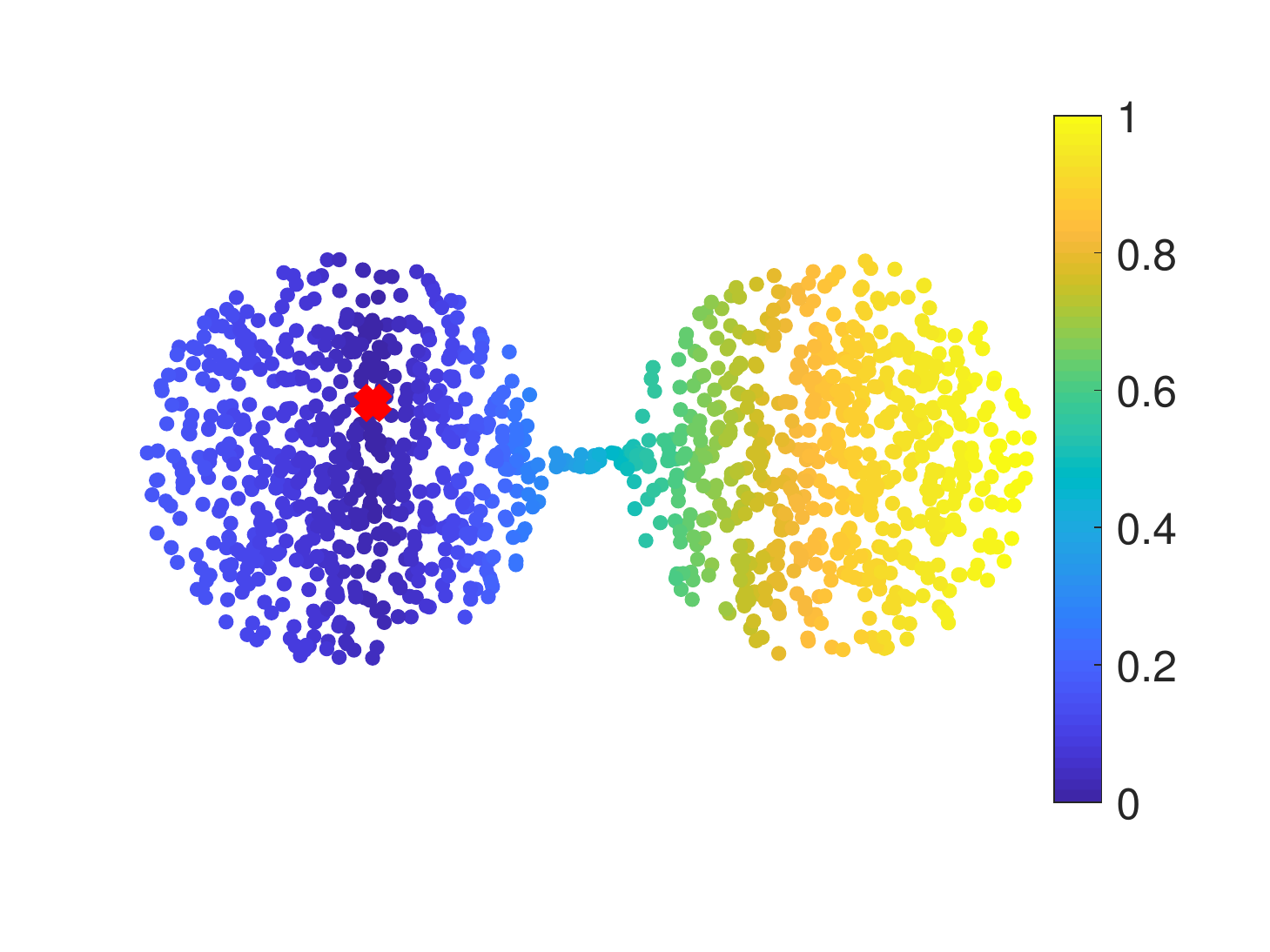}}
\caption{Diffusion distance illustration. (a) The Euclidean distances and (b) the diffusion distances from a reference point (marked as a red cross). In contrast to the Euclidean distances, the diffusion distances in the right cluster are larger compared to the diffusion distances in the left cluster.}
\label{Fig:EucDiffDist}
\end{center}
\end{figure}

An important property of the diffusion distance is that it captures both the local and the global structure of a data set. A prototypical example of a point cloud with a dumbbell shape is presented in Fig.~\ref{Fig:EucDiffDist}. Consider an arbitrary reference point in the left cluster, marked in red. The remaining points are colored according to their distance to this reference point. In Fig.~\ref{Fig:EucDist} the distance is the Euclidean distance and in Fig.~\ref{Fig:DiffDist} the distance is the diffusion distance.
We observe that the diffusion distance captures the two-cluster structure of the data set;
the diffusion distances of the points residing in the right cluster are larger than the diffusion distances of the points residing in the left cluster, with a sharp transition at the bottleneck. Conversely, the Euclidean distances do not demonstrate such a clear transition between the two clusters.

\section{The Normalized Graph Laplacian}

In many applications involving graphs, the construction of the normalized graph Laplacian $\mN$ and its spectral decomposition were shown to be useful. 
The normalized graph Laplacian is defined by
\begin{equation}
\label{eq:Definition_N}
    \mN=\mD^{-\frac{1}{2}}\left( \mI-\mM \right) \mD^{-\frac{1}{2}},
\end{equation}
where $\mM$ is the symmetric adjacency matrix and $\mD$ is corresponding diagonal degree matrix.

A symmetric adjacency matrix $\mM$ leads to a symmetric $\mN$. Therefore, $\mN$ has real eigenvalues, denoted by $\{\nu_i\}$, and orthonormal eigenvectors $\{\psi_i\}$.

The eigenvalue decomposition of the lazy random walk matrix $\mW$, given by $\mW=\frac{1}{2}(\mI+\mM\mD^{-1})$, is tightly related to the eigenvalue decomposition of the normalized graph Laplacian $\mN$. 
First, observe that $\mW$ can be recast as
$\mW=\mD^{\frac{1}{2}}\left(\mI-\frac{1}{2}\mN\right)\mD^{-\frac{1}{2}}=\mI - \frac{1}{2} \mD^{\frac{1}{2}} \mN \mD^{-\frac{1}{2}}$. Then, by definition, $\mN$ and $\mD^{\frac{1}{2}} \mN \mD^{-\frac{1}{2}}$ are similar, and so they share their real eigenvalues and the following holds: 
\begin{align}
    \label{eq:OmegaNuRelations}
    \omega_i&=\left(1-\frac{1}{2}\nu_i\right), \nonumber \\ 
    \phi_i&=\mD^{-\frac{1}{2}}\psi_i, \nonumber \\
    \tilde{\phi_i}&=\mD^{\frac{1}{2}}\psi_i,
\end{align}
where $\omega_i$ are the eigenvalues of $\mW$ and $\phi_i$ and $\tilde{\phi}_i$ are its left and right eigenvectors. As a consequence, Algorithm 1 in the paper can be implemented as follows. First, the normalized graph Laplacian $\mN$ is computed instead of $\mW$. Then, eigenvalue decomposition (EVD) is applied to $\mN$. Finally, $\{\omega_i\}$, $\{\phi_i\}$, and $\{\tilde{\phi}_i\}$ are computed based on (\ref{eq:OmegaNuRelations}), and the remainder of the algorithm is completed as is.

\section{Proofs}
\subsection{Proof of Proposition 1}% 

The setting and notation follow \citet{DanielSpielmanLectures}. 

We consider the lazy random walk matrix, defined by $\mW=\frac{1}{2}\mI+\frac{1}{2}\mM\mD^{-1}$ for a symmetric adjacency matrix $\mM$ and the corresponding diagonal degree matrix $\mD$. Using (\ref{eq:Definition_N}), we can rewrite $\mW$ as $\mW=\mD^{\frac{1}{2}}\left(\mI-\frac{1}{2}\mN\right)\mD^{-\frac{1}{2}}$. 

The probability distribution after $t$ steps of the lazy random walk is 
$\vp_t=
\mW^t\vp_0=
\mD^{\frac{1}{2}}\left(\mI-\frac{1}{2}\mN\right)^t\mD^{-\frac{1}{2}}\vp_0$,
for an initial distribution $\vp_0$.

Let $\nu_i$ and $\psi_i$ be the eigenvalues and eigenvectors of $\mN$.
Using (\ref{eq:OmegaNuRelations}), expanding $\mD^{-\frac{1}{2}}\vp_0$ in the orthonormal basis $\{\psi_i\}$, we get
\begin{equation*}
\vp_t=\mD^{\frac{1}{2}}\sum_{i\ge 1}\left(\mI-\frac{1}{2}\mN\right)^t c_i\psi_i
=\mD^{\frac{1}{2}}\sum_{i\ge 1}\omega_i^t c_i\psi_i,
\end{equation*}
where $\omega_i$ are eigenvalues of $\mW$ and $c_i=\psi_i^T\left(\mD^{-\frac{1}{2}}\vp_0\right)$.

Plugging in the explicit form of the coefficients $c_i$ yields
\begin{equation}
\label{eq:propagation}
    \vp_t=\mD^{\frac{1}{2}}c_1\psi_1+
\mD^{\frac{1}{2}}\sum_{i\ge{2}}\omega_i^t\psi_i^T\left(\mD^{-\frac{1}{2}}\vp_0\right)\psi_i.
\end{equation}

Let $\vd$ denote the degree vector, i.e. $\vd=\text{diag}(\mD)$. The principal eigenvector $\psi_1$ can be expressed as $\psi_1=\frac{\vd^{\frac{1}{2}}}{\Vert\vd^{\frac{1}{2}} \Vert}$ and $c_1$ as $c_1=\frac{1}{\Vert\vd^{\frac{1}{2}} \Vert}$.
Consequently, the first term in (\ref{eq:propagation}) is given by $\mD^{\frac{1}{2}}c_1\psi_1=\frac{\vd}{\Vert \vd^{\frac{1}{2}}\Vert^2}=\pi_0$, where $\pi_0$ is the stationary distribution of $\mM\mD^{-1}$ (and also of $\mW$), namely $\mM\mD^{-1}\pi_0=\pi_0$. 

Suppose the initial distribution $\vp_0$ is completely concentrated at a single state $s$, i.e.,
$\vp_0=\delta_s$, where $\delta_s$ is a vector of all zeros expect the $s$th entry.
The distribution after $t$ steps starting with $\delta_s$ is given by
\begin{equation}
    \label{eq:P_tPiOurFunc}
\vp_t^{(s)}=\mathrm{\pi_0}+
\mD^{\frac{1}{2}}\frac{1}{\sqrt{\vd\left(s\right)}}\sum_{i\ge{2}}\omega_i^t\psi_i^T\delta_s\psi_i.
\end{equation}
Using (\ref{eq:P_tPiOurFunc}), we compute the squared diffusion distance between two states, $s$ and $s'$, after $t$ steps, as follows:
\begin{equation*}
  D_t^2\left(s,s'\right)=\Vert\vp_t^{(s)}-\vp_t^{(s')}\Vert^2=
\Vert \mD^{\frac{1}{2}}\sum_{i\ge{2}}\omega_i^t\left(\frac{\psi_i\left(s\right)}{\sqrt{\vd\left(s\right)}}-\frac{\psi_i\left(s'\right)}{\sqrt{\vd\left(s'\right)}}\right)\psi_i\Vert ^2 = 
\Vert \sum_{i\ge{2}}\omega_i^t\left(\phi_i\left(s\right)-\phi_i\left(s'\right)\right)\Tilde{\phi_i}\Vert ^2  
\end{equation*}
using (\ref{eq:OmegaNuRelations}).
By expanding the norm, we get
\begin{equation*}
    D_t^2\left(s,s'\right)=
\Vert\sum_{i\ge{2}}\omega_i^t\phi_i\left(s\right)\Tilde{\phi_i}\Vert^2+
\Vert\sum_{i\ge{2}}\omega_i^t\phi_i\left(s'\right)\Tilde{\phi_i}\Vert^2-
2\sum_{i\ge{2}}\omega_i^t\phi_i\left(s\right)\Tilde{\phi_i}^T\sum_{j\ge{2}}\omega_j^t\phi_j\left(s'\right)\Tilde{\phi_j}.
\end{equation*}

Finally, by Lemma \ref{Lem:PhiSumZero}, we have
\begin{equation*}
  \left<D_t^2\left(s,s'\right)\right>_{s'\in\sS}=
\Vert\sum_{i\ge{2}}\omega_i^t\phi_i\left(s\right)\Tilde{\phi_i}\Vert^2+
\left<\Vert\sum_{i\ge{2}}\omega_i^t\phi_i\left(s'\right)\Tilde{\phi_i}\Vert^2\right>_{s'\in\sS}  
\end{equation*}
where the second term is a constant independent of $s$.

\begin{lemma}
\label{Lem:PhiSumZero}
Define $\tilde{\phi_i}$ as in (\ref{eq:OmegaNuRelations}). Then, assuming a connected graph, we have 
\begin{equation*}
    \langle \Tilde{\phi_i}(s) \rangle _{s \in \sS} =0 \hspace{1cm}  \forall i\ge2
\end{equation*}
\begin{proof}

The eigenvectors $\{\psi_i\}$ are orthonormal, so $\psi_1^T\psi_i=0$ for all  $i\ge2$.\newline
Using $\psi_1=\frac{\vd^{\frac{1}{2}}}{\Vert \vd^{\frac{1}{2}}\Vert}$ leads to 
$$
0=\psi_1^T\psi_i=
\frac{\left(\vd^{\frac{1}{2}}\right)^T}{\Vert \vd^{\frac{1}{2}}\Vert}\psi_i =
\frac{\left(\vd^{\frac{1}{2}}\right)^T}{\Vert \vd^{\frac{1}{2}}\Vert}\mD^{-\frac{1}{2}}\Tilde{\phi_i}=
\frac{1}{\Vert \vd^{\frac{1}{2}}\Vert}\mathbf{1}^T\Tilde{\phi_i} = \frac{1}{\Vert \vd^{\frac{1}{2}}\Vert} \sum _{s \in \sS} \Tilde{\phi_i}(s)
$$
\end{proof}
\end{lemma}

\subsection{Proof of Proposition 2}% 
\paragraph{Part 1.}
Combining (\ref{eq:P_tPiOurFunc}) and  (\ref{eq:OmegaNuRelations}) yields
\begin{equation*}
    \vp_t^{(s)}-\mathrm{\pi_0}=
\mD^{\frac{1}{2}}\frac{1}{\sqrt{\vd\left(s\right)}}\sum_{i\ge{2}}\omega_i^t\psi_i^T\delta_s\psi_i=
\sum_{i\ge{2}}\omega_i^t\phi_i\left(s\right)\Tilde{\phi_i},
\end{equation*}
and we have 
\begin{equation*}
    f_t(s)\triangleq
\Vert\sum_{i\ge{2}}\omega_i^t\phi_i\left(s\right)\Tilde{\phi_i}\Vert^2=\Vert\vp_t^{(s)}-\mathrm{\pi_0}\Vert^2.
\end{equation*}

\paragraph{Part 2.}
The concatenation of the eigenvectors $\{\psi_i\}$ as columns of a matrix forms an orthonormal matrix, whose rows are also orthonomal. So,
\begin{equation*}
    \sum_{i\ge{2}}\psi_i^2\left(s\right)=1-\psi_1^2\left(s\right)=
1-\frac{\vd\left(s\right)}{d\left(\sS\right)}=
\frac{d\left(\sS\right)-\vd\left(s\right)}{d\left(\sS\right)},
\end{equation*}
where $d(\sS)$ is the degree of the set of all states, $\sS$.
We therefore have
\begin{equation*}
    \frac{d\left(\sS\right)}{\vd\left(s\right)}\sum_{i\ge{2}}\psi_i^2=
\frac{d\left(\sS\right)-\vd\left(s\right)}{\vd\left(s\right)}=
\frac{d\left(\sS\right)}{\vd\left(s\right)}-1=
\frac{1}{\pi_0\left(s\right)}-1.
\end{equation*}
Finally, assuming the eigenvalues $\omega_i$ are in descending order, we have
\begin{align*}
    f_t\left(s\right)&=
\Vert\sum_{i\ge{2}}\omega_i^t\phi_i\left(s\right)\Tilde{\phi}_i\Vert^2 \le
\omega_2^{2t}\Vert\sum_{i\ge{2}}\phi_i\left(s\right)\Tilde{\phi}_i\Vert^2 \\
    &\le \omega_2^{2t}\sum_{i\ge{2}}\frac{\psi_i^2\left(s\right)}{\vd\left(s\right)}\Vert \mD^{\frac{1}{2}}\psi_i\Vert^2
\le\omega_2^{2t}\sum_{i\ge{2}}\frac{\psi_i^2\left(s\right)}{\vd\left(s\right)}d\left(\sS\right)\\
&=\omega_2^{2t}\left(\frac{1}{\pi_0\left(s\right)}-1\right).
\end{align*}

\section{Additional Experimental Results}

\subsection{Further Comparison with Eigenoptions} 
\begin{figure}
\centering
\subfigure[]{\includegraphics[width=.30\columnwidth]{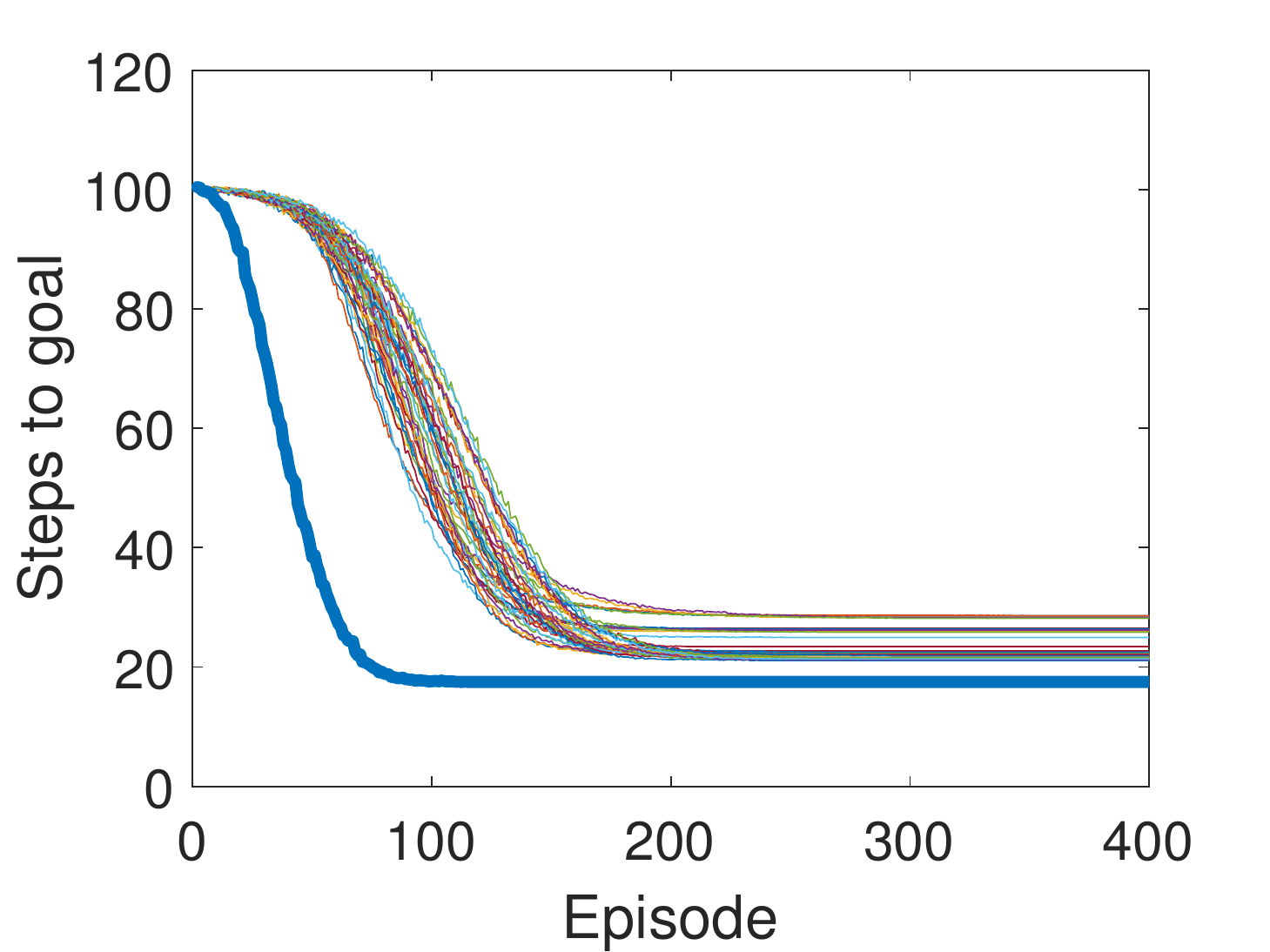}}
\subfigure[]{\includegraphics[width=.30\columnwidth]{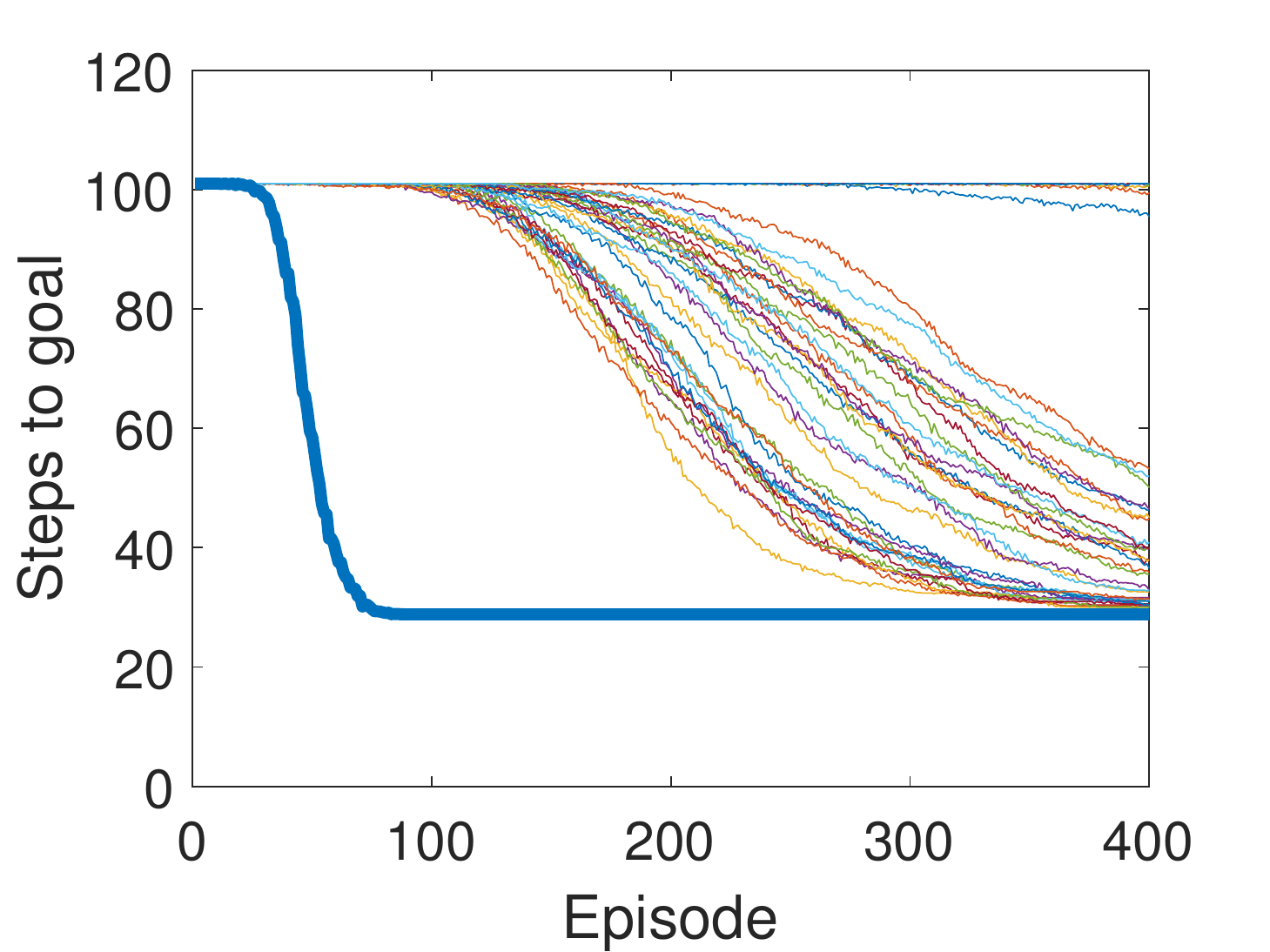}}
\subfigure[]{\includegraphics[width=.30\columnwidth]{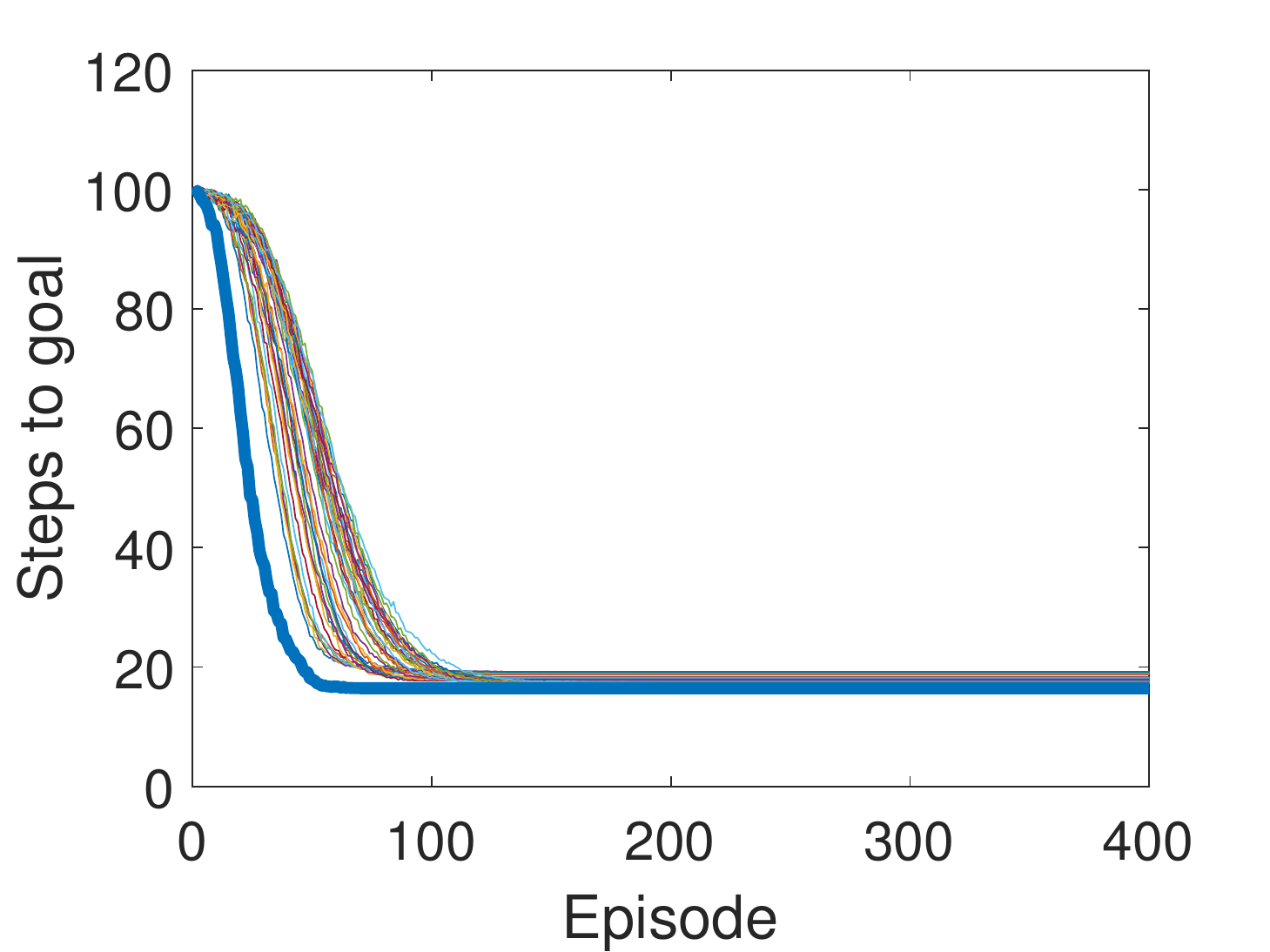}}
\caption{Learning performance of the diffusion options using $t=4$ (bold line) and different number of eigenoptions ranging between 10 and 50. (a) The Ring domain. (b) The Maze domain. (c) The 4Rooms domain. Due to the multiple curves, the standard deviation is omitted in this figure, but we report that it is similar to the standard deviation reported in Fig. 2 in the paper.}
\label{Fig:VarietyEigenoptionsDiffusionOptAllDomains}
\end{figure}

In the paper, we test the learning performance of the diffusion options for $t=4$, yielding $32$, $19$, and $20$ options for the Ring, Maze and 4Rooms domains, respectively. There, we compare it to the same number of eigenoptions. To complement this comparison, here we compare the performance of the diffusion options to different number of eigenoptions, ranging between 10 and 50, for all three domains. We use the same Q learning setting as described in section 4 in the paper. The results appear in Figure \ref{Fig:VarietyEigenoptionsDiffusionOptAllDomains}. We observe that in all the domains the diffusion options achieve faster convergence compared to the different number of eigenoptions. In other words, the diffusion options are superior even if the number of eigenoptions is tuned to attain maximal performance.

In Table 1 in the paper we reported the median and the interquartile range (IQR) of the number of steps between any pair of states. Here, in Table \ref{Table:StepsStatesMean}, we complete the picture and present the mean and standard deviation. Broadly, we observe similar trends, where the diffusion options obtain the best results in all three domains.
\setcounter{table}{1}
\begin{table*}[t]
\caption{Number of steps between any pair of states using options induced by $t=4$ and by $t=13$. Here we report the mean value and the stadard deviation.}
\label{Table:StepsStatesMean}
\begin{center}
\begin{tabular}{*{9}{c|}}
\multicolumn{1}{c|}{\bf Domain (\#states)}
&\multicolumn{1}{c|}{\bf t}
&\multicolumn{1}{c|}{\bf \#options}
&\multicolumn{2}{c|}{\bf Diffusion Options }
&\multicolumn{2}{c|}{\bf Eigenoptions }
&\multicolumn{2}{c|}{\bf Random Walk }
\\ 
                      &  &  &  Mean  & STD       &  Mean    & STD       &  Mean   &STD \\
\hline
\multirow{2}{*}{Ring (192)}     & 4 & 32  &  228       & 74                 &  327    & 139              &  618   & 134\\
                                & 13 & 28 &  249       & 149                 &  326    & 149              &  618  & 134\\
\hline
\multirow{2}{*}{Maze (148)}     & 4  & 19 &  296      & 271                  & 551      & 341             & 1323   & 639 \\
                                & 13 & 14 &  240      & 134                   & 1393      & 3260             & 1323   & 639 \\
\hline
\multirow{2}{*}{4Rooms (104)}  & 4  & 20 &   158     & 81                    &  183   & 88                 & 497       & 80\\
                                 & 13 & 15 &  155     & 97                       &  193   & 120                & 497    & 80 \\
\end{tabular}
\end{center}
% \label{Table:StepsStates}
\end{table*}

\subsection{The Scale Parameter $t$}

The scale parameter $t$ demonstrates a smoothing effect, similar to a low pass filter, on $f_t(s)$. Concretely, $f_t(s)$ displays fewer peaks, which results in a fewer number of diffusion options, as $t$ increases.

\begin{figure*}
\begin{center}
 \includegraphics[width=0.5\columnwidth]{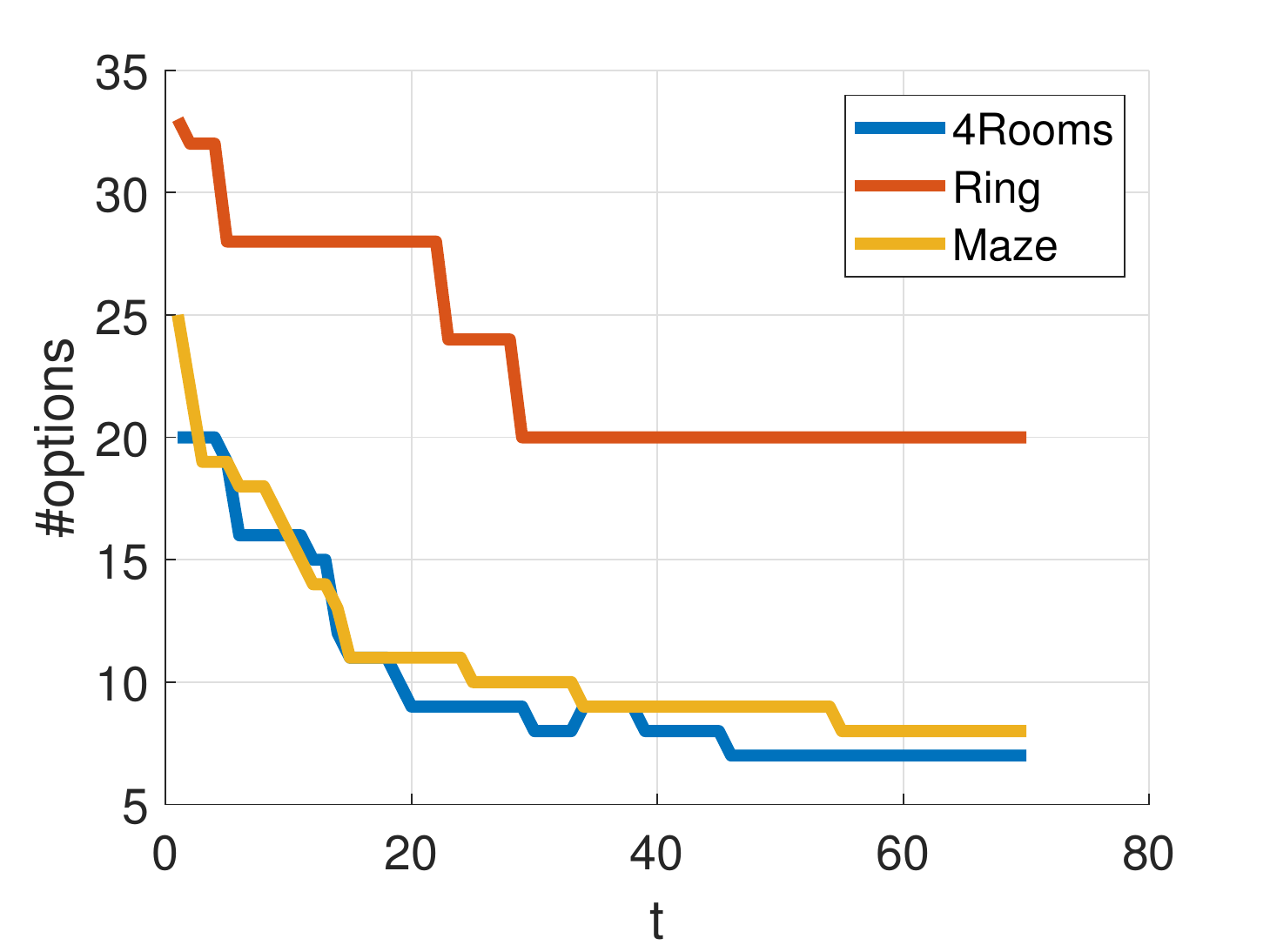}
 \end{center}
  \caption{The number of discovered options for different scale parameter $t$ for all 3 domains.}
    \label{Fig:KOptionsVsTime}
\end{figure*}

Figure \ref{Fig:KOptionsVsTime} shows the number of options derived from the local maxima of $f_t(s)$ as a function of $t$. First we observe that indeed the number of options is decreasing with $t$.
Second, we observe that above a certain value of $t$ (e.g., $t=30$ in the Ring domain), the number of options empirically reaches a steady state. 

\begin{figure}[t]
\centering 
\subfigure[]{\includegraphics[width=.19\columnwidth]{Pics/RingStartStateGoalState-eps-converted-to.pdf}}
\subfigure[]{\includegraphics[width=.16\columnwidth]{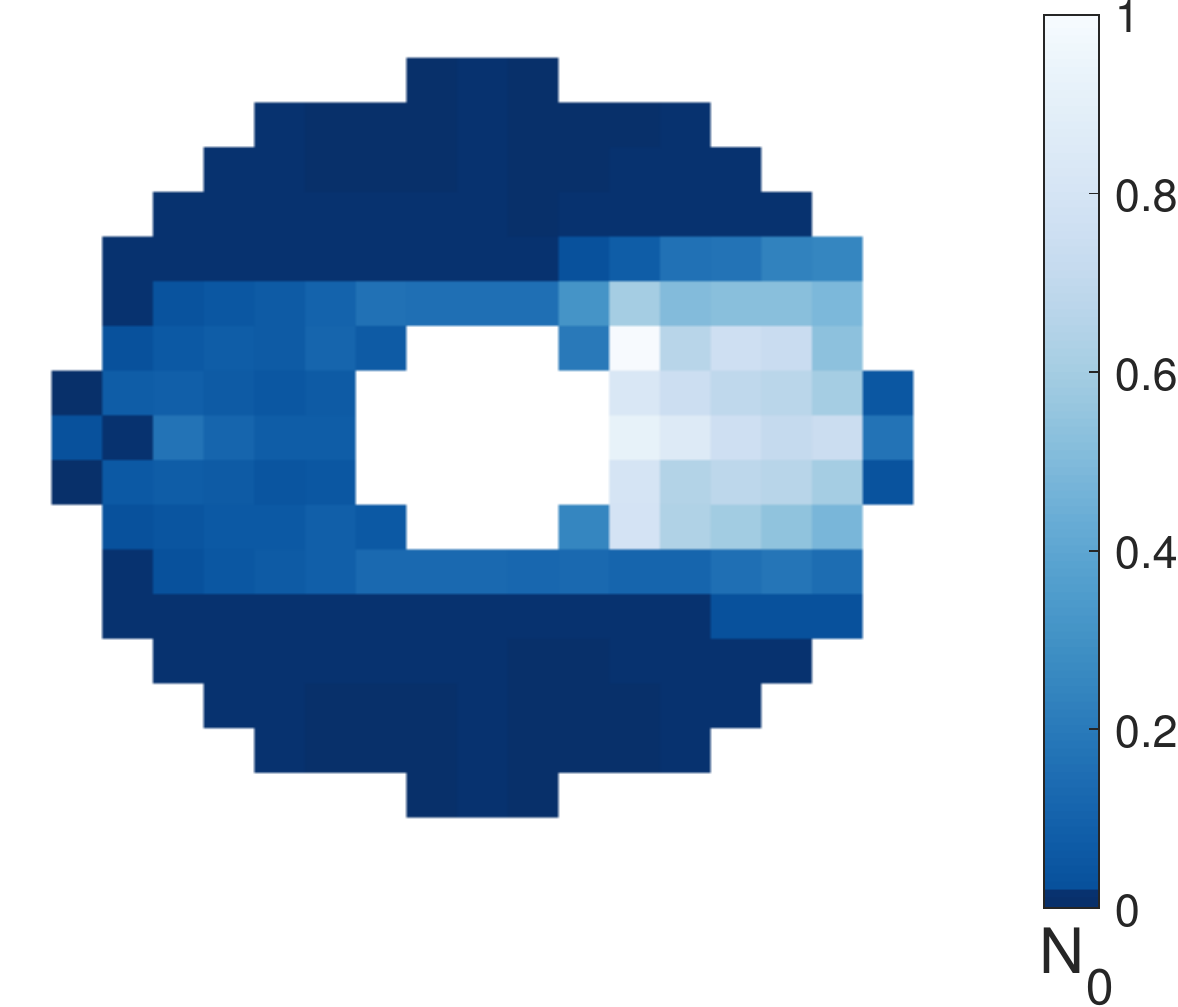}}
\subfigure[]{\includegraphics[width=.16\columnwidth]{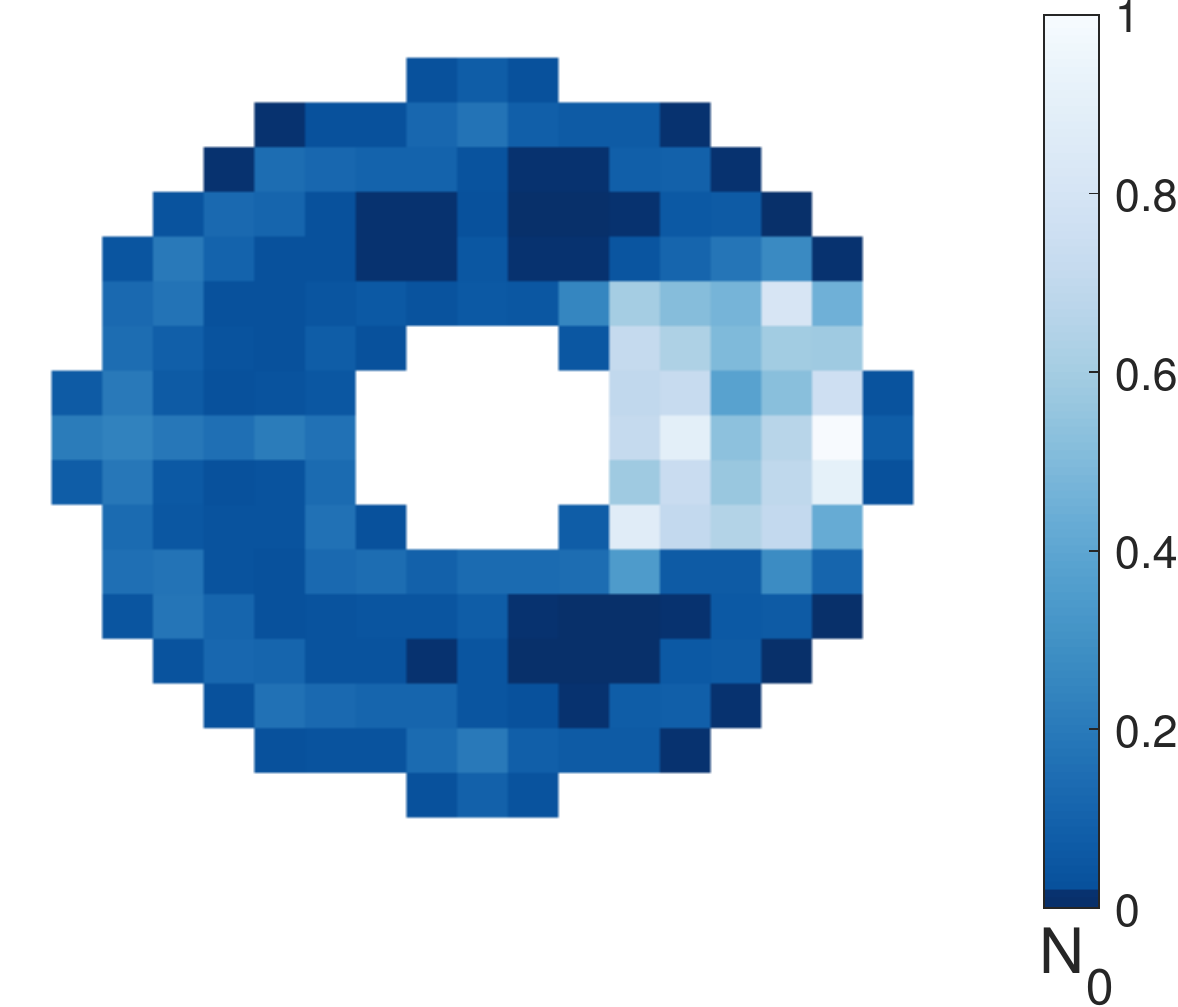}}
\subfigure[]{\includegraphics[width=.16\columnwidth]{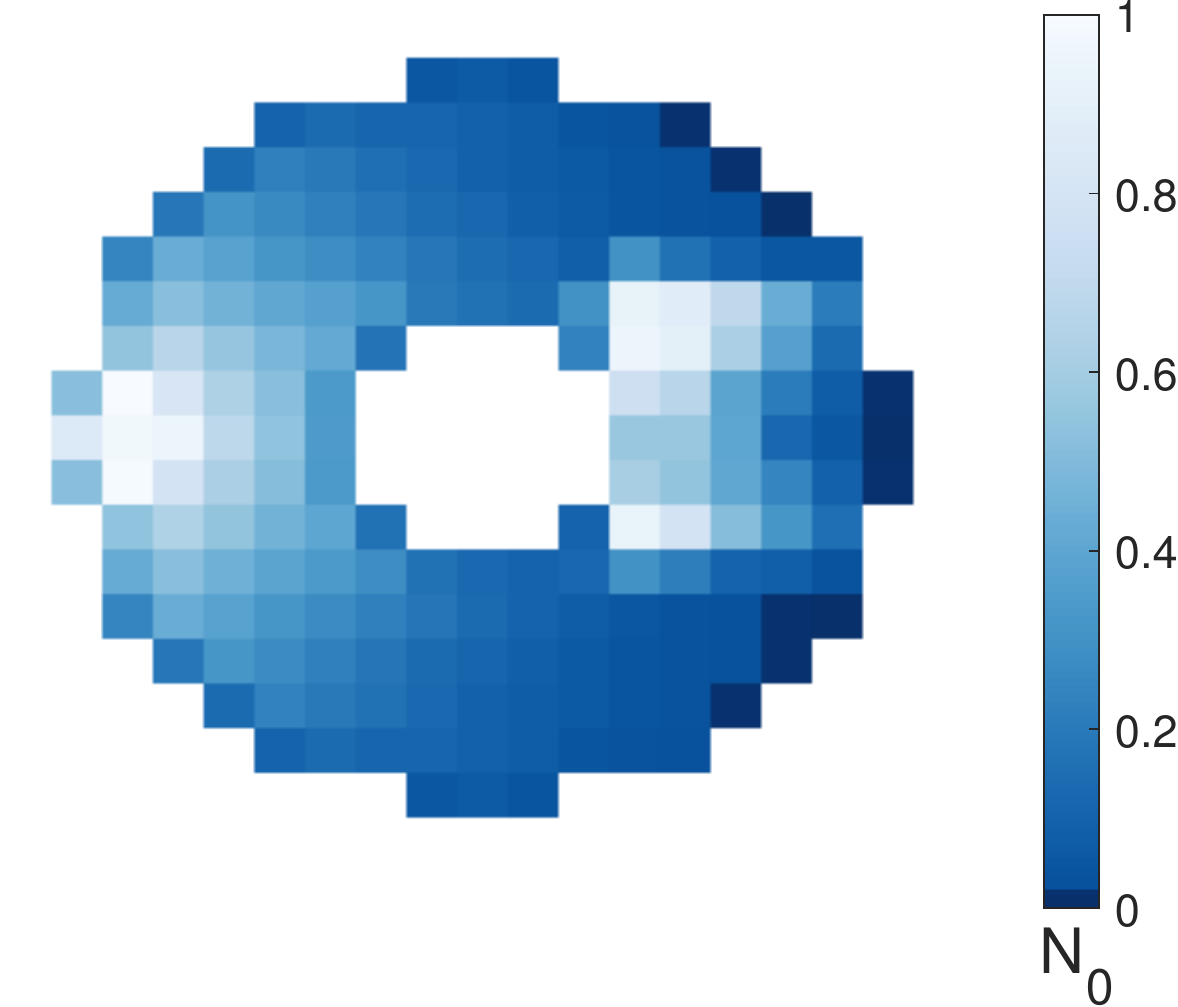}}
\subfigure[]{\includegraphics[width=.19\columnwidth]{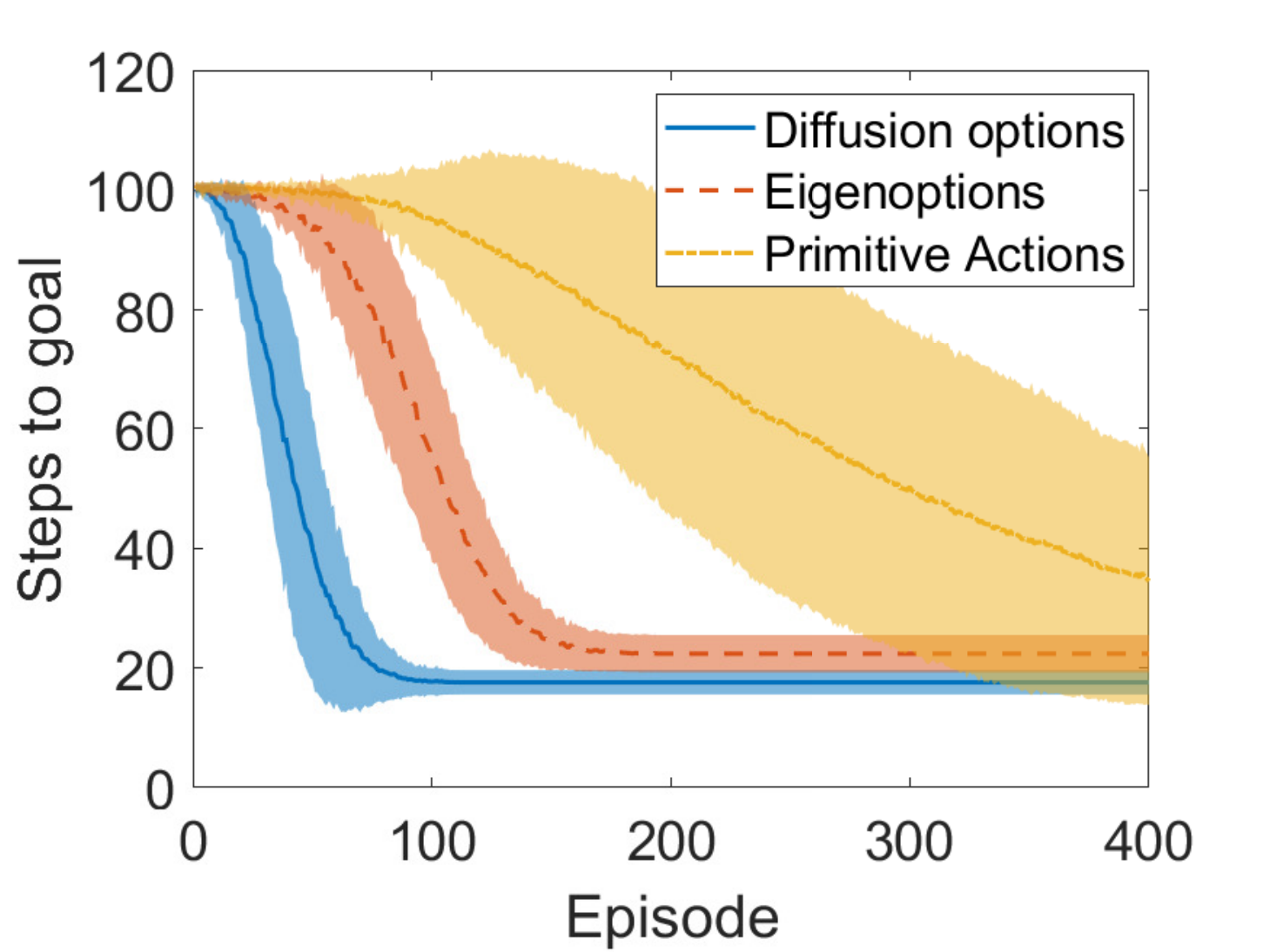}}
%maze
\subfigure[]{\includegraphics[width=.19\columnwidth]{Pics/MazeStartStateGoalState-eps-converted-to.pdf}}
\subfigure[]{\includegraphics[width=.16\columnwidth]{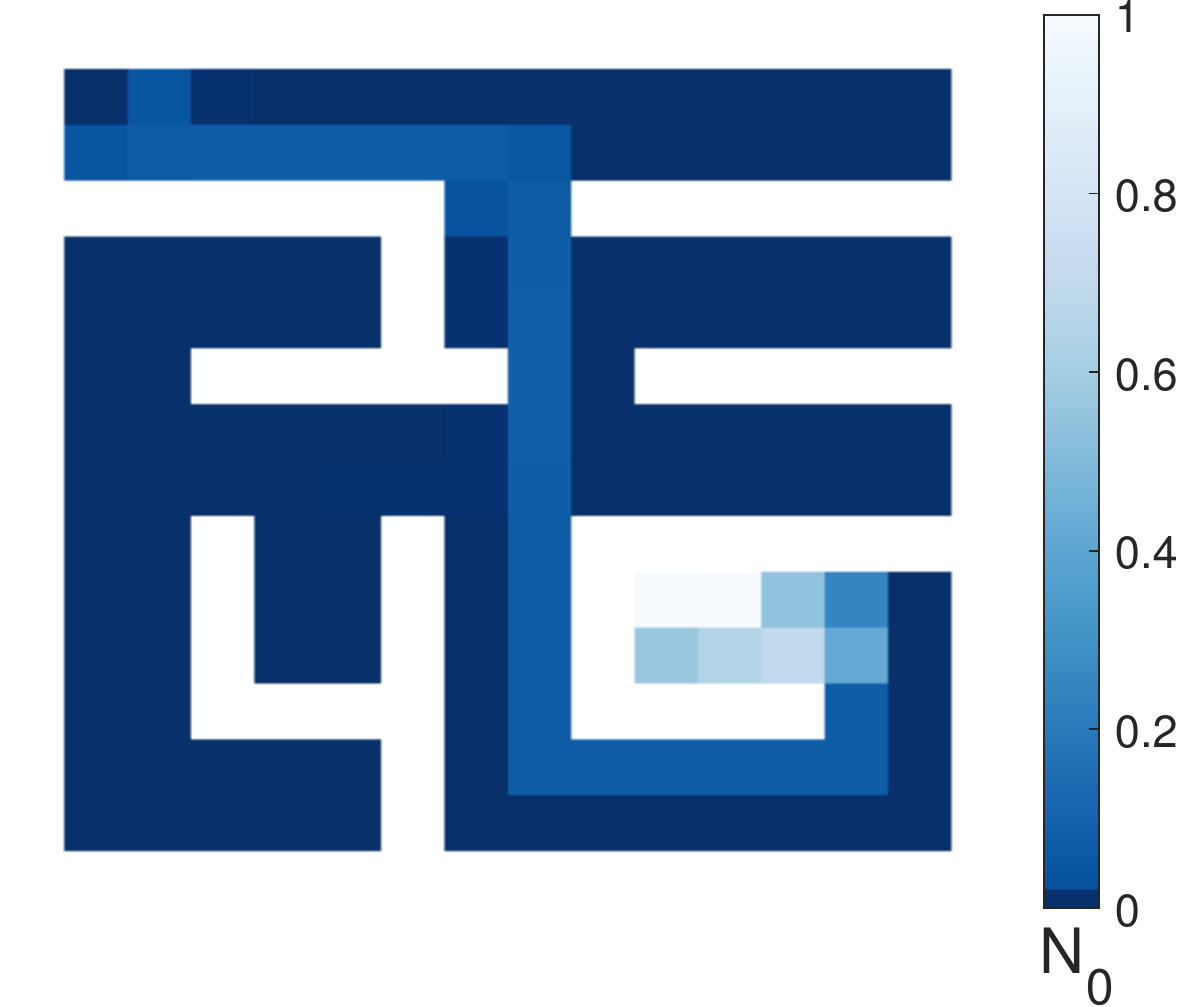}}
\subfigure[]{\includegraphics[width=.16\columnwidth]{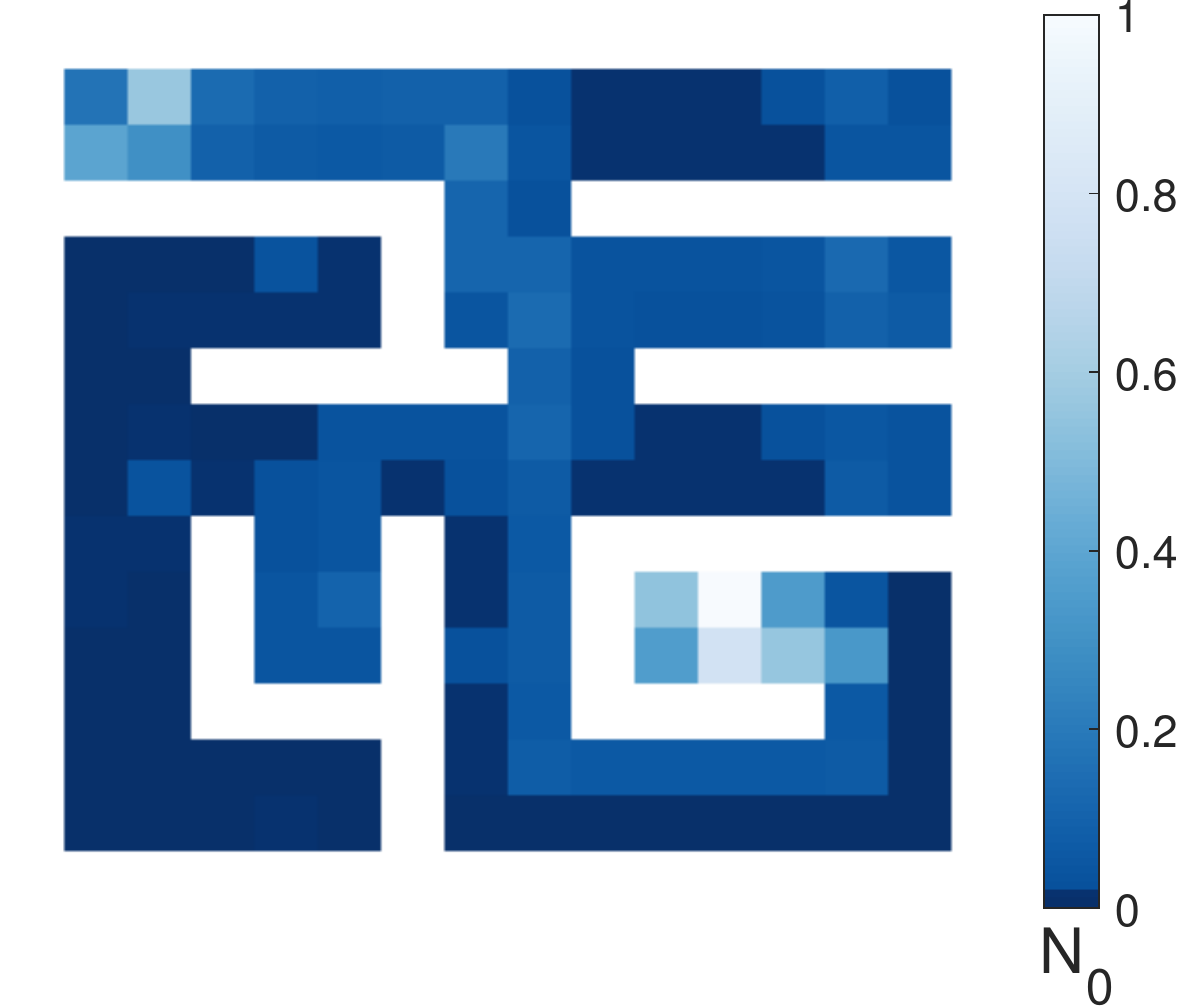}}
\subfigure[]{\includegraphics[width=.16\columnwidth]{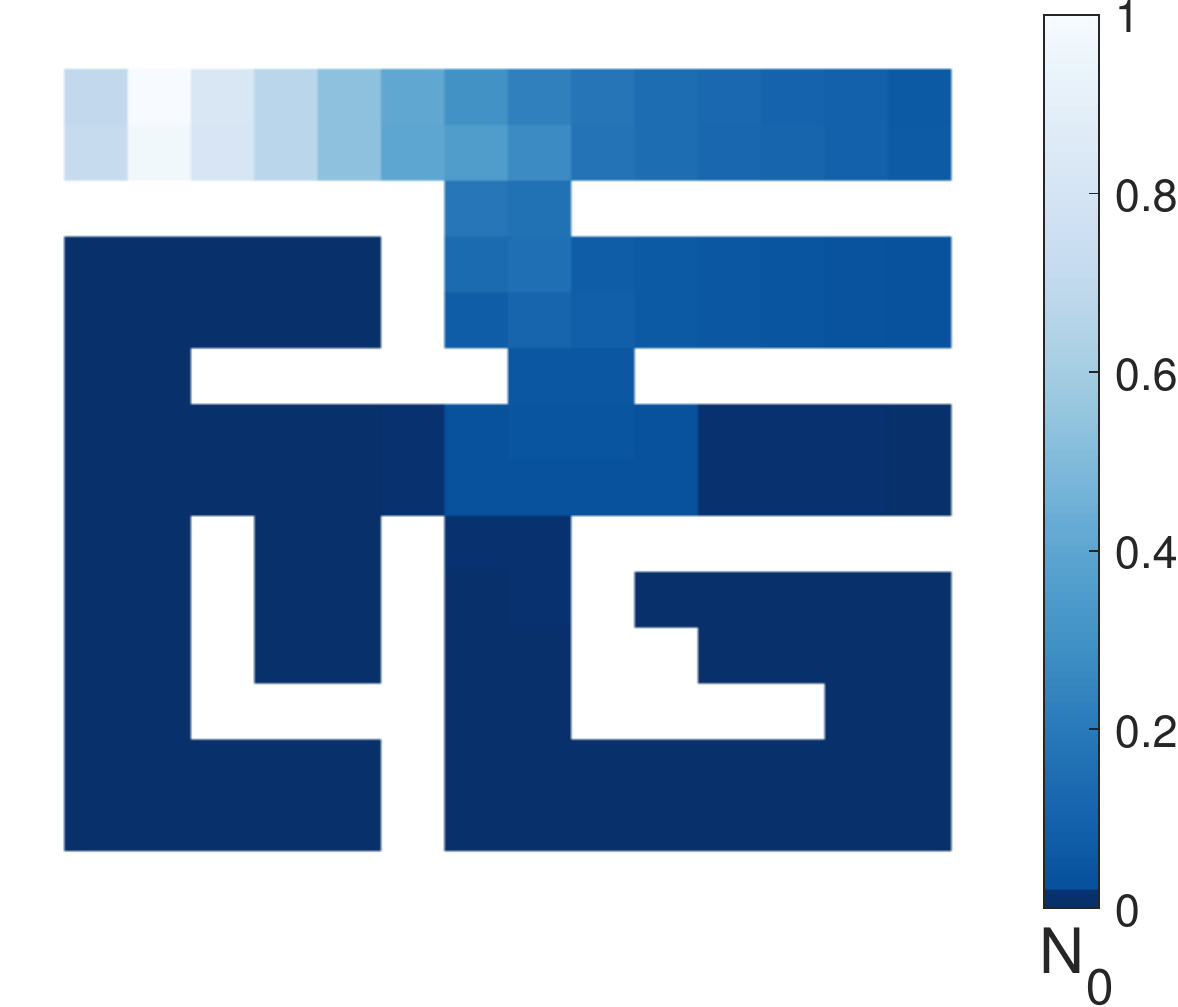}}
\subfigure[]{\includegraphics[width=.19\columnwidth]{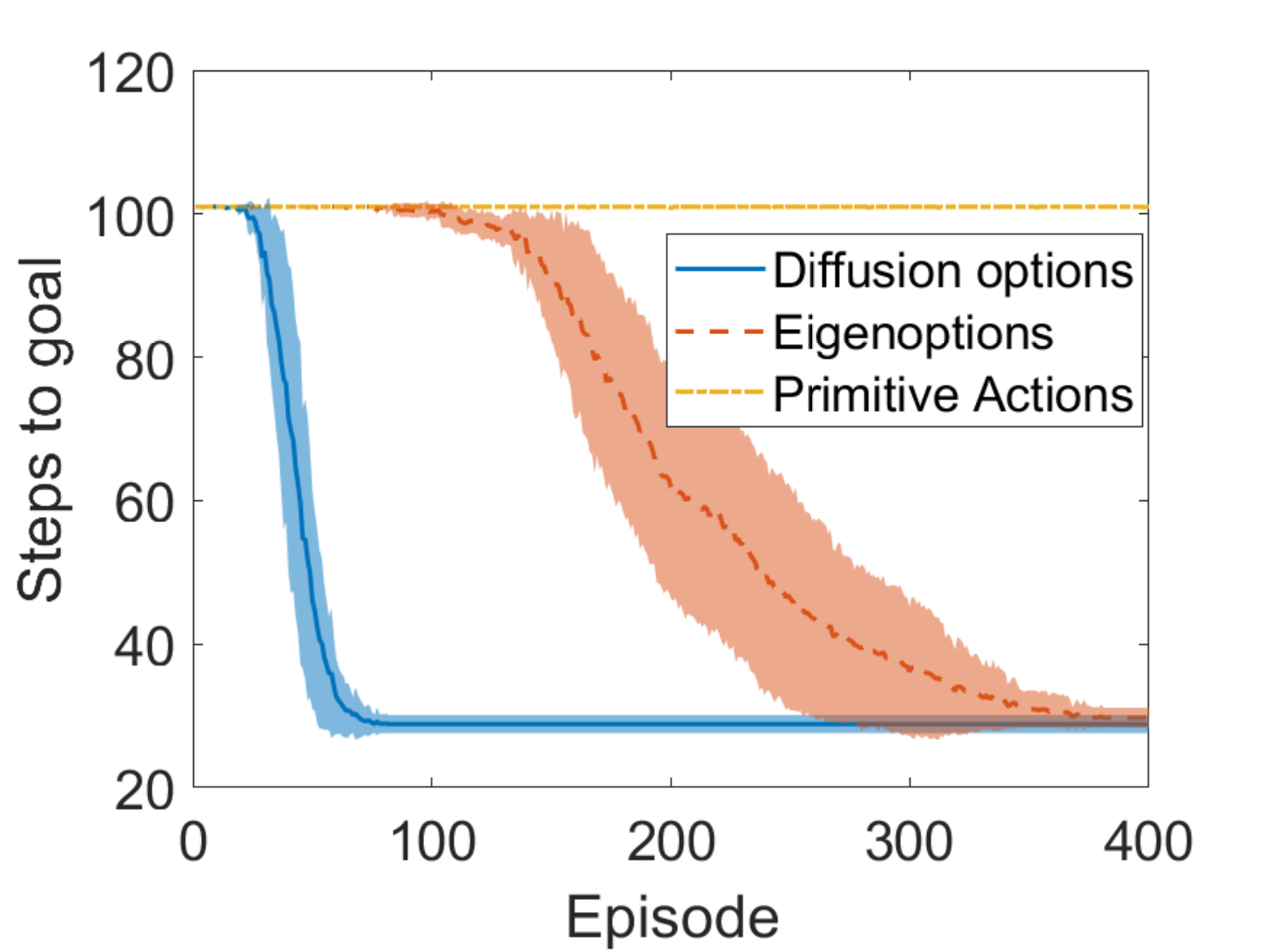}} 
%4Rooms
\subfigure[]{\includegraphics[width=.19\columnwidth]{Pics/4RoomsStartStateGoalState-eps-converted-to.pdf}} 
\subfigure[]{\includegraphics[width=.16\columnwidth]{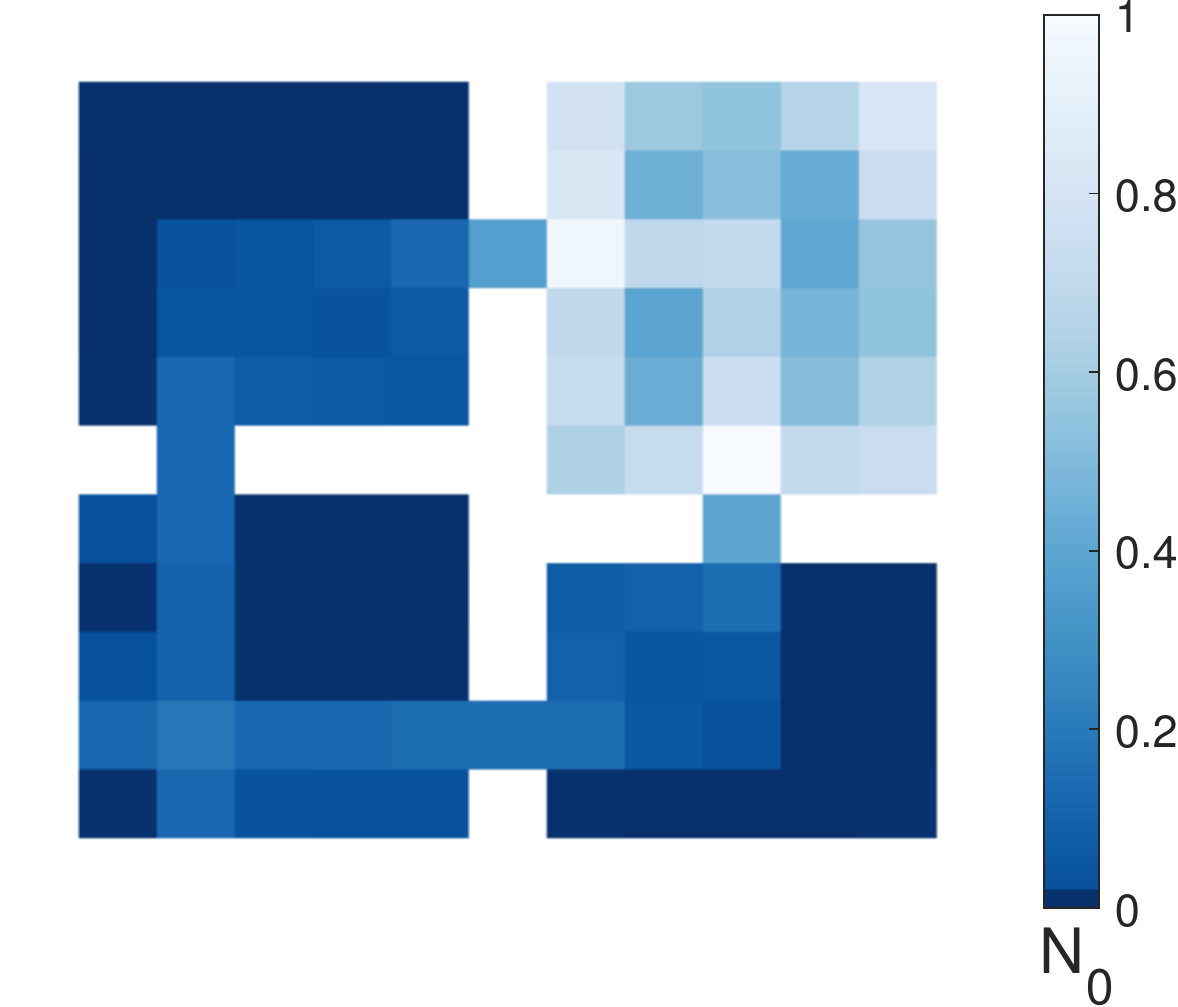}} 
\subfigure[]{\includegraphics[width=.16\columnwidth]{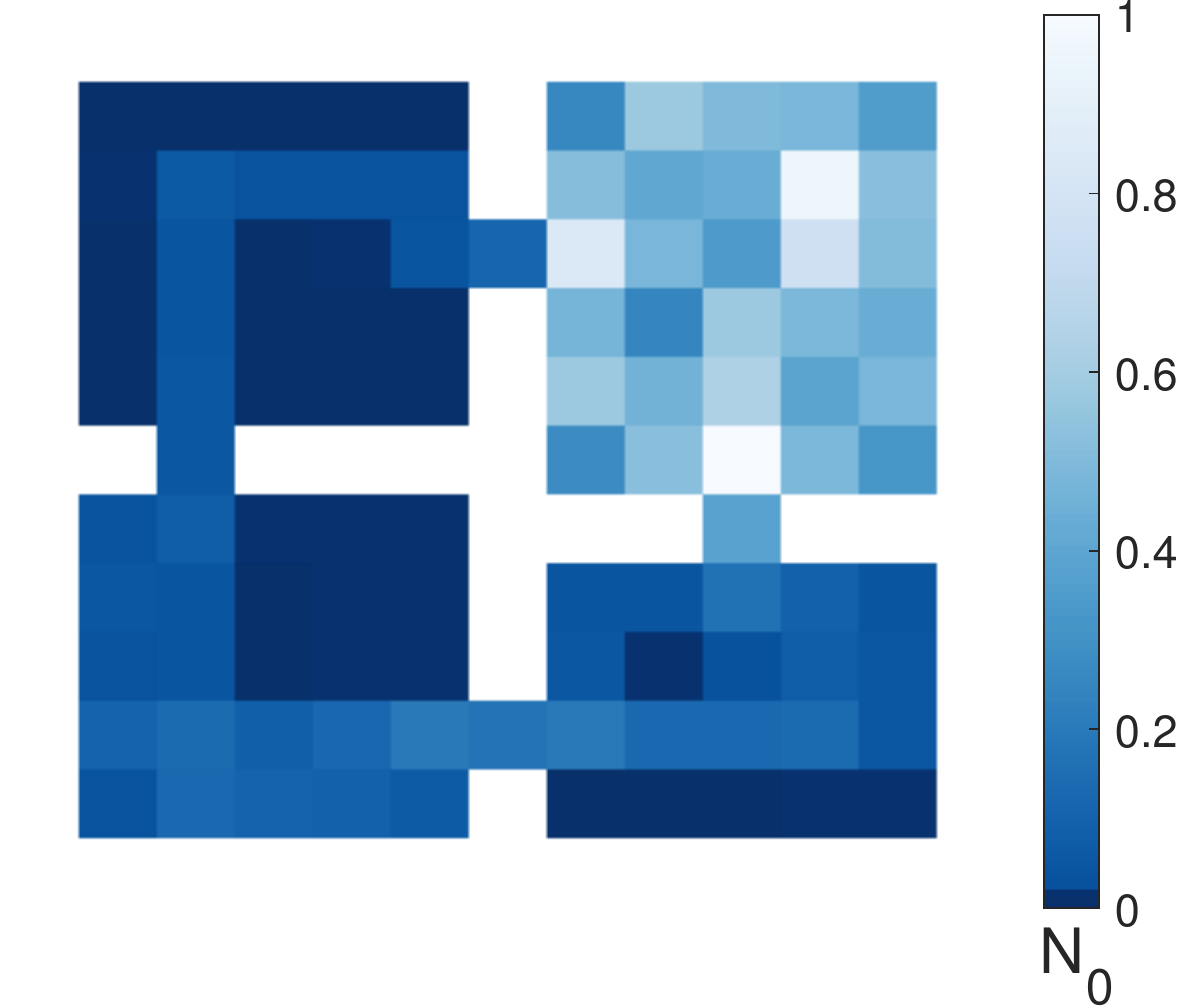}} 
\subfigure[]{\includegraphics[width=.16\columnwidth]{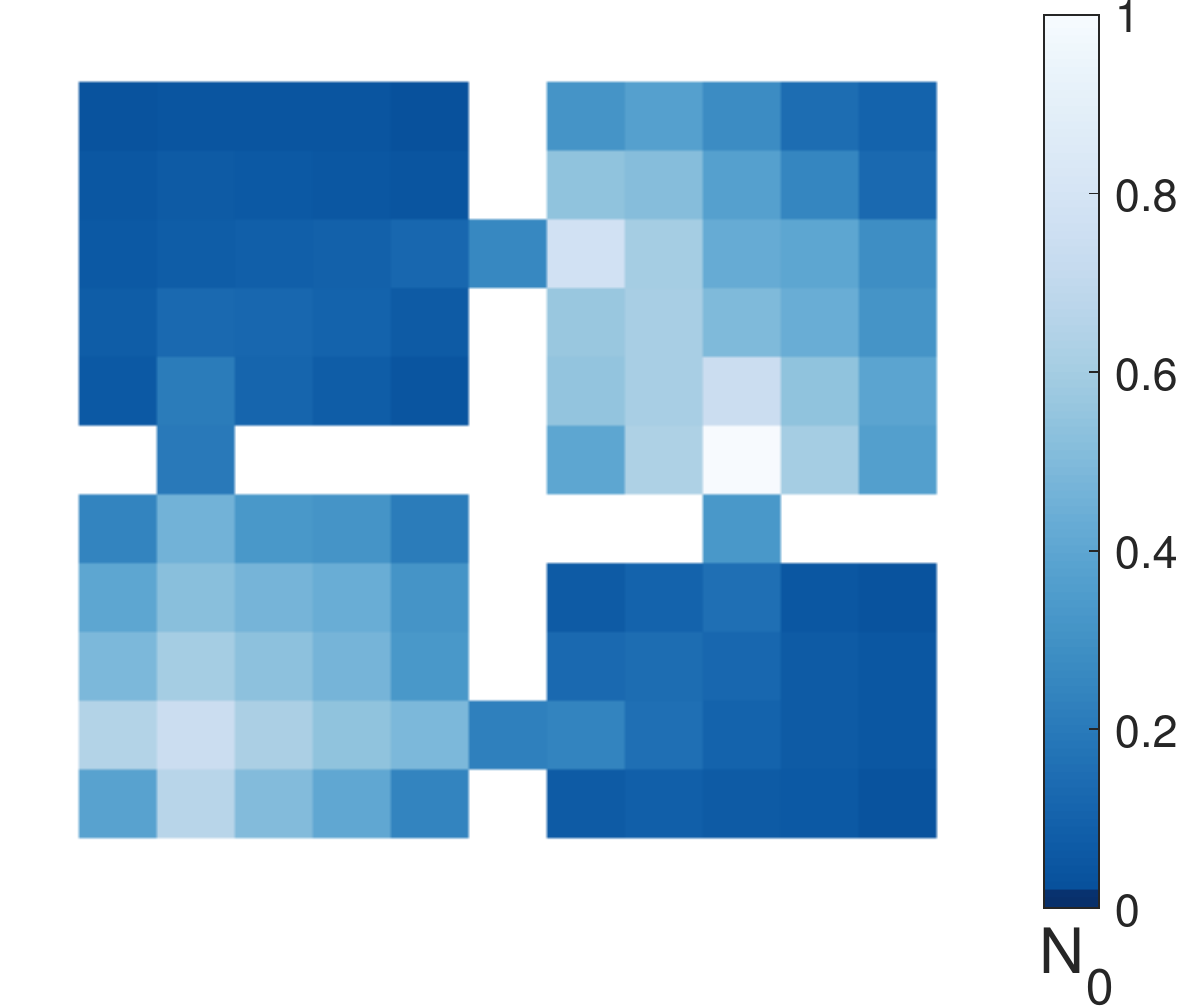}} 
\subfigure[]{\includegraphics[width=.19\columnwidth]{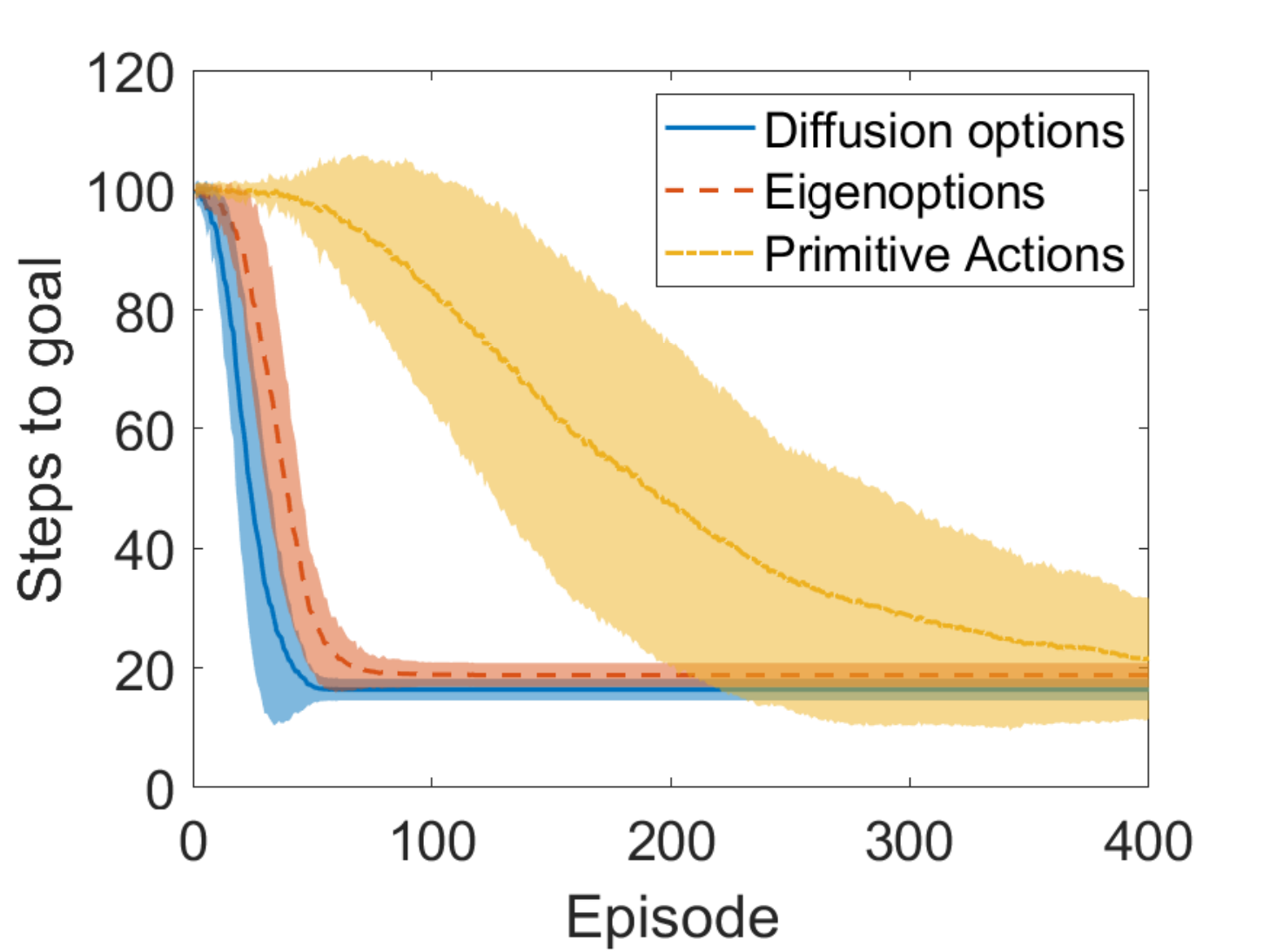}} 
\caption{Learning results obtained by setting $t=13$ in the Ring domain (top row), the Maze domain (middle row), and 4Rooms domain (bottom row). (a,f,k) The start state (green) and goal states (purple). (b-d,g-i,l-m) Normalized visitation count $N_0$ obtained based on (b,g,l) the diffusion options, (c,h,m) the eigenoptions, and (d,i,n) a random walk (d). For visualization purposes, the visitation number is normalized to the range of $[0,1]$ by dividing by the maximum number of visitations. (e,j,o) The learning convergence depicting the average number of steps to goal for each learning episode. The solid line represents the mean value and the light colors represent the standard deviation.}
\label{Fig:RingDomainQlearn_tEq13}
\end{figure}

In the paper, we show the performance of the diffusion options derived from $f_t(s)$ by setting $t=4$. Here, we show results obtained by setting $t=13$. Figure \ref{Fig:KOptionsVsTime} implies that $t=13$ results in fewer options than $t=4$ due to the smoothing effect. 
Figure \ref{Fig:RingDomainQlearn_tEq13} presents the results in the Ring, Maze and 4Rooms domains, measured by the learning convergence and the normalized visitation count during the learning process. 
Broadly, we observe that the obtained results for $t=13$ are similar to those reported in the paper for $t=4$, where the diffusion options are superior in comparison to the eigenoptions in the three tested domains. We report that similar results were obtained for other $t$ values as well, suggesting that the algorithm is not very sensitive to the particular value of $t$.

\subsection{Comparison with Random Options}
n order to highlight the importance of the option goal states derived in the proposed method, we compare the results to random option goal states.
To this end, we first randomly choose states. Then, we generate ``random options'', where each random option is associated with one random goal state. We implement these random options similarly to the diffusion options, so that their policies lead the agent to their respective random goal state via the shortest path.

We compare the performance of the diffusion options, the eigenoptions, and the random options in the 4Rooms domain, where the agent starts at the bottom left corner and its goal is to reach the top right corner. We implement Q learning (as in the paper) and compute the number of steps until the goal state is reached at each episode during the learning process. We repeat this test for $100$ Monte Carlo iterations. At each iteration a different set of random goal states are uniformly chosen, resulting in a different set of random options (while the diffusion options and the eigenoptions remain the same). We use $20$ options (derived by setting $t=4$) from each option kind. 

\begin{figure*}
\begin{center}
 \includegraphics[width=0.5\columnwidth]{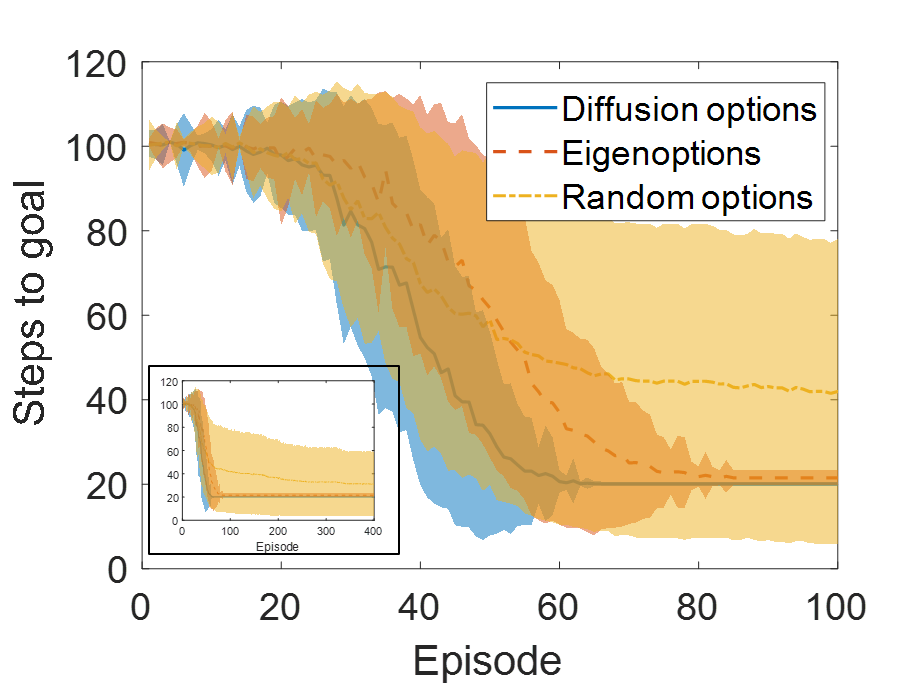}
 \end{center}
  \caption{Comparing the learning convergence to random options in the 4Rooms domain. 
  The learning convergence depicting the average number of steps to goal for each learning episode. The solid line represents the mean value and the light colors represent the standard deviation.
  The start state is the bottom left corner, and the goal state is at the top right corner. The inset presents the same plot for $400$ episodes.}
    \label{Fig:QLearn4Rooms1GoalState}
\end{figure*}

Figure \ref{Fig:QLearn4Rooms1GoalState} presents the learning performance for $100$ episodes, allowing to compare the learning rates, and an inset for $400$ episodes, including the learning convergence.
We observe that the diffusion options lead to the fastest convergence compared to the eigenoptions and the random options. Random options have a large learning variance and a slow convergence. Even after $300$ episodes, the random options' performance demonstrates relatively high variance, suggesting that their performance highly depends on the randomly chosen set of goal states.

\subsection{Stochastic Domain}

In a deterministic domain, the considered graph is undirected, where the adjacency matrix $\mM$ merely conveys the (binary) connectivity between states, and thereby,  is symmetric.
Conversely, in a stochastic domain, the connectivity between states is more evolved, depending on the transition probabilities induced by the stochasticity of the domain.
For example, in a domain with wind blowing downward, the agent experiences more transitions from states located at the top to states located at the bottom, than vice versa.
As a result, the corresponding graph is directed and the corresponding matrix $\mM$ is non-symmetric.
This requires slight modifications in Algorithm 1 in the paper, which are detailed below. 

Let $\mM$ be the weight matrix of a directed graph. Such a matrix could represent either the true transitions or the empirical transitions experienced by the agent in some exploration phase.
Forming the normalized graph Laplacian $\mN=\mI-\mD^{-\frac{1}{2}}\mM\mD^{-\frac{1}{2}}$ , where $\mD$ is the corresponding degree matrix, results in a non-symmetric matrix, to which we cannot apply EVD directly, as prescribed in Algorithm 1 in the paper. Instead, following \citet{mhaskar2018unified}, we use the polar decomposition $\mN=\mR\mU$, and apply EVD to $\mR$ , obtaining its eigenvectors $\{\tilde{\psi}_i\}$ and eigenvalues $\{\tilde{\nu}_i\}$. Then, by the relations in (\ref{eq:OmegaNuRelations}), $\{\phi_i\}$,$\{\tilde{\phi}_i\}$, and $\{\tilde{\omega}_i\}$ are computed, from which $f_t(s)$ is constructed. 
The entire modified algorithm is presented in Algorithm \ref{Alg:OptionsDiscoveryStochastic}. 

\begin{figure}[t]
\begin{center}
\subfigure[]{\includegraphics[width=0.3\textwidth]{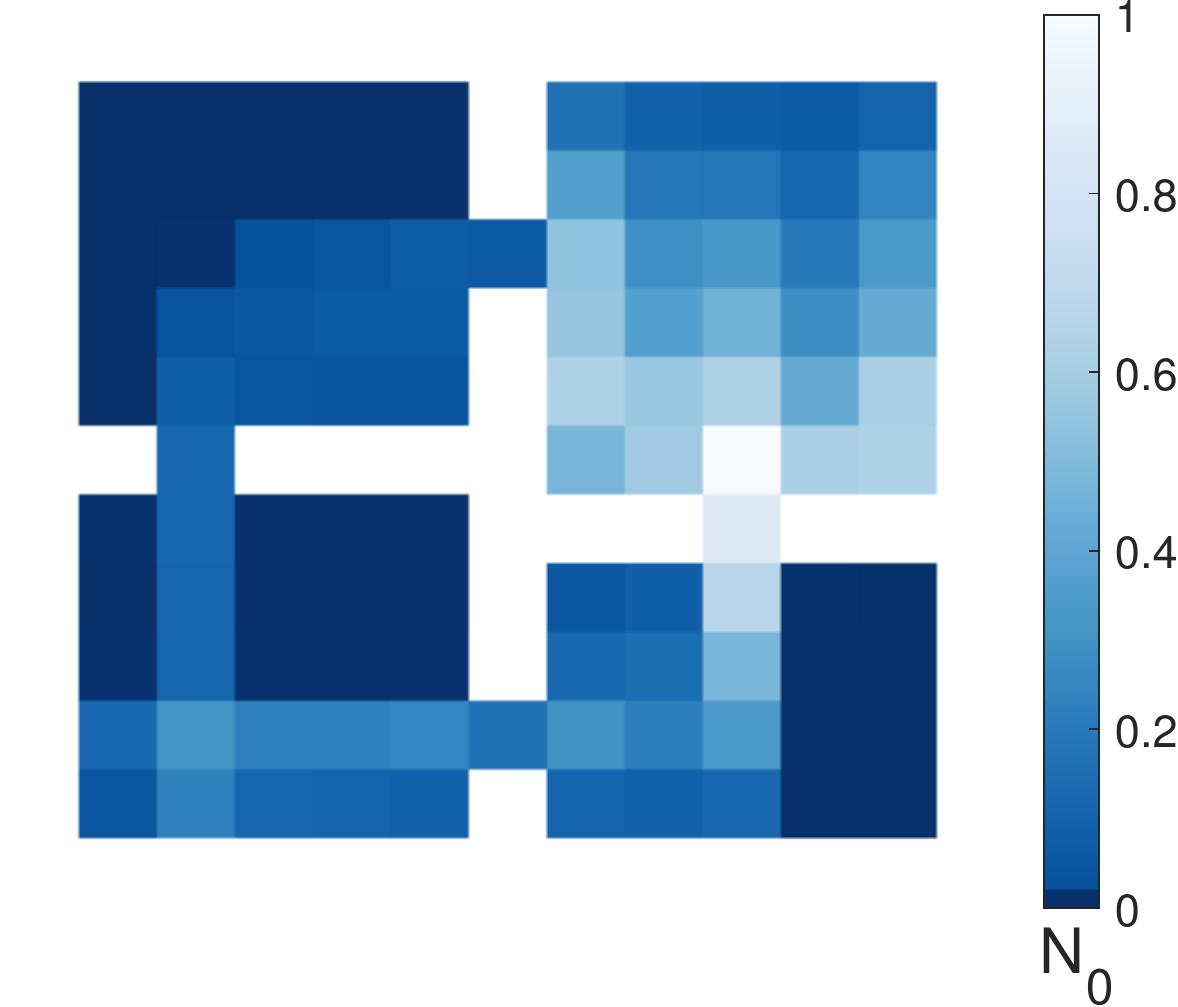}}
\subfigure[]{\includegraphics[width=0.3\textwidth]{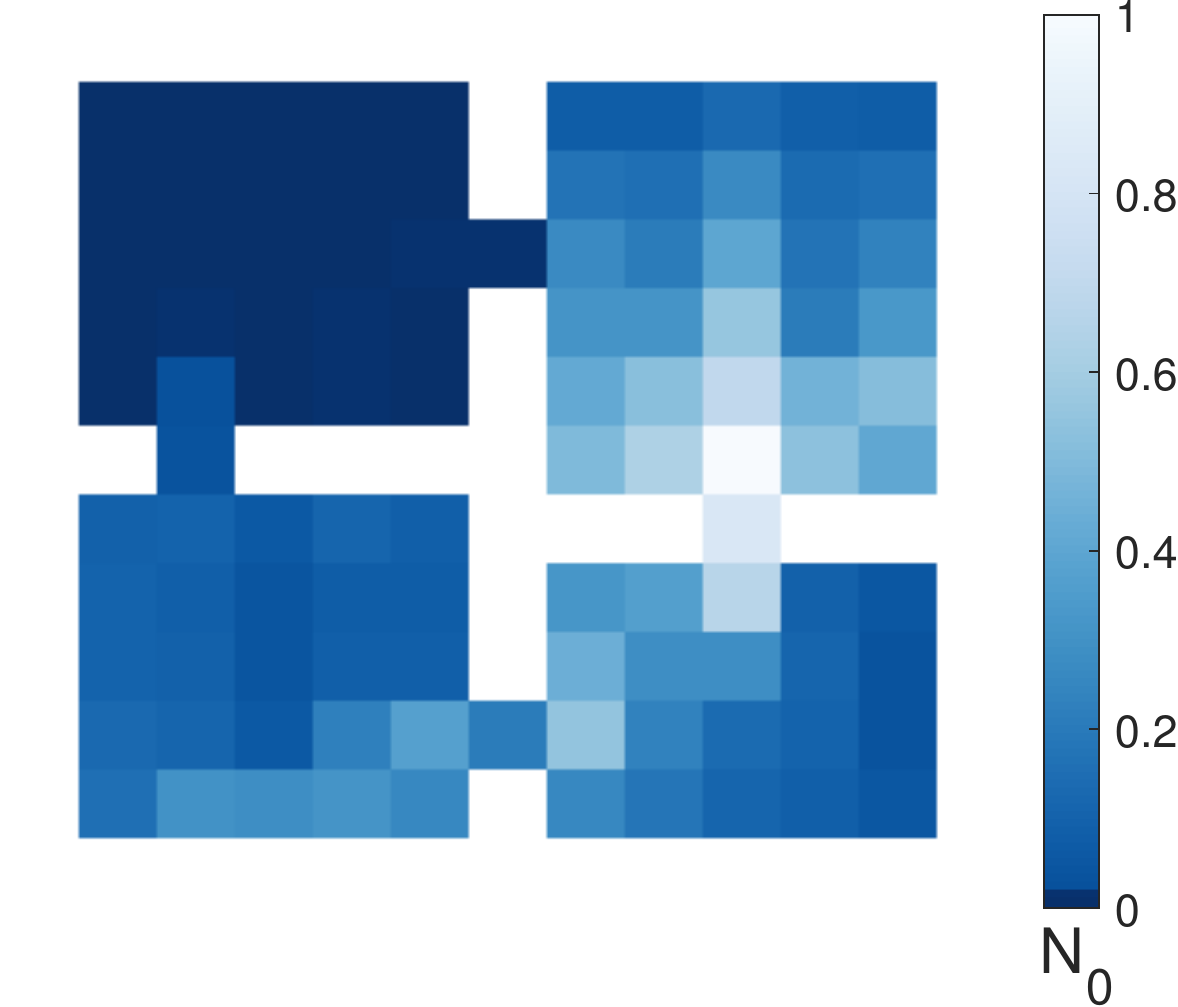}}
\subfigure[]{\includegraphics[width=.3\textwidth]{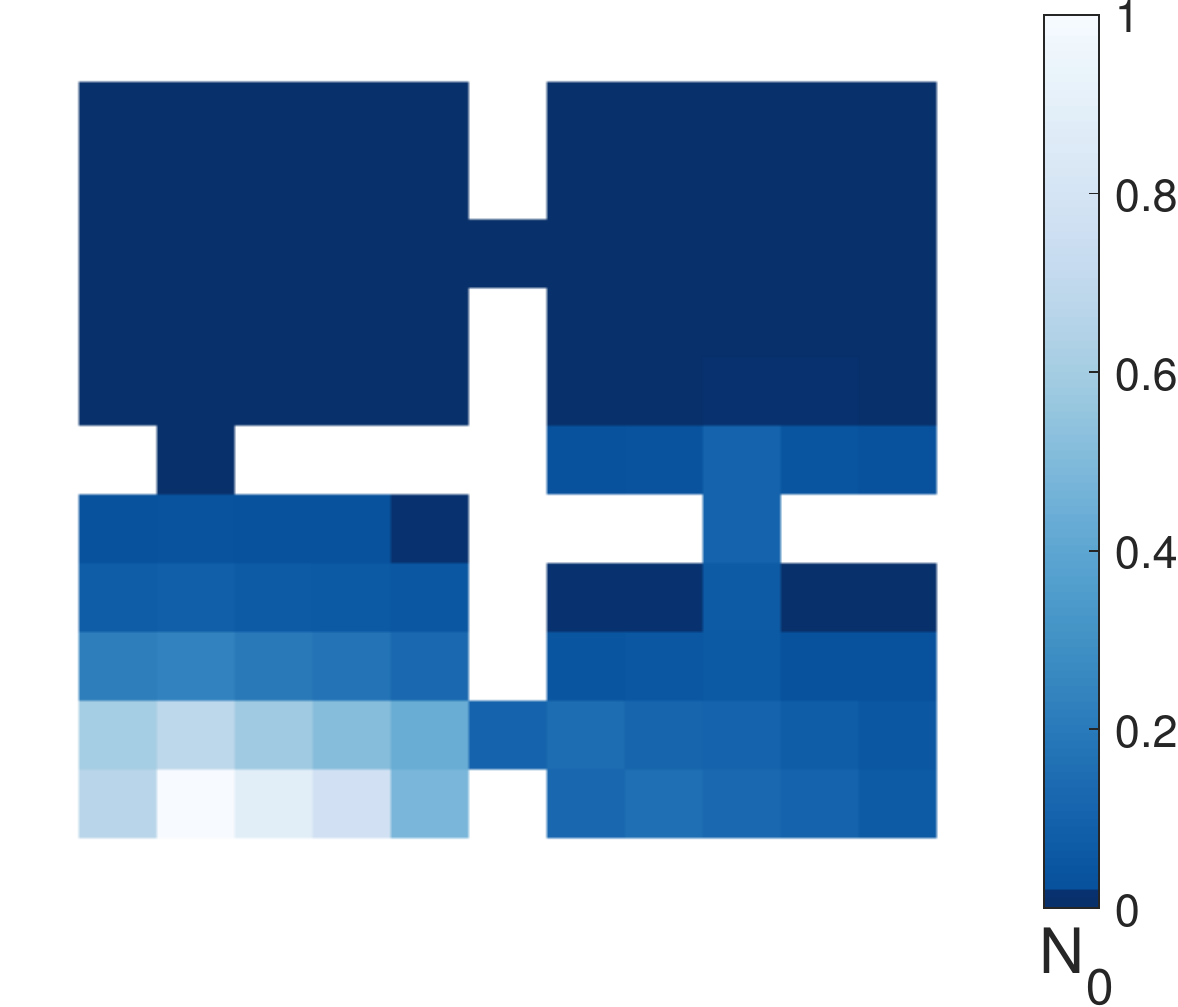}}
\caption{Normalized visitation count $N_0$ in the 4Rooms domain with stochastic wind blowing downward obtained based on (a) the diffusion options, (b) the eigenoptions, and (c) a random walk.}
\label{Fig:4Rooms_Stochastic_Visitaions_tEq4}
\end{center}
\end{figure}

We note that this modified version is also applicable as is to deterministic domains with asymmetric transitions between states. We observe that Algorithm 1 in the paper and Algorithm \ref{Alg:OptionsDiscoveryStochastic} here are very similar, and differ only in the matrix whose spectrum is used: $\mN$ in Algorithm 1, and $\mR$ in Algorithm \ref{Alg:OptionsDiscoveryStochastic}. 
\setcounter{algorithm}{1}
\begin{algorithm} 
    \caption{Diffusion Options for Stochastic Domains}
    \label{Alg:OptionsDiscoveryStochastic}
     \textbf{Input:} Adjacency matrix $\mM$ and scale parameter $t>0$
     
     \textbf{Output:} $K$ options with policies $\{\pi_o^{(i)}\}_{i=1}^{K}$ 

    \begin{algorithmic}[1] 
     \STATE Compute the degree matrix $\mD$ from $\mM$
     \STATE Compute the normalized graph Laplacian $\mN$ using (\ref{eq:Definition_N})
     \STATE Compute the polar decomposition $\mN=\mR\mU$ 
     \STATE Apply EVD to $\mR$ and obtain its eigenvectors \{$\Tilde{\psi}_i$\} and eigenvalues $\{\tilde{\nu}_i\}$
     \STATE Compute \{$\phi_i$\}, \{$\Tilde{\phi}_i$\} and $\{\omega_i\}$ using (\ref{eq:OmegaNuRelations}) with \{$\Tilde{\psi}_i$\} and $\{\tilde{\nu}_i\}$  
     \STATE Construct $f_t(s)= \Vert\sum_{i\ge{2}}\omega_i^t\phi_i\left(s\right)\Tilde{\phi}_i\Vert^2$ 
    
    \STATE Find the local maxima of $f_t(\cdot)$ -- $\{s_o^{(i)}\}_{i=1}^{K}$
    % \FOREACH {$s$}
    % \STATE a
    % \ENDFOR
    \FOR{$i \in \{1,\ldots,K\}$}
     \STATE Build an option with policy $\pi_o^{(i)}$ s.t. it leads to $s_o^{(i)}$ 
     \ENDFOR
\end{algorithmic}
\end{algorithm}
We examine the performance in the 4Rooms domain, making it stochastic by adding a wind blowing downward, as described in section 4.3 in the paper.
Figure \ref{Fig:4Rooms_Stochastic_Visitaions_tEq4} depicts the normalized visitations at each state during the learning process.
The effect of the downward wind on the visitation count is most prominent when using the primitive actions, where we observe that the agent rarely visits the top two rooms.
The effect of the wind on the diffusion options and the eigenoptions is milder. Compared to Fig. 2 in the paper, we now observe a tendency to prefer trajectories that go through the right bottom room rather than the left top room. Still, the advantage of the diffusion options compared to the eigenoptions is evident, where the diffusion options lead the agent to the goal via shorter paths.

\begin{figure*}
\begin{center}
 \includegraphics[width=0.5\columnwidth]{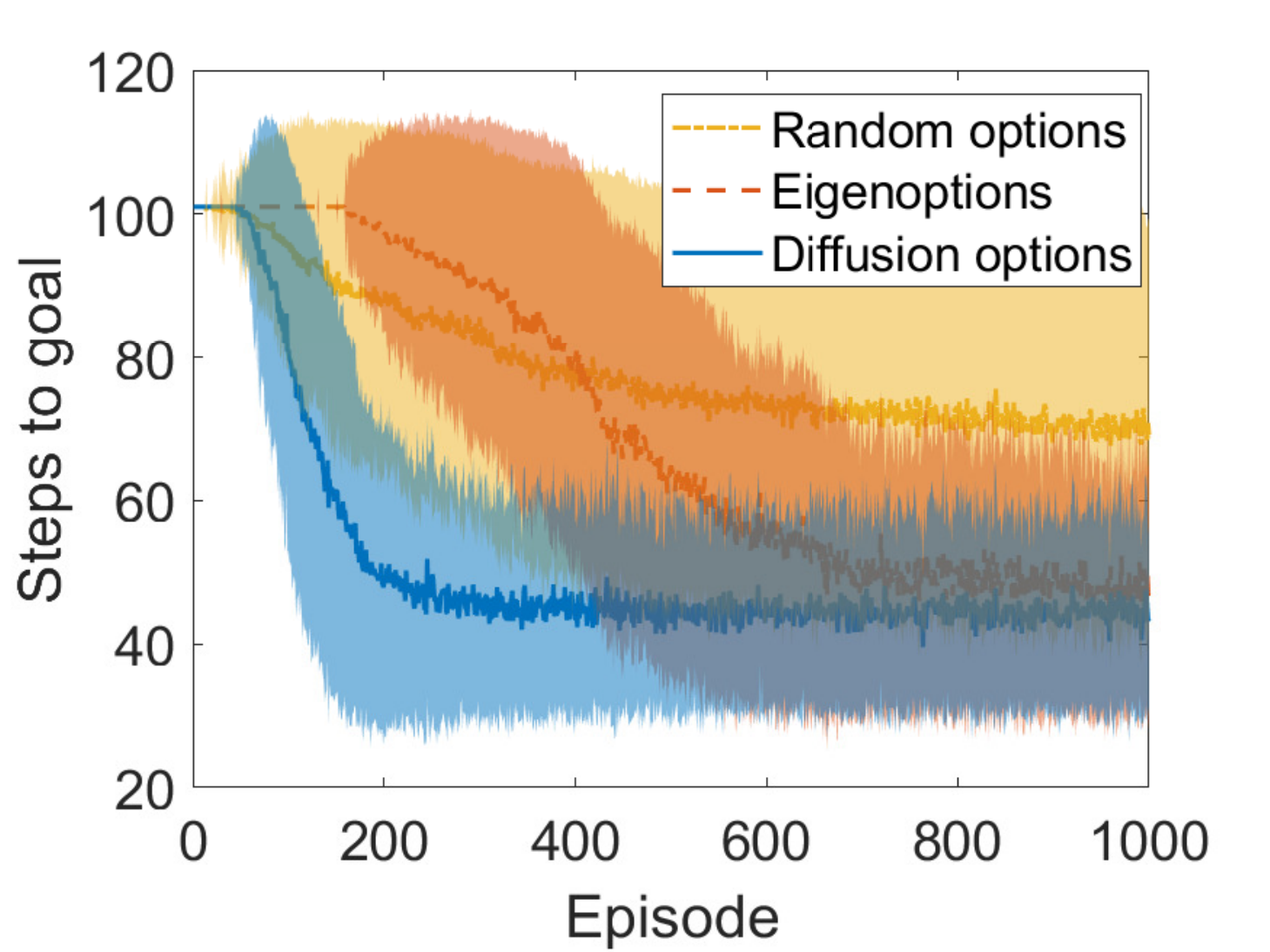}
 \end{center}
  \caption{Learning performance in the 4Rooms domain with stochastic wind. The start state is the bottom left corner and the goal state is the top right corner.}
    \label{Fig:QLearnRandomOptionsStochastDom}
\end{figure*}

Figure \ref{Fig:QLearnRandomOptionsStochastDom} shows the learning performance in the 4Rooms domain with the stochastic wind.
We observe that the diffusion options obtain the fastest convergence and the smallest standard deviation.
In comparison with Fig. \ref{Fig:QLearn4Rooms1GoalState}, we first observe that 
all the options lead to higher number of steps to goal, coinciding with the increase in the difficulty of the problem.
In addition, similar trends appear, where the diffusion options lead to the fastest convergence with the smallest variance.

\section{Extension to Large Scale Domains}
\label{Sec:ScalinkUp}

Assume that the set of states $\mathbb{S}$, which is known a-priori, is merely a subset of the states in the domain $\bar{\mathbb{S}}$, i.e., $\mathbb{S} \subset \bar{\mathbb{S}}$. Then, the question that arises is how to extend $f_t: \mathbb{S} \rightarrow \mathbb{R}$ in equation (1) in the paper from $\mathbb{S}$ to $\bar{\mathbb{S}}$, thereby allowing to discover options on the entire (unseen) domain.  

We consider the following extension $g_t:\bar{\mathbb{S}} \rightarrow \mathbb{R}$
\begin{equation}
	g_t(s) = \left\| \sum _{i \ge 2} \omega_i^t \varphi_i(s) \tilde{\varphi}_i\right\|^2, \forall s \in \bar{\mathbb{S}},
	\label{eq:g_t}
\end{equation}
where $\varphi_i,\tilde{\varphi}_i:\bar{\mathbb{S}}\rightarrow \mathbb{R}$ are out-of-sample extensions of $\phi_i$ and $\tilde{\phi}_i$, such that 
\begin{equation*}
	\varphi_i(s) = \phi_i(s), \tilde{\varphi}_i(s) = \tilde{\phi}_i(s), \forall s \in \mathbb{S}.
\end{equation*}
By definition, the extension is consistent in the sense that
\begin{equation*}
	g_t(s) = f_t(s), \forall s \in \mathbb{S}.
\end{equation*}

There exist many approaches and methods for out-of-sample extension of the eigenvectors of the graph Laplacian in the field of manifold learning, e.g. the classical Nystr\"{o}m extension \citep{fowlkes2001efficient,williams2001using} and geometric harmonics \citep{coifman2006geometric}. More recent extension methods also include methods based on neural networks \citep{chui2018deep,mishne2017diffusion}, and in the context of reinforcement learning and representation learning, a scalable approach  \citep{wu2018laplacian} using the spectral graph drawing objective \citep{koren2003spectral} while ensuring orthonormality.

The rationale behind the extension considered in (\ref{eq:g_t}) comes from the realm of manifold learning.
Suppose that the domain is a manifold and that the set of states is a sample from this continuous space. In such a setting, the graph can be viewed as a discrete proxy of the manifold, and the graph Laplacian as a discrete approximation of the Laplace-Beltrami operator of the manifold (see \citep{coifman2006diffusion,belkin2007convergence}). The particular interest in the Laplace-Beltrami operator stems from the fact that it contains all the geometric information on the manifold \citep{berard1994embedding}.
Under this setting, the two sets of states $\sS$ and $\bar{\sS}$ are just two samples from the manifold. Since their respective graph Laplacians are approximations of the same Laplace-Beltrami operator, it is reasonable to assume that the eigenvalues remain the same, and the eigenvectors are interpolations of one to the other. 

Once $g_t(s)$ is defined using the extended eigenvectors, and options are computed, the remaining question is how to compute the policy of each option, e.g., by approximating the Q function for $s \in \bar{\mathbb{S}}$. 
Building a policy that leads the agent toward any particular state in the domain can be implemented in multiple ways, e.g., using deep Q learning (DQN) architectures, which have shown success in much more involved tasks \citep{hessel2018rainbow,schaul2015universal}. Such an approach was recently demonstrated for cover options by \citet{jinnai2020exploration}.

The above description implies that the extension of the proposed method requires the extension of the eigenvectors of the graph Laplacian.
As described in Section 3.2 in the paper, these extensions in a similar setting have already been considered and tested \citep{wu2018laplacian,jinnai2020exploration,machado2017eigenoption,chui2018deep,mishne2017diffusion}.
Therefore, in principle, we can use any one of these methods.

% TODO: perhaps the following paragraph could be removed if not implemented and tested
%Another approach for computing the value function for each option is based on diffusion distance. Once an option is invoked, the diffusion distance between the current state and all other states can be computed using the extended eigenvectors $\psi_i$ and $\tilde{\psi}_i$. The value function associated with the option is set to the value of the diffusion distance between the current state and other states for a chosen $t$ value. This simplifies the value function calculation at the expense of online calculation.

% TODO: include? Eigenoptions estimation (which is essentially the estimation of the graph Laplacian's eigen functions $\{\psi(s)\}$) was extended to stochastic domains in \cite{machado2017eigenoption} based on raw pixels, using deep successor representation. As $f_t(s)=h(\{\psi(s)\})$ the same method could be applied.

While the extension is presented in the context of partially known models, it can also be applied as is for scaling up.
The complexity of Algorithm 1 for discovering the diffusion options is governed by the EVD applied to the lazy random walk matrix (or the normalized graph Laplacian). In domains with large state spaces, or even with continuous state spaces, the construction can be based on a small representative set of states and then extended to the entire domain. Such an approach was proposed by \citet{wu2018laplacian}, \citet{jinnai2020exploration}, and \citet{machado2017eigenoption}.

\section{The Stationary Distribution and $f_t(s)$ in a Multidimensional Grid Domain}

We further demonstrate the relations between the stationary distribution $\pi_0$ and $f_t(s)$ specifically in a multidimensional grid domain. 
Consider an $n$-dimensional grid domain with a set of states $\sS$.
Such a domain can be represented by the product of $n$ path graphs. Let $G^{(m)}$ be a path graph with a node set $\sS^{(m)}$ consisting of $N_m$ nodes. Then, the product graph $G=G^{(1)}\times G^{(2)} \times \cdots \times G^{(n)}$ defined on the node set $\sS = \sS^{(1)} \times \sS^{(2)} \times \cdots \times \sS^{(n)}$ represents the $n$-dimensional grid domain.

The stationary distribution at state $s$ is given by $\pi_0(s)=\frac{\vd(s)}{d(\sS)}$ \citep{DanielSpielmanLectures}, namely the stationary distribution at a certain state is proportional to its degree. This immediately yields that $\underset{s}{\arg\min}\{\pi_0(s)\}=\underset{s}{\arg\min}\{\vd(s)\}$, which implies that the minima of the stationary distribution are located at states with minimal degree. In the grid domain, those states are the corner states. 

\begin{figure}
\center
\subfigure[$f_4(s)$]{\includegraphics[width=.29\textwidth]{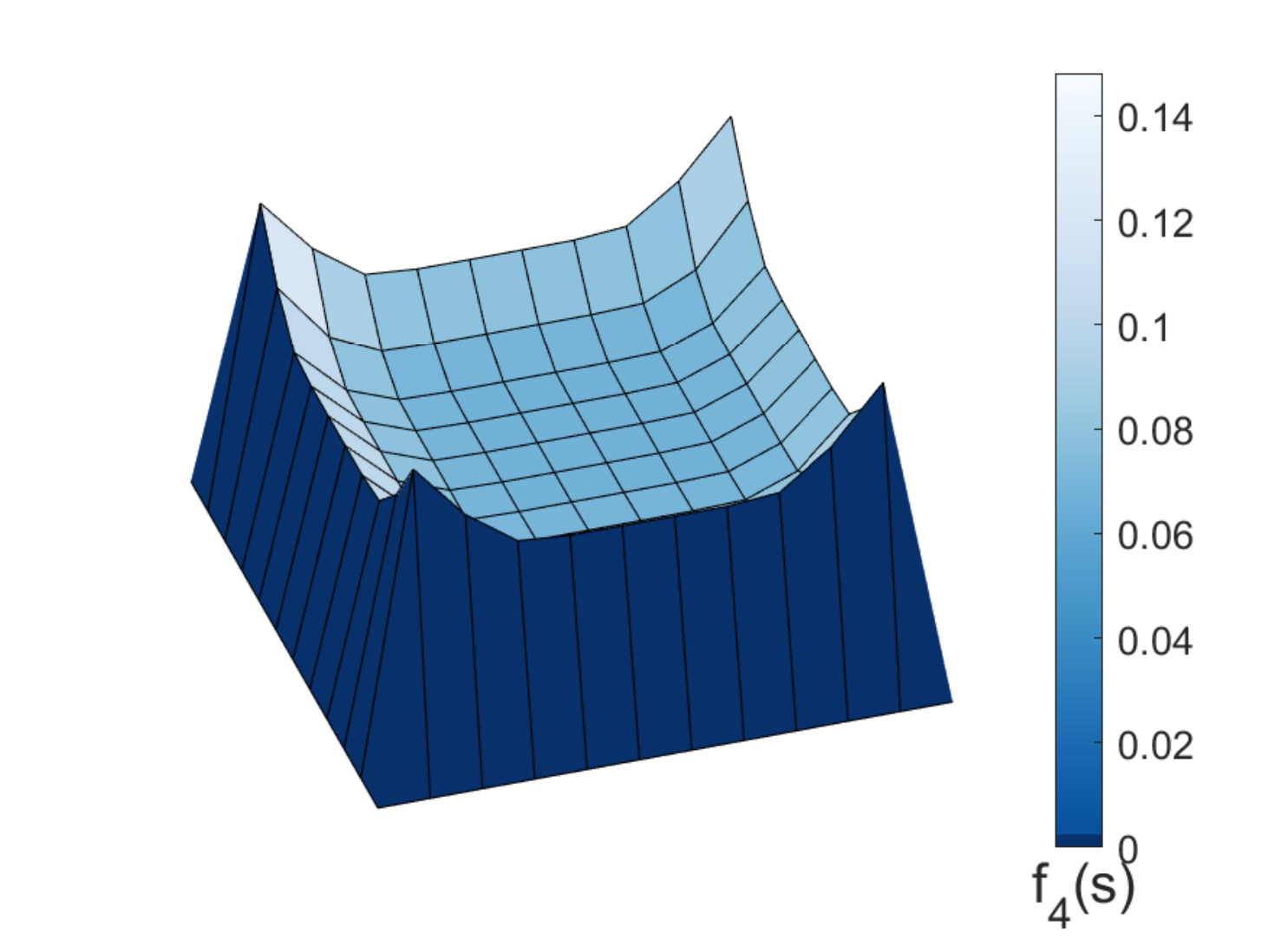}}
\subfigure[$f_{13}(s)$]{\includegraphics[width=.29\textwidth]{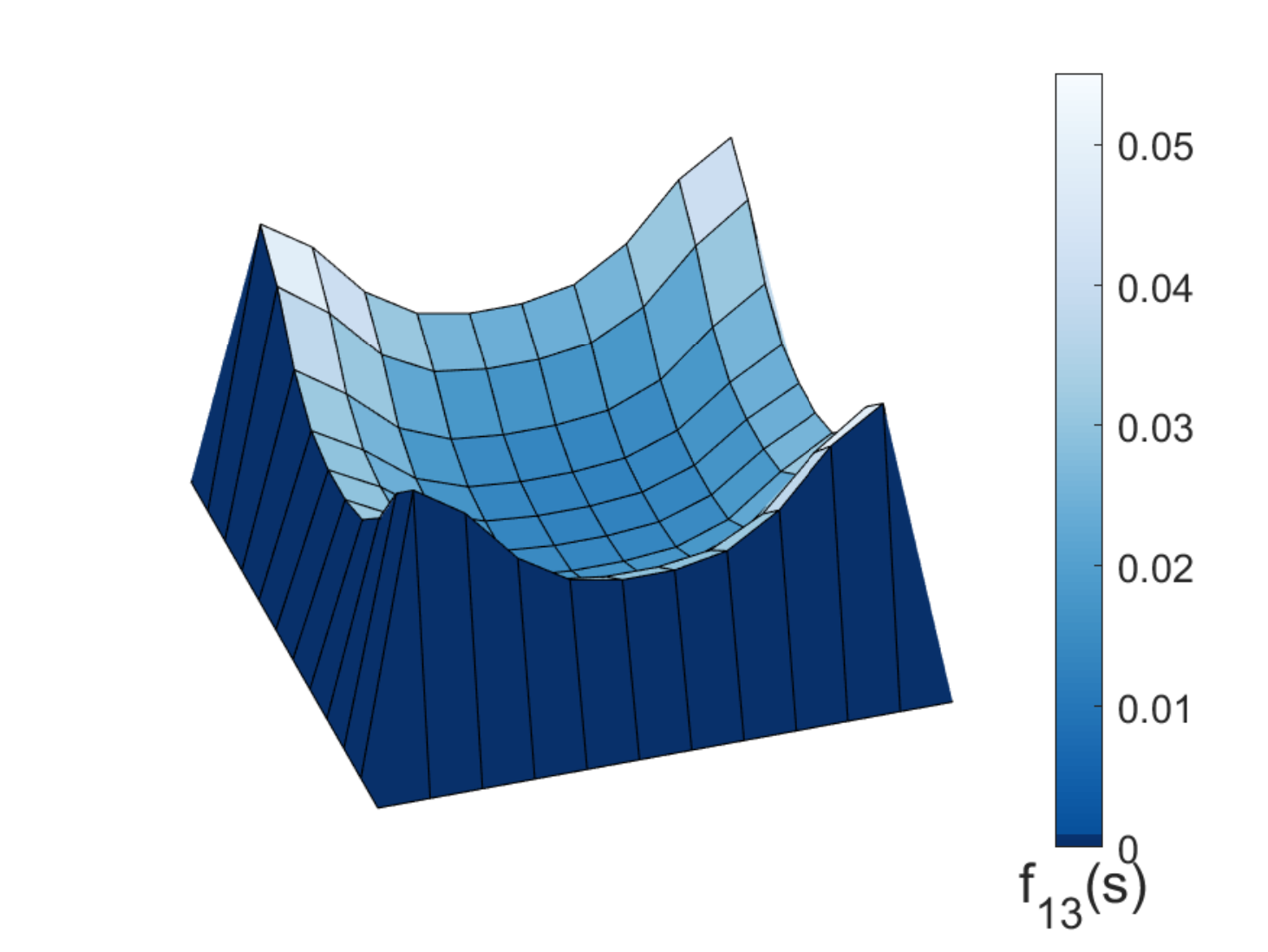}}
\subfigure[$f_{100}(s)$]{\includegraphics[width=.29\textwidth]{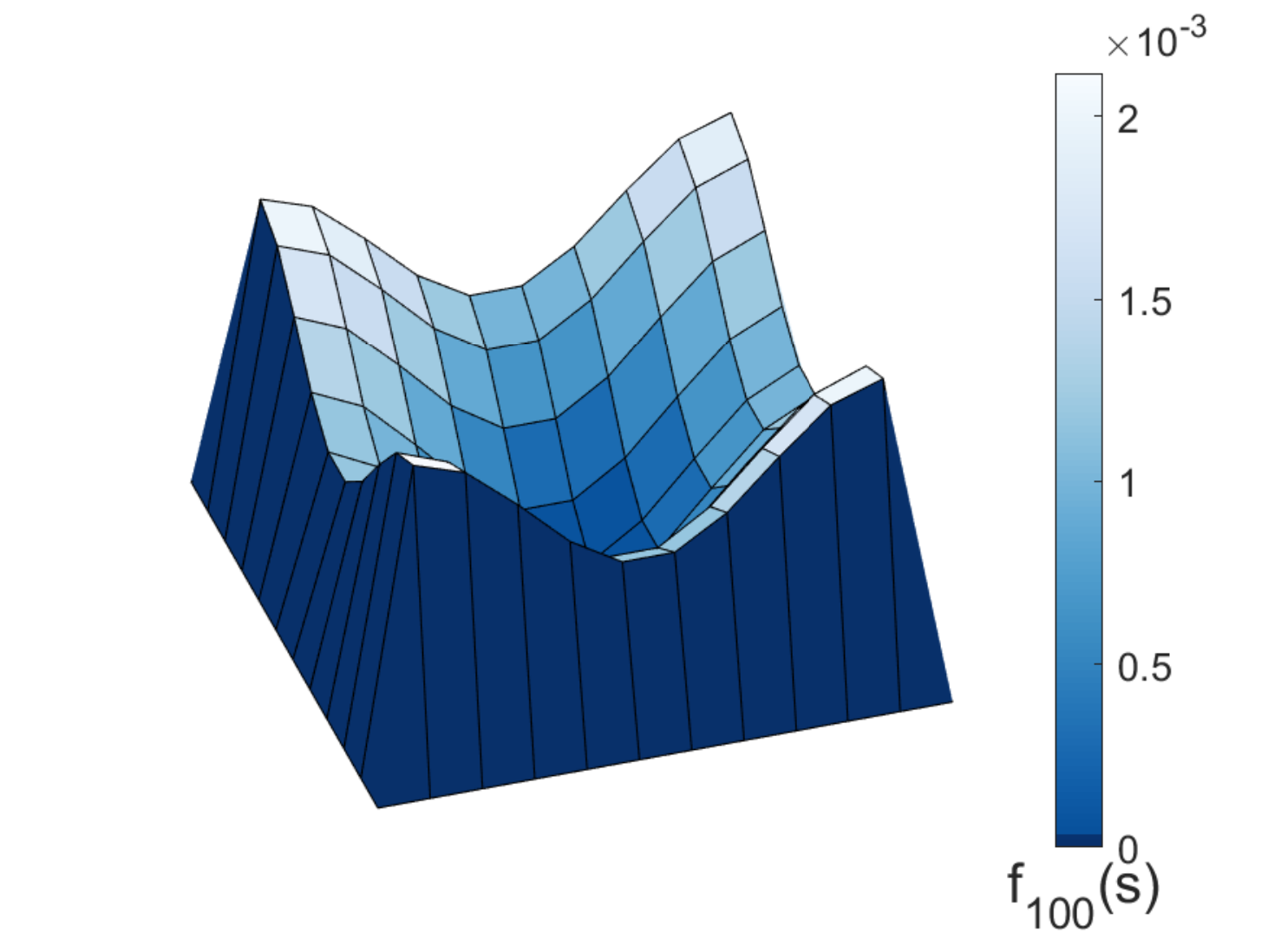}}
\caption{$f_t(s)$ for different scale parameter $t$ values for the $2$D grid domain. As $t$ increases, $f_t(s)$ becomes smoother, yet the corner states remain the maxima. We note that here we use a 3D representation for a better visualization of the maxima.}
\label{Fig:f_tForDiffTimes}
\end{figure}

We demonstrate in a simulation that the maxima of $f_t(s)$ in a $2$D grid domain are located at the corners as well. 
Figure \ref{Fig:f_tForDiffTimes} shows $f_t(s)$ for  $t=4$, $t=13$, and $t=100$. Indeed, we observe that $f_t(s)$ has maxima at the corners. In addition, we observe that $f_t(s)$ assumes a similar shape for the different $t$ values, and that higher $t$ values lead to a smoother function.

In addition, the corner states of a grid domain admit another property that makes them good option goal states candidates. To show this, we turn to the continuous analogue of the discrete grid domain and focus on $2$ dimensions for simplicity.
Namely, we present the result in the $2$D $[0,L]^2$ domain.
\begin{prop}
The expected distance between a uniformly distributed state to the corners of a grid domain is the largest compared to the expected distance to any other state. 
Concretely, let $\vx_s \in [0,L]^2$ sampled uniformly at random.
We have
$$
\mathbf{E}_{\vx_s\in[0,L]^2}\mathbf{E}_{\vx_c\in\{0,L\}^2}\left[\Vert \vx_s- \vx_c\Vert^2\right]=
\mathrm{max}_{\vx_o \in [0,L]^2}\mathbf{E}_{\vx_s\in [0,L]^2}\left[\Vert \vx_s-\vx_o\Vert^2\right].
$$
\label{prop:PropEucDistGridDomain}
\end{prop}
Proposition \ref{prop:PropEucDistGridDomain} suggests that by moving to the corners, the agent covers maximal distance. In the discrete counterpart, this could imply that by doing so, the agent visits the maximal number of states, thereby encouraging exploration.
\begin{proof}
First, we compute:
\begin{equation*}
   \mathbf{E}_{\vx_c\in\{0,L\}^2}\left[\Vert\vx_s-\vx_c\Vert^2\right]=
x_s^2+y_s^2+L^2-Lx_s-Ly_s 
\end{equation*}
Then, the left hand side can be recast as:
\begin{align*}
    \mathbf{E}_{\vx_s\in[0,L]^2}\mathbf{E}_{\vx_c\in\{0,L\}^2}\left[\Vert\vx_s-\vx_c\Vert^2\right] &=
\int_0^{L}\int_0^{L}\frac{1}{L^2}\mathbf{E}_{\vx_c\in\{0,L\}^2}\left[\Vert\vx_s-\vx_c\Vert^2\right]dx_sdy_s \\
&= \frac{1}{L^2}\int_0^{L}\int_0^{L}\left[x_s^2+y_s^2+L^2-Lx_s-Ly_s\right]dx_sdy_s=\frac{2}{3}L^2.
\end{align*}

We continue with the right hand side:

\begin{align*}
    \mathbf{E}_{\vx_s\in[0,L]^2}\left[\Vert\vx_s-\vx_o\Vert^2\right]&= 
\int_0^{L}\int_0^{L}\frac{1}{L^2}\left(x_o^2+y_o^2+x_s^2+y_s^2-2x_ox_s-2y_oy_s\right)dx_sdy_s \\
&=x_o^2+y_o^2-Lx_o-Ly_o+\frac{2}{3}L^2.
\end{align*}
This expression gets its maximum value at $x_o=\{0,L\}$ and $y_o=\{0,L\}$, so we have

\begin{equation*}
    \mathrm{max}_{\mathbf{x}_o}\mathbf{E}_{\vx_s\in[0,L]^2}\left[\Vert\vx_s-\vx_o\Vert^2\right]=\frac{2}{3}L^2
\end{equation*}
\end{proof}

\section{Code}
The code is available at \href{https://github.com/amitaybar/Diffusion-options}{https://github.com/amitaybar/Diffusion-options}.

% \includepdf[pages=-,pagecommand={}]{Supplementary Material for Option Discovery in the Absence of Rewards with Manifold Analysis.pdf}
\end{document}